\documentclass{article}
\usepackage[utf8]{inputenc}
\usepackage{main}
\usepackage{microtype}
\usepackage{graphicx}
\usepackage{times}
\usepackage{amsmath}
\usepackage{amssymb}
\usepackage{enumitem}
\usepackage{booktabs}
\usepackage{xcolor}     
\usepackage{natbib}
\definecolor{mydarkblue}{rgb}{0,0.08,0.45}
\usepackage[colorlinks=true,linkcolor=mydarkblue,citecolor=mydarkblue,filecolor=mydarkblue,urlcolor=mydarkblue]{hyperref}
\usepackage{fancyhdr}
\usepackage{makecell}

\definecolor{rowgray}{RGB}{245,245,245}
\definecolor{lightred}{RGB}{190,40,40}
\definecolor{lightgreen}{RGB}{40,130,60}
\definecolor{lightblue}{RGB}{30,90,160}
\definecolor{gred}{RGB}{250, 210, 207}
\definecolor{coolblue1}{rgb}{0.91, 0.94, 0.98}
\definecolor{coolblue2}{rgb}{0.76, 0.85, 0.94}
\definecolor{coolblue3}{rgb}{0.54, 0.72, 0.87}
\definecolor{coolblue4}{rgb}{1, 1, 1}

\usepackage[table]{xcolor}  

\usepackage{xspace}
\usepackage{cleveref}
\usepackage[]{multicol, multirow}
\usepackage[most]{tcolorbox}
\usepackage{etoc}

\tcbuselibrary{listingsutf8}
\tcbuselibrary{listingsutf8,breakable}

\newtcolorbox[auto counter]{observation}[1][]{
  colback=black!5!white,
  colframe=black!70!white,
  fonttitle=\bfseries,
  title=Observation~\thetcbcounter,
  enhanced,
  boxrule=0.6pt,
  left=1mm,right=1mm,top=1mm,bottom=1mm,
  #1
}

\newtcolorbox[auto counter]{takeaway}[1][]{
  colback=teal!3!white,
  colframe=teal!55!black,
  fonttitle=\bfseries,
  title=Takeaway~\thetcbcounter,
  enhanced,
  boxrule=0.5pt,
  left=1mm,right=1mm,top=1mm,bottom=1mm,
  #1
}

\newtcolorbox[auto counter]{practicalguidance}[1][]{
  colback=cyan!3!white,
  colframe=cyan!60!black,
  fonttitle=\bfseries,
  title=Practical Guidance~\thetcbcounter,
  enhanced,
  boxrule=0.5pt,
  left=1mm,right=1mm,top=1mm,bottom=1mm,
  #1
}

\newtcolorbox[auto counter]{discussion}[1][]{
  colback=violet!4!white,
  colframe=violet!60!black,
  fonttitle=\bfseries,
  title=Discussion~\thetcbcounter,
  enhanced,
  boxrule=0.5pt,
  left=1mm,right=1mm,top=1mm,bottom=1mm,
  #1
}

\usepackage{caption}
\usepackage{placeins}
\usepackage{subcaption}
\usepackage{color}
\usepackage{tabularx}
\usepackage{tabu}
\usepackage{booktabs}
\usepackage{colortbl}
\usepackage{multirow}
\usepackage{longtable}
\usepackage{placeins}
\usepackage{todonotes}
\usepackage{pifont}
\usepackage{wrapfig}
\usepackage{pgfplots}
\pgfplotsset{compat=1.18}
\usepackage{tikz}
\usepackage{colortbl}
\definecolor{rowgray}{RGB}{245,245,245}
\definecolor{lightred}{RGB}{190,40,40}
\definecolor{lightgreen}{RGB}{40,130,60}
\definecolor{lightblue}{RGB}{30,90,160}
\usepackage{placeins}
\definecolor{darkblue}{rgb}{0, 0, 0.5}
\hypersetup{colorlinks=true, citecolor=darkblue, linkcolor=darkblue, urlcolor=darkblue}
\usepackage{booktabs}
\usepackage{multirow}
\usepackage[table]{xcolor}
\usepackage{colortbl}
\usepackage{tcolorbox}
\usepackage{xcolor}
\tcbuselibrary{skins, breakable}
\definecolor{rowgray}{RGB}{245,245,245}
\definecolor{lightred}{RGB}{190,40,40}
\definecolor{lightgreen}{RGB}{40,130,60}
\usepackage{booktabs}
\usepackage{multirow}
\usepackage{xcolor}
\usepackage{colortbl}
\definecolor{losegray}{gray}{0.88}

\definecolor{gptblue}{RGB}{16,110,190}
\definecolor{gptbluebg}{RGB}{219,234,248}
\definecolor{claudeteal}{RGB}{0,139,139}
\definecolor{claudetealbg}{RGB}{210,240,240}
\definecolor{gemred}{RGB}{200,60,40}
\definecolor{gemredbg}{RGB}{252,228,224}
\definecolor{qwenorange}{RGB}{210,120,0}
\definecolor{qwenorangebg}{RGB}{255,241,210}
\definecolor{deepgray}{RGB}{80,80,80}
\definecolor{deepgraybg}{RGB}{235,235,235}
\definecolor{openbrown}{RGB}{150,90,20}
\definecolor{openbrowrbg}{RGB}{250,235,210}
\definecolor{sftpurple}{RGB}{120,60,160}
\definecolor{sftpurplebg}{RGB}{237,224,252}
\definecolor{rlgreen}{RGB}{30,130,50}
\definecolor{rlgreenbg}{RGB}{210,244,218}
\definecolor{gtgreen}{RGB}{34,139,34}
\definecolor{gtgreenbg}{RGB}{220,245,220}
\definecolor{analysisbg}{RGB}{232,245,233}
\definecolor{analysisborder}{RGB}{46,125,50}
\usepackage{graphicx} 
\usepackage{float}
\usepackage{amssymb}

\usepackage{hyperref}
\usepackage{tablefootnote}
\usepackage{xspace}

\usepackage{float}
\usepackage{amsmath}
\usepackage{bm}
\usepackage{algorithmicx}
\usepackage{algorithm}
\usepackage{algpseudocode}

\usepackage{arydshln}
\usepackage{amssymb}
\usepackage{pifont}
\usepackage{enumitem}

\newcommand{\ours}{ActReview\xspace}
\newcommand{\data}{ActReview-40K\xspace}
\newcommand{\bench}{ActReview-Bench\xspace}
\newenvironment{itemize*}%
 {\leftmargini=10pt\begin{itemize}%
  \setlength{\itemsep}{0pt}%
  \setlength{\parskip}{0pt}%
  }%
 {\end{itemize}}
\newenvironment{enumerate*}%
 {\begin{enumerate}%
  \setlength{\itemsep}{0pt}%
  \setlength{\parskip}{0pt}}%
 {\end{enumerate}}

\usepackage[most]{tcolorbox}
\definecolor{myblue}{rgb}{0.18,0.45,0.73}

\usepackage[table]{xcolor}
\usepackage[most]{tcolorbox}

\definecolor{inputbg}{RGB}{248,250,252}
\definecolor{inputborder}{RGB}{148,163,184}

\definecolor{refgreenbg}{RGB}{240,253,244}
\definecolor{refgreen}{RGB}{22,101,52}

\definecolor{openbrownbg}{RGB}{254,243,199}
\definecolor{openbrown}{RGB}{146,64,14}

\definecolor{predbluebg}{RGB}{239,246,255}
\definecolor{predblue}{RGB}{30,64,175}

\definecolor{errorredbg}{RGB}{254,242,242}
\definecolor{errorred}{RGB}{185,28,28}

\definecolor{graybg}{RGB}{248,250,252}
\definecolor{grayborder}{RGB}{203,213,225}

\definecolor{grpours}{RGB}{231,245,255}   
\definecolor{grpllm}{RGB}{240,240,240}    
\definecolor{grpother}{RGB}{237,250,233}  

\usepackage{pgfplots}
\pgfplotsset{compat=1.18}
\usepackage{xcolor}

\usepackage{booktabs}
\usepackage{siunitx}
\definecolor{customblue}{HTML}{286dc0}

\definecolor{customgreen}{HTML}{2ca02c}

\usepackage{wrapfig}
\usepackage{multirow}
\newtcolorbox{blueBox}[1][]{
  enhanced,
  colback=customblue!5!white,
  colframe=customblue,
  boxrule=0.8pt,
  arc=1mm,
  left=5pt,
  right=5pt,
  top=6pt,
  bottom=6pt,
  title={\centering #1},
  width=\linewidth
}
\usepackage{soul}
\sethlcolor{green!25}
\newcommand{\bestbase}[1]{\colorbox{green!22}{#1}}

\newtcolorbox{analysisbox}[1][]{
    enhanced,
    colframe=customgreen,
    colback=customgreen!10!white,
    sharp corners,
    boxsep=0pt,
    left=5pt,
    right=5pt,
    top=6pt,
    bottom=6pt,
    boxrule=0pt,
    leftrule=4pt,
    width=\linewidth,
    #1
}

\usepackage{microtype}

\definecolor{YaleBlue}{RGB}{0, 53, 107}
\definecolor{UChiRed}{RGB}{128, 0, 0}
\definecolor{TCSC}{RGB}{1, 126, 199}

\newcommand{\huggingface}{\raisebox{-1.5pt}{\includegraphics[height=1.05em]{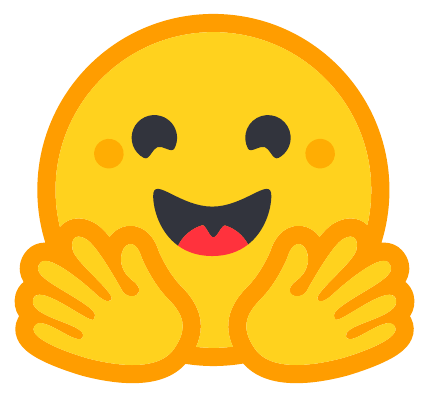}}\xspace}
\newcommand{\github}{\raisebox{-1.5pt}{\includegraphics[height=1.05em]{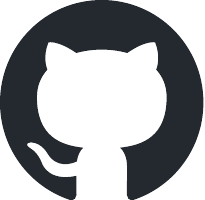}}\xspace}
\newcommand{\Yale}{\hspace{.1em}^{\textcolor{YaleBlue}{\boldsymbol{Y}}}}
\newcommand{\TCS}{\hspace{.1em}^{\textcolor{TCSC}{\boldsymbol{T}}}}
\newcommand{\UChicago}{\hspace{.1em}^{\textcolor{UChiRed}{\boldsymbol{C}}}}

\begin{document}

\title{\textsc{\ours}: Rebuttal-Guided Training Data and Rubric Rewards for Actionable Peer Review Generation}

\author{
\textbf{Yiling Ma}$\Yale$ \quad
\textbf{Yilun Zhao}$\Yale$* \quad
\textbf{Sihong Wu}$\Yale$ \quad
\textbf{Ziyu Chen}$\UChicago$ \quad
\textbf{Manasi Patwardhan}$\TCS$ \quad
\textbf{Arman Cohan}$\Yale$
\protect\\[7pt]
$\Yale$Yale University \quad
$\UChicago$University of Chicago \quad
$\TCS$TCS Research
\protect\\[6pt]
{\small *Correspondence to: Yilun Zhao (\texttt{yilun.zhao@yale.edu})}
}

\maketitle
\thispagestyle{fancy}
\fancyhead{}
\lhead{%
    \includegraphics[height=1.1cm]{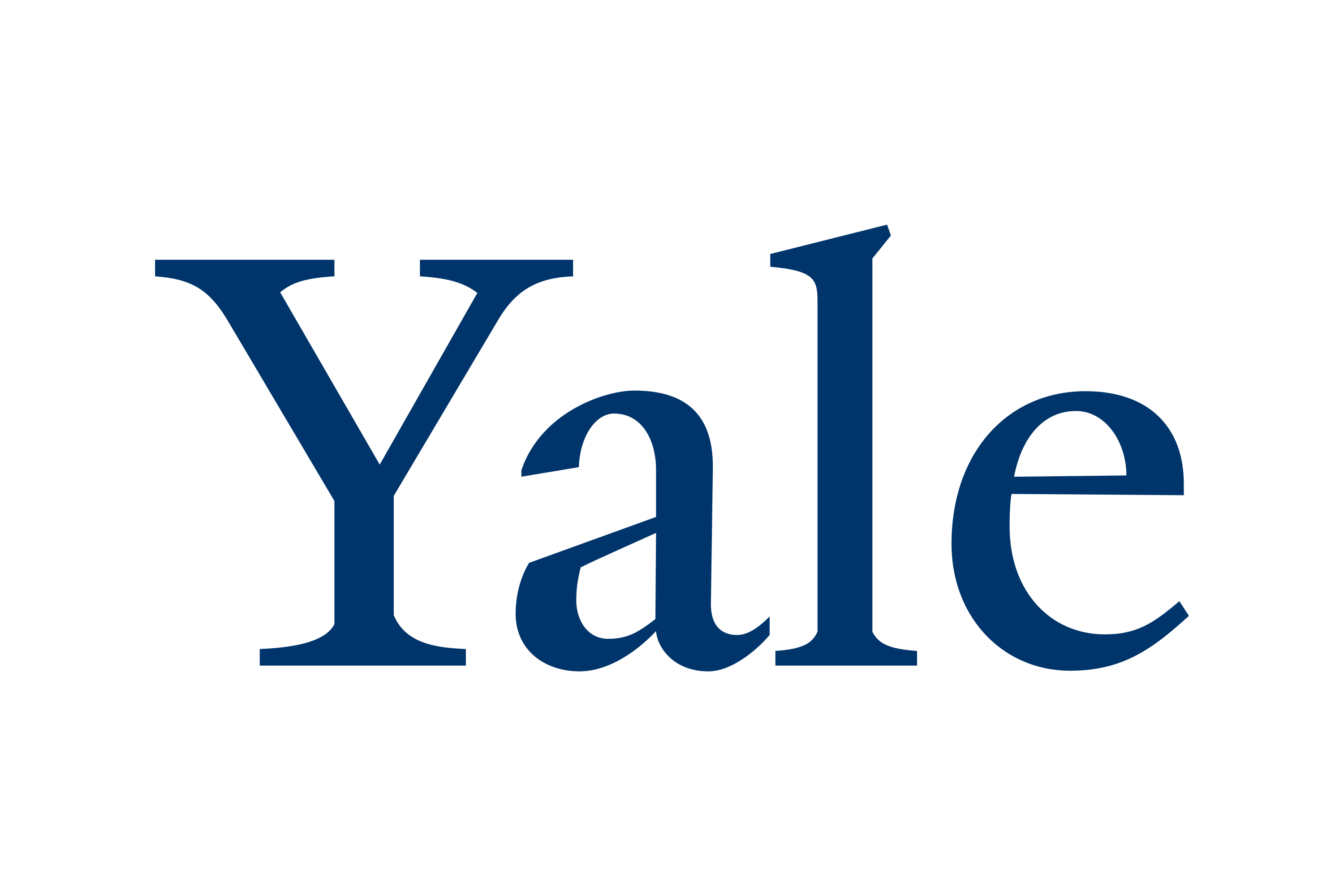}\hspace{0.2cm}%
    \includegraphics[height=1.05cm]{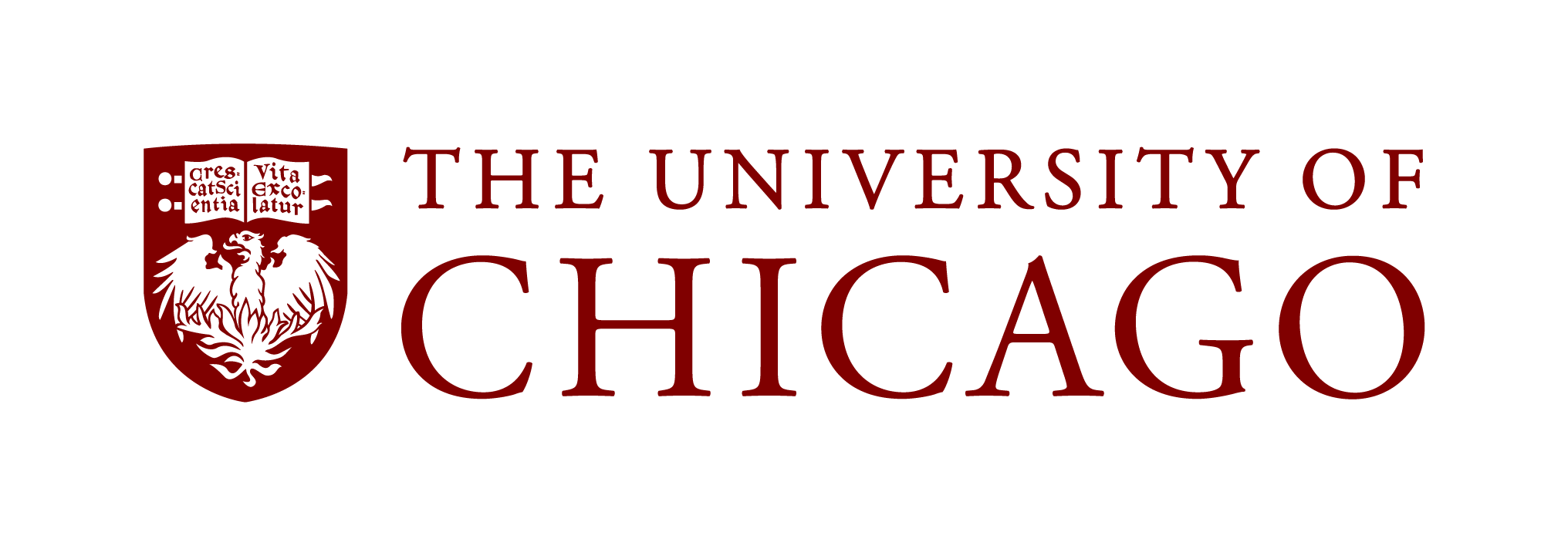}\hspace{0.3cm}%
    \raisebox{0.05cm}{\includegraphics[height=0.8cm]{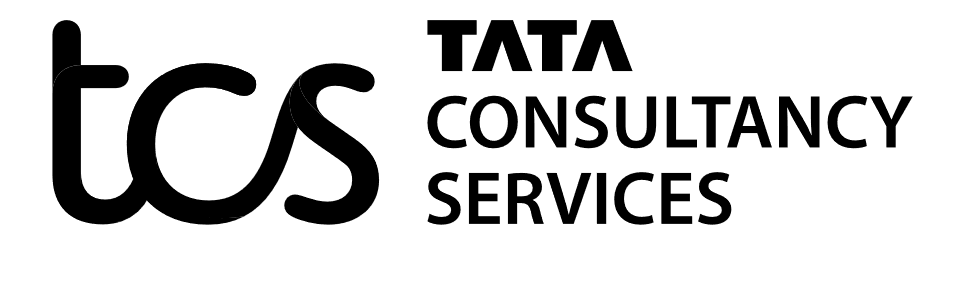}}%
}

\fancyfoot[C]{\thepage}
\renewcommand{\headrulewidth}{0pt}
\setlength{\headheight}{12pt}
\addtolength{\topmargin}{0pt}
\setlength{\headsep}{3mm}

\vspace{-1.0em}

\pagestyle{plain}

\begin{abstract}
As LLMs are increasingly used for pre-submission self-review, there is growing demand
for feedback that not only identifies weaknesses but also guides authors toward concrete
revisions. We study this as \emph{Actionable Peer-review Generation} and decompose it
into two subtasks: diagnostic claim generation and revision suggestion generation. We
introduce \ours, a rebuttal-guided post-training framework that connects paper-specific
diagnoses to concrete, grounded revision plans. Our central insight is that author
rebuttals reveal plausible actions for addressing reviewer concerns and can therefore
provide \emph{latent supervision} for revision-oriented feedback. From real
review--rebuttal threads on OpenReview, we construct \data by aligning reviewer
weaknesses with author responses and grounding the resulting feedback in localized paper
evidence. We post-train Qwen3-8B-Base with multi-task supervised fine-tuning followed by
GRPO using candidate-aware, weakness-specific rubric rewards. We also introduce \bench,
a human-curated benchmark of 1,000 instances for evaluating diagnostic quality and
revision usefulness. Experiments show that \ours outperforms prior specialized
review-generation models on actionability and grounding while remaining competitive with
strong prompt-based LLMs. Human evaluation confirms improved revision usefulness while
revealing a remaining gap in technical accuracy, and additional analyses support
generalization to held-out papers and robustness across independent judges.

\begin{center}
\begin{tabular}{cl@{\hspace{5em}}cl}
\huggingface & \href{https://huggingface.co/datasets/YilingMa/ActReview-40K}{\textbf{Data:} \data} &
\github & \href{https://github.com/Yiling-Ma/ActReview}{\textbf{Code:} \ours}
\end{tabular}
\end{center}
\vspace{5pt}
\end{abstract}

\begin{figure*}[h]
    \centering
    \includegraphics[width=0.95\textwidth]{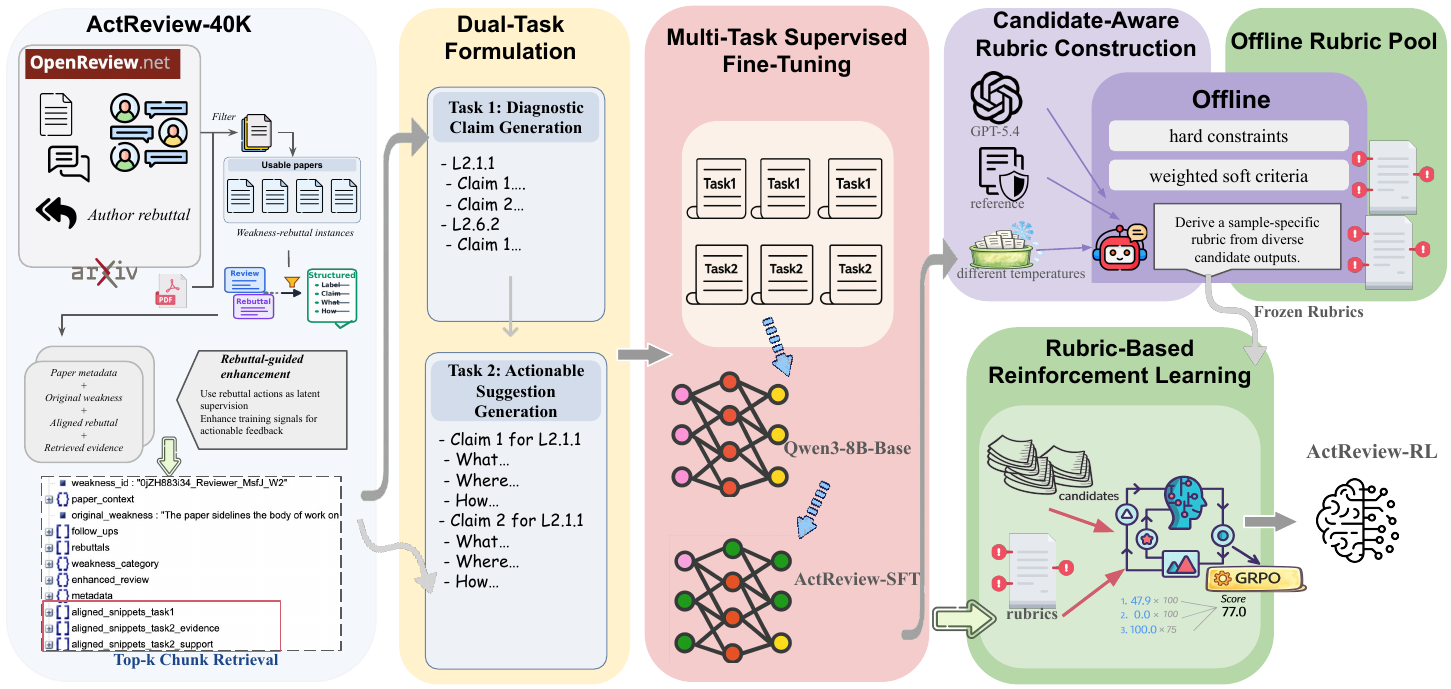}
    \caption{Overview of the \ours framework. From OpenReview's review–rebuttal threads, we construct ActReview-40K and retrieve top-k paper chunks as localized context. The model is first trained via multi-task supervised fine-tuning on Qwen3-8B-Base for claim generation (Task 1) and actionable suggestion generation (Task 2), then further optimized with GRPO using weakness-specific rubrics constructed offline from ground-truth references and diverse candidate outputs scored by an LLM judge.
    }
    \label{fig:overview}
\end{figure*}
\section{Introduction}
\label{sec:intro}

The rapid growth of scholarly publications has placed increasing pressure on the peer-review system. At the same time, reviewers and authors are increasingly exploring LLMs to support review writing and pre-submission feedback~\citep{liang2024mapping, wu2026can}. A key limitation is that existing LLM-based systems predominantly generate \emph{descriptive} rather than \emph{prescriptive} feedback: they can identify issues, but often fail to specify how those issues should be addressed~\citep{jin2024agentreview, gao2025reviewagents, wu2026rbtact}. This limits their usefulness when authors need concrete next steps for revision.

Recent work has begun to study review feedback that goes beyond problem identification~\citep{d2024marg,zhu2025deepreview,wu2026rbtact}. However, two gaps remain. First, existing formulations often treat review generation as a single task~\citep{idahl2025openreviewer}, without separating \emph{problem diagnosis} from \emph{revision guidance}. In practice, useful feedback must first identify what is wrong in the paper, and then explain how the issue can be revised in a grounded and implementable way. Second, existing peer-review resources \citep{zhu2025deepreview, zhang2025re, zhu2025your}
rarely provide reference signals that explicitly connect reviewer-identified weaknesses with concrete revision guidance. As a result, it is difficult to evaluate whether a generated comment merely diagnoses a problem or also provides useful guidance about what to revise, where to revise, and how to carry out the revision.

To address these gaps, we formulate \emph{Actionable Peer-review Generation} as a dual-task problem. Given a paper and a target weakness category, the model generates \emph{diagnostic claims} that identify concrete paper deficiencies and \emph{actionable suggestions} that describe how those deficiencies should be addressed. To support this setting, we construct \bench, a benchmark grounded in real review--rebuttal threads. We use rebuttal-derived author actions as grounded reference signals, since rebuttals often reveal plausible ways to address reviewer concerns. This enables evaluation of both diagnostic quality and practical revision usefulness.

We then present \ours, a rebuttal-guided post-training framework for this setting. We construct \data by aligning atomic reviewer weaknesses with rebuttal spans and transforming them into structured initial-review-style feedback. The rebuttal is used only as latent supervision: it helps infer a concrete revision path, but the resulting feedback is written from the perspective of an initial reviewer. Unlike prior systems that condition on full-paper inputs~\citep{zhu2025deepreview, idahl2025openreviewer}, our framework retrieves localized paper evidence for the target weakness, reducing irrelevant context and providing more focused supervision.

We fine-tune Qwen3-8B-Base~\citep{yang2025qwen3} on \data with multi-task supervised fine-tuning. Since supervised models can still produce fluent but generic comments, we further apply GRPO~\citep{shao2024deepseekmath} with candidate-aware, weakness-specific rubric rewards. These rubrics provide instance-level criteria for assessing diagnostic precision, grounding, and revision usefulness, encouraging the model to connect paper-specific weaknesses with concrete revision plans.

Experiments on \bench{} show that \ours{} improves over prior specialized
review-generation models~\citep{zhu2025deepreview,idahl2025openreviewer,wu2026rbtact}
and remains competitive with strong prompt-based LLMs. Human evaluation
shows stronger revision-oriented performance while identifying remaining
challenges in technical accuracy. Controlled and independent analyses further
validate the two-task formulation, generalization to held-out 2025--2026
papers, and robustness across human and independent LLM judges
(Appendix~\ref{app:additional_analysis}).

Our main contributions are as follows:
\begin{itemize}[leftmargin=*]
    \item We formulate \emph{Actionable Peer-review Generation} as 
    a dual-task problem covering diagnostic claim generation and 
    actionable suggestion generation, and introduce \bench, a 
    human-curated benchmark for evaluating revision-oriented feedback 
    (\S\ref{sec:task}).

    \item We construct \data, a large-scale rebuttal-guided training dataset 
    that converts aligned weakness--response pairs into structured 
    initial-review-style feedback (\S\ref{sec:data}).
    
    \item We propose \ours, combining multi-task SFT with GRPO and
    candidate-aware, weakness-specific rubric rewards, and validate it
    through automatic, judge-based, and human evaluation~(\S\ref{sec:train}).
\end{itemize}
\section{Related Work}
\paragraph{Peer-review Generation and Assistance.}
Recent work has explored LLMs for peer-review assistance, including review generation,
rebuttal generation, reviewer--author interaction, and weakness discovery. Prompt-driven
and agent-based systems support tasks such as author response generation, review
discussion simulation, and critique discovery~\citep{ma2026paper2rebuttal, han2026drpg,
ruan2026author, jin2024agentreview, zou2026diagpaper}. Another line of work trains or
prompts models to generate review scores, strengths, weaknesses, questions, or full
reviews~\citep{weng2024cycleresearcher, gao2025reviewagents, wu2026rbtact,
sharmaintelliask, zhu2025deepreview, idahl2025openreviewer}. These systems improve
automated review assistance, but most formulations either treat review generation as a
single output task or condition on full-paper inputs. LimitGen~\citep{DBLP:conf/acl/Xu00VC25}
benchmarks paper limitation identification using a limitation taxonomy, while
AbGen~\citep{DBLP:conf/acl/0001CX0W0VC25} evaluates context-grounded ablation study
design. In contrast, our work separates
reviewer-side feedback into diagnostic claim generation and revision suggestion
generation, and uses review--rebuttal interactions to construct structured supervision
for both subtasks.
\paragraph{Post-training for Non-verifiable Tasks.}
Our work is also related to post-training for non-verifiable tasks, where output quality
cannot be measured by exact-match rewards. RLHF and related alignment methods often rely
on scalar preference signals, which can be expensive to collect and too coarse for
open-ended tasks requiring multi-dimensional judgments. Recent work therefore explores
LLM-based evaluators, self-rewarding methods, and reference-based judges for
open-ended evaluation and scalable reward modeling~\citep{zheng2023judging,
yuan2024self, liu2026examining, shi2026references, DBLP:conf/nips/ZhaoZHWBLTCDBZH25}.
Other work replaces generic scalar rewards with structured rubrics
or checklists to better capture task-specific quality criteria~\citep{gunjal2025rubrics,
viswanathan2025checklists}. At the same time, ranked or relative optimization methods have been explored to improve robustness under noisy or coarse reward signals~\citep{choi2026gopo}. 
Our method builds on rubric-based reward design, but constructs weakness-specific rubrics
for individual paper--weakness instances, using both reference outputs and model
candidate failures to define more targeted reward criteria.

\section{Actionable Peer-Review Generation}
\label{sec:task}

\subsection{Problem Formulation}
\label{sec:problem_formulation}

As illustrated in Figure~\ref{fig:overview}, we formulate \emph{Actionable Peer-review Generation} as a dual-task problem that covers both problem diagnosis and revision guidance. The input consists of a paper context $C$ and a target weakness label $w$. The paper context includes paper metadata and retrieved paper chunks. The weakness label $w$ specifies the type of concern to evaluate, such as missing or insufficient theoretical justification. These labels are drawn from our data-driven two-level weakness taxonomy, which is induced from peer-review data through LLM-assisted category discovery and human refinement. Full taxonomy details are provided in Appendix~\ref{app:taxonomy}.

\paragraph{Diagnostic claim generation.}
A \emph{diagnostic claim} is a brief reviewer-side statement that identifies what is wrong with the paper under the target weakness label, rather than how to fix it. Given $C$ and $w$, the model generates a set of claims
\[
\mathcal{C} = \mathcal{G}_{\mathrm{claim}}(w, C) = \{c_1, \ldots, c_k\}, \quad k \geq 0.
\]
Each claim should identify a concrete paper deficiency supported by the context. If the context does not support the target weakness, the model should return no claim ($k=0$) rather than hallucinate one.

\paragraph{Actionable suggestion generation.}
An \emph{actionable suggestion} is revision guidance associated with a diagnostic claim. Given a claim $c \in \mathcal{C}$, the weakness label $w$, and paper context $C$, the model generates
\[
r_{\mathrm{action}}^{(c)} = \mathcal{G}_{\mathrm{action}}(c, w, C).
\]
The suggestion specifies what should be revised, where the revision should appear, how it can be implemented, and what outcome the revision is expected to achieve.

This formulation separates two abilities that are often conflated in review generation: detecting a paper-specific weakness and translating it into a concrete revision plan. The two tasks can be trained and evaluated separately, while also supporting an end-to-end review-to-revision workflow. A controlled comparison with joint end-to-end generation shows better suggestion
quality and claim--suggestion alignment~(Appendix~\ref{app:formulation_ablation}).

\paragraph{Weakness-conditioned inference.}
Our formulation is weakness-conditioned: each instance queries one target weakness label $w$, rather than requiring the model to consider all labels simultaneously. For full-paper auditing, the model can be applied independently to each label in the taxonomy. In each run, the model generates diagnostic claims only when the paper contains evidence supporting the queried weakness; otherwise, it should abstain and return \texttt{None}. Importantly, \data and the main \bench contain only reviewer-raised weakness--paper pairs and therefore provide no explicit negative training examples. Appendix~\ref{app:abstention} constructs a separate held-out set of $N_{\text{neg}}{=}360$ screened non-supported weakness--paper pairs and evaluates this behavior as zero-shot abstention.
\subsection{\bench Construction}
\label{sec:benchmark}

Existing peer-review resources mainly preserve raw reviews, scores, or rebuttal discussions, but rarely provide structured reference signals that connect reviewer-identified weaknesses to concrete revision guidance~\citep{zhang2025re, wu2026rbtact, zhu2025deepreview, idahl2025openreviewer}. To address this gap, we construct \bench from real review--rebuttal threads, using rebuttal-derived author actions as grounded reference signals rather than unique gold answers.\footnote{Rebuttals are post-hoc author responses and may reflect one plausible revision path rather than the only valid recommendation. We therefore use rebuttal-derived actions as grounded reference signals and latent supervision, while acknowledging that alternative actionable suggestions may also be valid.}

We begin with 2,000 candidate instances sampled from 4,000 automatically aligned weakness--response pairs and retain 1,000 high-quality benchmark instances after human annotation. Each candidate instance is independently annotated by two annotators. Annotators judge whether the rebuttal span (1) directly and substantively addresses the reviewer concern, (2) contains a concrete revision-oriented action, and (3) when follow-up reviewer discussion is available, remains supported by the discussion. Annotators also mark the rebuttal span that captures the relevant author action. Disagreements on the final keep/filter decision or the marked span are resolved by a third adjudicator.

Table~\ref{tab:human_validation} summarizes the human validation results. Using adjudicated labels as the reference, benchmark filtering achieves strong agreement and high filtering quality, indicating that the retained instances reliably satisfy our benchmark criteria. Appendix~\ref{app:benchmark_details} covers annotation, retained/filtered examples, and benchmark scope. 


\begin{wraptable}{r}{0.50\textwidth}
\centering
\scriptsize
\setlength{\tabcolsep}{1.8pt}
\resizebox{\linewidth}{!}{%
\begin{tabular}{@{}l l c c c c@{}}
\toprule
\textbf{Stage} & \textbf{Setting} & \textbf{Prec.} & \textbf{Rec.} & \textbf{F1} & \textbf{$\kappa$} \\
\midrule
Weakness--Rebuttal Alignment
  & Filtered set & 0.95 & 0.93 & 0.94 & 0.85 \\
\midrule
Benchmark Filtering
  & Retained set & 0.93 & 0.86 & 0.89 & 0.84 \\
\bottomrule
\end{tabular}
}
\caption{Human validation of alignment and filtering quality. 
Alignment is evaluated against adjudicated gold spans with token-level IoU $\geq 0.5$ and filtering is evaluated against adjudicated keep/filter decisions. 
$\kappa$ reports inter-annotator agreement.}
\label{tab:human_validation}
\end{wraptable}

\begin{figure*}[t]
    \centering
    \includegraphics[width=\textwidth]{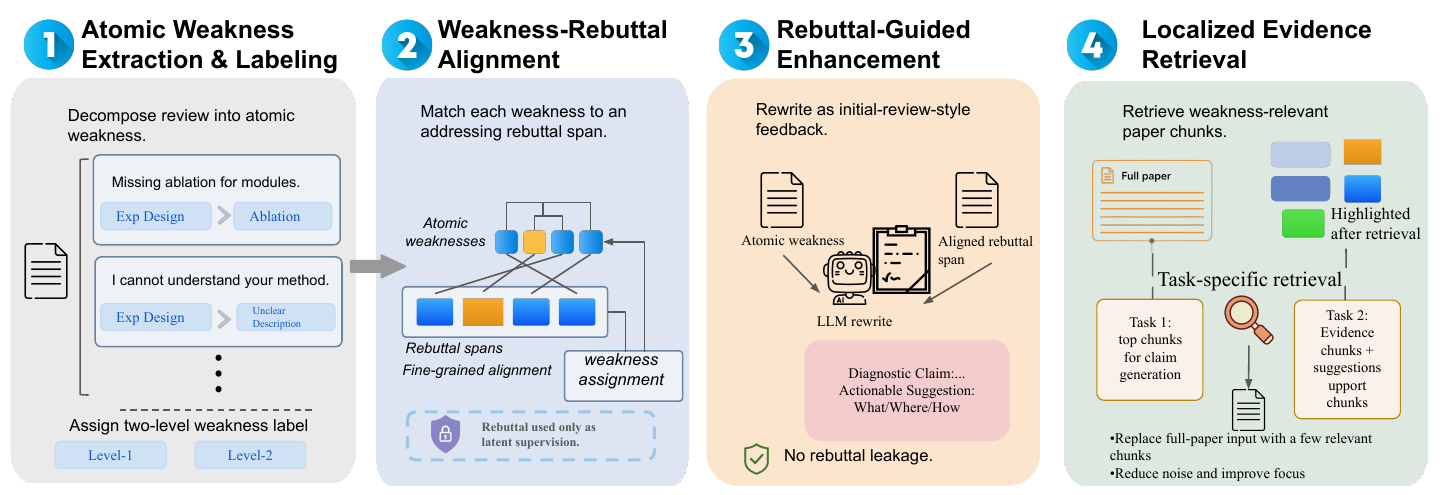}
    \caption{
    Overview of the ActReview-40K construction pipeline. Starting from real review--rebuttal threads, we build structured training instances through four stages: atomic weakness extraction and labeling, weakness--rebuttal alignment, rebuttal-guided feedback enhancement, and localized evidence retrieval. The aligned rebuttal span is used only as latent supervision to rewrite reviewer weaknesses into initial-review-style diagnostic claims and actionable suggestions, while the final training inputs contain only paper metadata, weakness labels, and localized paper evidence (at \bench{} evaluation time, localized evidence is replaced with full-paper context, as described in Section~\ref{sec:retrieval}).
    }
    \label{fig:data_construction_pipeline}
\end{figure*}

\section{\data}
\label{sec:data}
We construct \data through a multi-stage pipeline to provide large-scale supervision for the two subtasks in Section~\ref{sec:task}, with an overview shown in \autoref{fig:data_construction_pipeline}.
Each instance starts from a real review--rebuttal thread and is converted into structured initial-review-style feedback. 
The construction has four steps: (1) extracting and labeling atomic reviewer weaknesses, (2) aligning each weakness with the rebuttal span that addresses it, (3) rewriting the aligned signal into diagnostic claims and revision suggestions, and (4) retrieving localized paper evidence.

\subsection{Data Sources and Weakness Taxonomy}

We collect review--rebuttal threads from OpenReview across ICLR, NeurIPS, and EMNLP (Table~\ref{tab:source_distribution}). 
After filtering for papers with usable reviewer concerns and author responses, we obtain 15,819 papers and approximately 40K weakness--response instances. 
Each weakness is assigned to a two-level taxonomy of paper weaknesses, induced from review data through LLM-assisted category discovery and human refinement. 
Full taxonomy definitions and validation details are provided in Appendix~\ref{app:taxonomy}.

\subsection{Weakness Extraction and Rebuttal Alignment}
\label{sec:alignment}

Because review paragraphs often contain multiple concerns, we decompose each review into \emph{atomic weakness units}, where each unit expresses one critique or request for improvement. 
This enables fine-grained alignment between reviewer concerns and author responses. 
We then align each atomic weakness to the rebuttal span that substantively addresses it using a two-stage procedure: candidate span retrieval based on structural and lexical cues, followed by LLM-based semantic alignment. 
Human validation on a stratified sample confirms strong alignment quality (Table~\ref{tab:human_validation}); protocol details and residual errors are provided in Appendix~\ref{app:alignment_quality}.

\subsection{Rebuttal-Guided Feedback Enhancement}
\label{sec:feedback_enhancement}

Raw reviewer comments often identify a problem without specifying a concrete revision path, while author rebuttals frequently reveal how the concern can be addressed. 
We use this signal as \emph{latent supervision}: the aligned rebuttal span helps infer a plausible revision action, but the generated feedback must be written from the perspective of an initial reviewer and must not mention the rebuttal, author response, or any post-submission change.

For each aligned weakness--rebuttal pair, we prompt GPT-5.4 to produce two structured fields. 
The first is a \emph{diagnostic claim}, which states the core paper-specific deficiency. 
The second is a set of \emph{revision suggestions}, which specify what should be revised, where the revision should appear, how it can be implemented, and what outcome the revision is expected to achieve. 
The aligned rebuttal span is included only as background evidence during data construction, and is never provided to the model during training or evaluation.

We apply automatic filters to remove enhancement artifacts, including rebuttal leakage, retrospective rewrites, near-copying of rebuttal spans, and underspecified suggestions. 
A human audit of 500 stratified enhanced instances yields substantial agreement on binary acceptability ($\kappa=0.78$) and an overall acceptability rate of 84.8\%, suggesting that most enhanced instances provide reliable supervision for revision-oriented feedback generation. 
Full prompts, filtering rules, and criterion-level audit results are provided in Appendix~\ref{app:construct_enhancement} and Appendix~\ref{app:enhancement_quality}.

\subsection{Localized Evidence Retrieval}
\label{sec:retrieval}

For \ours-40K construction, we retrieve localized paper chunks rather than
using the full paper as training context. Papers are converted into structured
text and segmented into paragraph-level chunks with page and section metadata.
Task~1 retrieves evidence for weakness diagnosis, while Task~2 uses
evidence-support and revision-support channels to identify chunks that ground
the concern and localize concrete revisions. Rebuttal-derived fields are used
only offline as latent supervision for retrieval; the selected content is
extracted entirely from the paper and contains no rebuttal text. At
\bench{} evaluation time, all systems receive the same full-paper context,
paper metadata, and target weakness label. Retrieval validation and analysis
of globally distributed evidence are provided in
Appendix~\ref{app:retrieval} and Appendix~\ref{app:global_context}.
\section{Post-Training Framework}
\label{sec:train}

We train the model in two stages. First, supervised fine-tuning (SFT) teaches the model the dual-task output format and reviewer-style generation. Second, GRPO post-training uses weakness-specific rubric rewards to encourage diagnostic precision, contextual grounding, and revision usefulness. We partition the enhanced instances into 90\% for SFT and 10\% for RL. In the RL stage, reference claims and suggestions are not used as direct supervision; they are used only offline to construct frozen instance-specific reward rubrics for RL training instances. The RL training and rubric-construction instances are disjoint from \bench{} evaluation instances.

\subsection{Multi-Task Supervised Fine-Tuning}

We perform full-parameter SFT on Qwen3-8B-Base using the enhanced instances from Section~\ref{sec:feedback_enhancement}. Task~1 maps a weakness label and retrieved paper chunks to diagnostic claims, while Task~2 maps a diagnostic claim and related paper context to structured revision suggestions. The two tasks share a unified instruction format and are trained jointly within a single multi-task model. We optimize the standard autoregressive objective over the enhanced training set $\mathcal{D}_{\text{SFT}}$. Although SFT learns the schema and reviewer-style phrasing, it can still produce fluent but generic feedback, motivating a second stage with more targeted reward signals.

\subsection{Candidate-Aware Rubric Construction}
\label{sec:rubric}

Generic rubrics apply the same criteria across instances and cannot specify which evidence, experiment, or revision location matters for a particular paper. We therefore construct candidate-aware weakness-specific rubrics. For each RL instance, we sample diverse SFT outputs and combine them with a human-written reference and a GPT-5.4-generated reference. GPT-5.4 then synthesizes a frozen rubric with hard constraints and weighted soft requirements. This rubric captures both desirable revision targets and common failure modes in candidate outputs. Appendix~\ref{app:rubric} provides examples, and Section~\ref{sec:ablation} evaluates the effect of reward design.

\subsection{Rubric-Based Reinforcement Learning}

We clarify that only the SFT-trained model is further optimized with GRPO and all other baselines are not subject to RL training. Specifically, \ours-SFT serves as the initialization policy, and GRPO is applied to obtain \ours-RL ~\citep{shao2024deepseekmath}. For each prompt, the policy samples $K{=}8$ responses, each scored by the frozen rubric. Outputs violating hard constraints receive zero reward; otherwise, the reward is computed from weighted soft-requirement scores, using deterministic checks for format-related criteria and GPT-5.4 for semantic criteria. GRPO updates the policy with group-relative rewards and KL regularization toward the SFT reference model. 
\begin{table*}[!t]
\centering
\scriptsize
\addtolength{\tabcolsep}{-0.55em}
\resizebox{\textwidth}{!}{%
\begin{tabular}{l cccccc cccccc}
\toprule
\multirow{2}{*}{\textbf{Baseline}}
& \multicolumn{6}{c}{\textbf{Task 1: Weakness claim discovery}}
& \multicolumn{6}{c}{\textbf{Task 2: Actionable suggestions}} \\
\cmidrule(lr){2-7} \cmidrule(lr){8-13}
& \multicolumn{3}{c}{\textbf{Content quality}}
& \multicolumn{3}{c}{\textbf{Task-specific}}
& \multicolumn{3}{c}{\textbf{Content quality}}
& \multicolumn{3}{c}{\textbf{Task-specific}} \\
\cmidrule(lr){2-4} \cmidrule(lr){5-7}
\cmidrule(lr){8-10} \cmidrule(lr){11-13}
& R-L\textsuperscript{$\uparrow$}
& Sem.sim\textsuperscript{$\uparrow$}
& BLEU\textsuperscript{$\uparrow$}
& $n$ claims
& Spec.\textsuperscript{$\uparrow$}
& Rubric\textsuperscript{$\uparrow$}
& R-L\textsuperscript{$\uparrow$}
& Sem.sim\textsuperscript{$\uparrow$}
& BLEU\textsuperscript{$\uparrow$}
& Sugg.sim\textsuperscript{$\uparrow$}
& Evid.spec\textsuperscript{$\uparrow$}
& Rubric\textsuperscript{$\uparrow$} \\
\midrule
GPT-5.1
& 16.32 & \bestbase{49.69} & 5.42 & 2.84 & 0.31 & \bestbase{0.52}
& 26.86 & \bestbase{83.71} & 23.46 & 1.62 & 0.61 & \bestbase{0.59} \\
Gemini-3.1-Pro-Pre
& 17.64 & 48.30 & 5.92 & 2.91 & 0.27 & 0.48
& 28.16 & 81.03 & 25.90 & 2.00 & 0.72 & 0.58 \\
\midrule
DeepReviewer-14B
& 16.11 & 44.74 & 3.60 & 2.93 & \bestbase{0.39} & 0.30
& 25.01 & 67.54 & 13.73 & 1.95 & 0.00 & 0.12 \\
OpenReviewer-8B
& 19.21 & 35.92 & 7.30 & 3.71 & 0.10 & 0.19
& 23.21 & 70.72 & 4.50 & 0.00 & 0.00 & 0.00 \\
RbtAct-8B
& \bestbase{23.89} & 37.78 & \bestbase{10.54} & 2.67 & 0.36 & 0.44
& 27.21 & 75.78 & \bestbase{28.73} & 1.81 & \bestbase{1.57} & 0.57 \\
Qwen3-32B
& 22.07 & 36.13 & 8.85 & 1.60 & 0.20 & 0.40
& \bestbase{30.27} & 69.20 & 25.41 & \bestbase{2.01} & 1.20 & 0.54 \\
\midrule
\ours-SFT
& 24.60$^{\scriptstyle\textcolor{lightred}{+0.71}}$
& 47.90$^{\scriptstyle\textcolor{lightgreen}{-1.79}}$
& 11.12$^{\scriptstyle\textcolor{lightred}{+0.58}}$
& 1.74
& 0.96$^{\scriptstyle\textcolor{lightred}{+0.57}}$
& 0.59$^{\scriptstyle\textcolor{lightred}{+0.07}}$
& 35.50$^{\scriptstyle\textcolor{lightred}{+5.23}}$
& \textbf{84.10}$^{\scriptstyle\textcolor{lightred}{+0.39}}$
& 34.10$^{\scriptstyle\textcolor{lightred}{+5.37}}$
& 2.40$^{\scriptstyle\textcolor{lightred}{+0.39}}$
& 3.30$^{\scriptstyle\textcolor{lightred}{+1.73}}$
& 0.71$^{\scriptstyle\textcolor{lightred}{+0.12}}$ \\
\rowcolor{rowgray}
\textbf{\textcolor{lightred}{\ours-RL}}
& \textbf{28.24}$^{\scriptstyle\textcolor{lightred}{+4.35}}$
& \textbf{52.10}$^{\scriptstyle\textcolor{lightred}{+2.41}}$
& \textbf{12.04}$^{\scriptstyle\textcolor{lightred}{+1.50}}$
& \textbf{1.24}
& \textbf{0.57}$^{\scriptstyle\textcolor{lightred}{+0.18}}$
& \textbf{0.65}$^{\scriptstyle\textcolor{lightred}{+0.13}}$
& \textbf{37.54}$^{\scriptstyle\textcolor{lightred}{+7.27}}$
& 83.89$^{\scriptstyle\textcolor{lightred}{+0.18}}$
& \textbf{36.78}$^{\scriptstyle\textcolor{lightred}{+8.05}}$
& \textbf{2.54}$^{\scriptstyle\textcolor{lightred}{+0.53}}$
& \textbf{3.39}$^{\scriptstyle\textcolor{lightred}{+1.82}}$
& \textbf{0.78}$^{\scriptstyle\textcolor{lightred}{+0.19}}$ \\
\bottomrule
\end{tabular}%
}
\caption{
Main results on \bench for Task 1 and Task 2. All models are evaluated on the same test set. LLM-based evaluation is conducted on 1,000 samples, while human evaluation is conducted on a 200-sample subset for annotation cost reasons. \textbf{Rubric} denotes the rubric-based LLM-judge overall score in $[0,1]$. The best-performing baseline for each metric is highlighted in green. Red superscripts denote the change of \ours-SFT and \ours-RL relative to the best baseline for that metric.
}
\label{tab:auto}
\end{table*}

\section{Experiments}
\label{sec:exp}
We evaluate whether \ours generates feedback that is paper-specific, grounded, and useful for revision. 
Our experiments address four questions: 
(1) how \ours compares with strong prompt-based LLMs and specialized review-generation systems; 
(2) whether LLM-judge results are supported by human evaluation; 
(3) whether gains are reflected in automatic and reference-based metrics; and 
(4) which components of data construction, retrieval, and reward design drive the improvements.
\subsection{Baselines}

We compare against two groups of baselines. 
First, we evaluate strong prompt-based LLMs, including GPT-5.1 and Gemini-3.1-Pro-Preview. 
Each is tested in both a free-form zero-shot setting and a schema-controlled setting that follows our diagnostic-claim format for Task~1 and the What/Where/How structure for Task~2. 
This controls for the possibility that improvements come merely from output formatting. 
Second, we compare with specialized or strong open-source review-generation systems, including DeepReviewer-14B~\citep{zhu2025deepreview}, OpenReviewer~\citep{idahl2025openreviewer}, RbtAct~\citep{wu2026rbtact}, and Qwen3-32B~\citep{yang2025qwen3}, evaluated by direct inference with their released checkpoints, without any \bench-specific fine-tuning.
These systems were pretrained on different data and, in most cases, different tasks. This is intentional: we measure their out-of-the-box transfer to \bench, so the reported gap partly reflects baseline transfer, not only capability.
We also report \textsc{\ours-SFT} and \textsc{\ours-RL}, corresponding to our supervised-only and full post-training variants. During \ours-Bench evaluation, all systems receive identical paper metadata,
the target weakness label, and full-paper context. 
Rebuttals are used only to construct data and rewards, never as evaluation input. 

\begin{table*}[!t]
\centering
\scriptsize
\setlength{\tabcolsep}{3.2pt}
\renewcommand{\arraystretch}{1.08}
\resizebox{\textwidth}{!}{%
\begin{tabular}{l cc cc cc cc @{\hspace{0.8em}} cc cc cc cc}
\toprule
\multirow{3}{*}{\textbf{Baseline}}
& \multicolumn{8}{c}{\textbf{Task 1: Diagnostic Claims}}
& \multicolumn{8}{c}{\textbf{Task 2: Actionable Suggestions}} \\
\cmidrule(lr){2-9} \cmidrule(lr){10-17}
& \multicolumn{2}{c}{\textbf{Tech. Acc.}}
& \multicolumn{2}{c}{\textbf{Depth}}
& \multicolumn{2}{c}{\textbf{Spec./Gnd.}}
& \multicolumn{2}{c}{\textbf{Overall}}
& \multicolumn{2}{c}{\textbf{Tech. Acc.}}
& \multicolumn{2}{c}{\textbf{Depth}}
& \multicolumn{2}{c}{\textbf{Action.}}
& \multicolumn{2}{c}{\textbf{Overall}} \\
\cmidrule(lr){2-3}\cmidrule(lr){4-5}\cmidrule(lr){6-7}\cmidrule(lr){8-9}
\cmidrule(lr){10-11}\cmidrule(lr){12-13}\cmidrule(lr){14-15}\cmidrule(lr){16-17}
& LLM & Hum.
& LLM & Hum.
& LLM & Hum.
& LLM & Hum.
& LLM & Hum.
& LLM & Hum.
& LLM & Hum.
& LLM & Hum. \\
\midrule

GPT-5.1
& 48.3 & 49.6 & 51.5 & 54.7 & 56.3 & 55.84 & 51.2 & 55.1
& 48.2 & 51.3 & 53.6 & 55.3 & 57.1 & 56.3 & 51.9 & 56.8 \\

GPT-5.1$_{\text{schema}}$
& 55.7 & 53.3 & 53.6 & 57.8 & 56.8 & 58.9 & 55.1 & 54.3
& 56.3 & 56.9 & 55.4 & 57.2 & 64.0 & 59.7 & 58.8 & 57.1 \\

Gemini-3.1-Pro-Pre
& 50.3 & 50.1 & 54.9 & 55.6 & 55.5 & 60.6 & 53.0 & 56.5
& 54.3 & 53.1 & 52.1 & 53.3 & 57.0 & 63.5 & 55.5 & 53.2 \\

Gemini$_{\text{schema}}$
& 57.1 & 54.9 & 55.6 & 58.1 & 58.3 & 63.7 & 56.9 & 59.9
& 59.2 & 56.1 & 54.7 & 59.2 & 60.1 & 64.9 & 56.6 & 54.7 \\

RbtAct
& 61.8 & 64.4 & 55.7 & 57.8 & 62.2 & 58.7 & 57.3 & 62.0
& 57.5 & 58.3 & 70.2 & 67.5 & 53.3 & 54.5 & 56.2 & 59.8 \\

DeepReviewer-14B
& 75.5 & 76.2 & 68.2 & 71.9 & 67.8 & 66.1 & 70.4 & 72.8
& 69.7 & 69.4 & 68.2 & 70.9 & 84.4 & 77.3 & 68.2 & 71.9 \\

OpenReviewer-8B
& 67.5 & 69.1 & 65.3 & 65.8 & 70.1 & 69.1 & 66.7 & 66.0
& 63.2 & 64.7 & 72.0 & 70.5 & 80.1 & 74.5 & 67.8 & 72.7 \\

Qwen3-32B
& 67.2 & 69.8 & 62.6 & 61.3 & 72.5 & 67.2 & 65.9 & 62.7
& 56.0 & 61.6 & 56.5 & 54.3 & 57.9 & 59.7 & 60.0 & 57.6 \\

\bottomrule
\end{tabular}
}
\caption{
Pairwise LLM-as-a-Judge and human evaluation. Values report the adjusted win rate of \ours-RL against each baseline, where ties count as half a win:
$\text{Adj. Win Rate}=\text{Win}+0.5\times\text{Tie}$.
LLM columns use GPT-5.4 on 1,000 instances, while Hum. columns use human annotations on a 200-instance subset, enabling direct comparison under the same win-rate scale. GPT-5.1$_{\text{schema}}$ and Gemini$_{\text{schema}}$ denote schema-guided prompting variants with structured output constraints.
For Task~1, Tech. Acc. = Technical Accuracy, Depth = Depth \& Constructiveness, and Spec./Gnd. = Specificity \& Grounding.
For Task~2, Tech. Acc. and Depth have the same meanings, and Action. measures how directly useful the suggestion is for revision.
}
\label{tab:joint_pairwise_eval}
\end{table*}

\subsection{Evaluation Protocol}

We use three complementary evaluation protocols. 
First, we conduct pairwise LLM-as-a-judge evaluation using GPT-5.4 on the full evaluation set. 
For each instance, \ours-RL is compared with a baseline output under the same input. 
The judge selects \ours-RL, the baseline, or tie. 
We report adjusted win rate, where a win counts as 1, a tie as 0.5, and a loss as 0; thus, 50 indicates parity. 
Both tasks are evaluated on Technical Accuracy, Depth \& Constructiveness, and Overall quality, with one task-specific dimension: Specificity \& Grounding for Task~1 and Actionability for Task~2.

\paragraph{Human Evaluation.} We conduct human evaluation on a 200-instance subset using the same pairwise protocol. 
Each comparison is independently annotated by two graduate-level annotators with peer-review experience. 
Model identities are hidden and output order is randomized. 
The annotators make identical categorical choices in 93\% of pairwise judgments, suggesting high agreement.
Our primary comparisons do not rely on exact matching to the rewritten reference. 
The pairwise LLM and human evaluations compare model outputs directly under the same paper context and target weakness label, rather than scoring surface similarity to the reference. 
We use reference-based metrics only as complementary diagnostics of content overlap, granularity, and evidence specificity.

\paragraph{Automated Evaluation.} We also report automatic and reference-based metrics as complementary diagnostics. 
We include ROUGE-L, BLEU, sentence-level semantic similarity, and task-specific metrics such as claim count, specificity, suggestion similarity, evidence specificity, and rubric score. 
Because actionable feedback is open-ended, we treat these metrics as supporting evidence rather than as the sole measure of output quality. 
Metric definitions, judge prompts, bootstrap confidence intervals, and human annotation details are provided in Appendix~\ref{app:evaluation}.

\begin{table*}[!t]
\centering
\scriptsize
\setlength{\tabcolsep}{2pt}
\renewcommand{\arraystretch}{1.08}

\resizebox{\textwidth}{!}{%
\begin{tabular}{ll ccccc ccccc}
\toprule
\multirow{2}{*}{\textbf{Ablation}} 
& \multirow{2}{*}{\textbf{Variant}}
& \multicolumn{5}{c}{\textbf{Task 1: Diagnostic Claims}}
& \multicolumn{5}{c}{\textbf{Task 2: Actionable Suggestions}} \\
\cmidrule(lr){3-7} \cmidrule(lr){8-12}
& 
& R-L$\uparrow$ 
& Sem.sim$\uparrow$ 
& BLEU$\uparrow$ 
& $n$ claims 
& Spec.$\uparrow$
& R-L$\uparrow$ 
& Sem.sim$\uparrow$ 
& BLEU$\uparrow$ 
& Sugg.sim$\uparrow$ 
& Evid.spec$\uparrow$ \\
\midrule

\multirow{2}{*}{Data}
& Raw review data
& 26.40 & 42.20 & 10.81 & 1.00 & 0.27
& 23.90 & 66.71 & 6.21 & 1.00 & 0.05 \\

& Rebuttal-enhanced data
& 24.60{\tiny$^{\textcolor{red!70!black}{-1.80}}$}
& \textbf{47.90}{\tiny$^{\textcolor{green!60!black}{+5.70}}$}
& \textbf{11.12}{\tiny$^{\textcolor{green!60!black}{+0.31}}$}
& \textbf{1.74}
& \textbf{0.96}{\tiny$^{\textcolor{green!60!black}{+0.69}}$}
& \textbf{35.50}{\tiny$^{\textcolor{green!60!black}{+11.60}}$}
& \textbf{84.11}{\tiny$^{\textcolor{green!60!black}{+17.40}}$}
& \textbf{34.10}{\tiny$^{\textcolor{green!60!black}{+27.89}}$}
& \textbf{2.39}{\tiny$^{\textcolor{green!60!black}{+1.39}}$}
& \textbf{3.31}{\tiny$^{\textcolor{green!60!black}{+3.26}}$} \\

\midrule

\multirow{2}{*}{Context}
& Retrieved chunks
& 24.60 & \textbf{47.96} & \textbf{11.12} & 1.74 & 0.96
& \textbf{35.50} & \textbf{84.10} & 34.10 & 2.40 & 3.30 \\

& Full paper text
& 24.10{\tiny$^{\textcolor{red!70!black}{-0.50}}$}
& 43.26{\tiny$^{\textcolor{red!70!black}{-4.70}}$}
& 10.80{\tiny$^{\textcolor{red!70!black}{-0.32}}$}
& \textbf{1.77}
& \textbf{1.00}{\tiny$^{\textcolor{green!60!black}{+0.04}}$}
& 34.67{\tiny$^{\textcolor{red!70!black}{-0.83}}$}
& 82.13{\tiny$^{\textcolor{red!70!black}{-1.97}}$}
& \textbf{35.19}{\tiny$^{\textcolor{green!60!black}{+1.09}}$}
& \textbf{2.51}{\tiny$^{\textcolor{green!60!black}{+0.11}}$}
& \textbf{3.37}{\tiny$^{\textcolor{green!60!black}{+0.07}}$} \\

\bottomrule
\end{tabular}%
}

\caption{
Ablation studies on data construction and training context.
\textbf{Data} compares raw-review and rebuttal-enhanced training.
\textbf{Context} compares retrieved-chunk and full-paper training; both are evaluated with full-paper input.
Deltas are relative to the first variant in each block.
}
\label{tab:data_context_ablation}
\end{table*}

\subsection{Main Results}
\paragraph{Automatic and reference-based results.}
Table~\ref{tab:auto} shows that \ours-RL improves over \ours-SFT on most reference-based and task-specific metrics. 
For Task~1, it improves ROUGE-L, BLEU, semantic similarity, specificity, and rubric score. 
It also reduces the over-generation observed in zero-shot baselines, producing 1.24 claims per instance, which is closer to the reference average of 1.37 claims per instance than baselines that often generate two or more claims. 
We interpret claim count as a calibration signal rather than a standalone quality measure: overly many claims may indicate unfocused weakness discovery, while overly few claims may under-cover valid paper weaknesses. 
For Task~2, \ours-RL achieves the strongest trained-model results on ROUGE-L, BLEU, suggestion similarity, evidence specificity, and rubric score, suggesting better alignment with reference revision guidance and paper-specific evidence.

\paragraph{Pairwise LLM and human evaluation.}
Table~\ref{tab:joint_pairwise_eval} 
reports GPT-5.4 and human pairwise comparisons under the same adjusted-win-rate scale. 
Both signals show that \ours-RL is clearly preferred over prior specialized review-generation baselines, especially on Task~2 Actionability and Overall quality. 
Against GPT-5.1, Gemini, and their schema-controlled variants, preferences are closer to parity, indicating that output structure explains part of the gap. 
Technical Accuracy is also the most competitive dimension for frontier LLMs, so our results should not be read as universal technical superiority. 
Rather, the main gains are concentrated on revision-oriented dimensions: more specific diagnostic claims, deeper constructive feedback, and more actionable suggestions. 
The agreement between human and GPT-5.4 trends supports using GPT-5.4 as a
scalable proxy for aggregate comparison, while human evaluation remains the
higher-trust validation signal. An independent Claude evaluation yields
similar aggregate trends and instance-level agreement with humans
(Appendix~\ref{app:cross_judge}).

\subsection{Ablation Study}

\begin{table}[t]
\centering
\small
\captionsetup{skip=4pt}
\setlength{\tabcolsep}{3.2pt}
\renewcommand{\arraystretch}{1.08}
\begin{tabular}{lcccc}
\toprule
\textbf{Reward Setting}
& \textbf{T1 Tech.}
& \textbf{T1 Overall}
& \textbf{T2 Action.}
& \textbf{T2 Overall} \\
\midrule
SFT only                         & 62.1 & 59.4 & 65.2 & 62.8 \\
GRPO w/ fixed task-level rubric  & 60.0 & 60.7 & 64.1 & 61.0 \\
GRPO w/ direct judge reward      & 65.8 & 65.1 & 72.4 & 68.7 \\
GRPO w/ candidate-aware rubric   & 68.5 & 69.3 & 77.6 & 73.4 \\
\bottomrule
\end{tabular}
\caption{
Reward design ablation. We compare SFT only with GRPO variants using increasingly structured reward signals: direct LLM-judge reward without rubrics, a fixed task-level rubric shared across instances, and our candidate-aware rubric. Scores are GPT-5.4 judge scores on \bench.
}
\label{tab:reward_ablation}
\end{table}

\begin{table}[t]
\centering
\small
\captionsetup{skip=4pt}
\setlength{\tabcolsep}{3.0pt}
\renewcommand{\arraystretch}{1.08}
\begin{tabular}{lccccc}
\toprule
\textbf{Formulation}
& \textbf{T1 Spec./Gnd.}
& \textbf{T1 Overall}
& \textbf{T2 Action.}
& \textbf{T2 Overall}
& \textbf{Claim--Sugg. Align.} \\
\midrule
End-to-end SFT
& 2.91
& 2.84
& 3.07
& 2.76
& 3.13 \\

Two-task SFT
& 3.15{\tiny$^{\textcolor{green!60!black}{+0.24}}$}
& 3.02{\tiny$^{\textcolor{green!60!black}{+0.18}}$}
& 3.31{\tiny$^{\textcolor{green!60!black}{+0.24}}$}
& 3.18{\tiny$^{\textcolor{green!60!black}{+0.42}}$}
& 3.48{\tiny$^{\textcolor{green!60!black}{+0.35}}$} \\
\bottomrule
\end{tabular}
\caption{
Task-formulation ablation comparing end-to-end and two-task SFT under the same backbone, training instances, and evaluation protocol. Scores are GPT-5.4 rubric ratings on a 0--5 scale over 200 \bench{} instances. Superscripts report improvements over end-to-end SFT.
}
\label{tab:formulation_ablation}
\end{table}

\label{sec:ablation}


Tables~\ref{tab:data_context_ablation} and
\ref{tab:reward_ablation} show that each major component contributes to the
final system. Rebuttal-enhanced training data yields particularly large gains
for Task~2, supporting the use of author responses as latent supervision for
revision planning. Localized training context remains competitive with
full-paper training on the aggregate benchmark while using shorter and more
focused inputs. However, a targeted analysis shows that full-paper training
is preferable for concerns requiring cross-section reasoning
(Appendix~\ref{app:global_context}).

Candidate-aware rubric rewards outperform both direct judge rewards and a
fixed task-level rubric, demonstrating the value of instance-specific reward
criteria. We additionally compare the two-task SFT formulation against an
end-to-end SFT variant under the same backbone, training instances, and
evaluation protocol. The two-task formulation improves Task~2 Overall by
0.42 points and Claim--Suggestion Alignment by 0.35 points, indicating that
explicitly separating diagnosis from revision planning strengthens the
connection between the identified weakness and the proposed action
(Appendix~\ref{app:formulation_ablation}).

\subsection{Generalization and Reliability}
\label{sec:generalization_reliability}

Additional evaluations support the robustness of our findings. On 175
independently annotated instances from held-out 2025--2026 papers,
\textsc{\ours-RL} remains comparable to GPT-5.1$_{\mathrm{schema}}$ in
weakness validity and technical correctness while improving resolution
sufficiency and reviewer usefulness
(Appendix~\ref{app:independent_eval}). Independent Claude and human judgments
show trends consistent with GPT-5.4
(Appendix~\ref{app:cross_judge}). On the same 175 instances, fine-grained
technical-error analysis shows \textsc{\ours-RL} has the lowest severe-error
rate (6.29\%) among all evaluated systems and an overall technical-error rate
(28.00\%) no higher than the baselines', so the gains in actionability are not
offset by more severe scientific failures
(Appendix~\ref{app:technical_errors}).
Full-paper training also improves cross-section reasoning over
localized-chunk training (mean score 1.49 vs.\ 1.17) at a small cost in
resolution sufficiency, and on 50 hard cases with an incomplete rebuttal,
\textsc{\ours-RL} extends or rejects the original resolution 66\% of the
time rather than reproducing it
(Appendices~\ref{app:global_context} and~\ref{app:hard_cases}).
\textsc{\ours-RL} also abstains correctly on 78.3\% of 360 screened
non-supported weakness--paper pairs while maintaining a 95.4\% Answer Rate
on supported instances, substantially outperforming the strongest baseline
evaluated (Appendix~\ref{app:abstention}).

\section{Conclusion}

We introduced \ours, a Review2Revise framework for actionable 
peer-review generation that separates diagnostic claim generation 
from revision suggestion generation. Using real-world 
review--rebuttal threads, we construct \data by treating rebuttals 
as latent supervision to rewrite raw reviewer weaknesses into 
structured revision-oriented feedback, and further optimize the 
model with candidate-aware weakness-specific rubric rewards. 
Experiments on \bench show that \ours consistently outperforms 
prior specialized review-generation systems on actionability and 
grounding, and remains competitive with strong prompt-based LLMs 
despite an order-of-magnitude parameter gap. Ablation studies 
confirm that rebuttal-guided enhancement and candidate-aware 
rubric rewards each contribute independently to output quality. 
More broadly, our results suggest that review--rebuttal 
interactions are a valuable and underexploited source of 
supervision, and that structurally separating weakness diagnosis 
from revision guidance produces more targeted and executable 
feedback than treating review generation as a single task.
\section*{Limitations}

Our work has three main limitations. 
First, \bench focuses on rebuttal-resolvable weaknesses, where reviewer concerns can be linked to concrete author-side revision actions. 
This enables grounded evaluation of actionable feedback, but excludes concerns such as fundamental novelty disputes, deep conceptual disagreements, or irreparable methodological flaws, which may require broader literature comparison or interactive research guidance. Second, while \ours improves revision-oriented dimensions such as grounding and actionability, it does not uniformly outperform strong proprietary LLMs on technical accuracy. This suggests that our framework is most effective at making feedback more specific and useful for revision, while deeper technical judgment remains an important direction for future work. Third, because GPT-5.4 is used for data enhancement, rubric construction, semantic reward scoring, and scalable pairwise evaluation, our pipeline may still be affected by judge-model coupling. 
We reduce this risk through offline rubric construction, disjoint evaluation instances, direct pairwise comparison under the same paper context, and human evaluation on a 200-instance subset, where human preferences largely follow the LLM-judge trends. 
Future work should further validate actionable peer-review generation with more diverse independent judges and larger-scale expert evaluation.
\bibliography{custom}

\clearpage
\appendix

\clearpage

\section{Benchmark Annotation Details}
\label{app:benchmark_details}

To ensure the reliability of \bench, we curate the benchmark through full independent human annotation with explicit quality-control criteria. The goal is to retain review--rebuttal instances in which author responses provide grounded reference signals for evaluating actionable peer-review generation. Importantly, rebuttal-derived author actions are not treated as unique gold answers. They are used as realistic examples of how reviewer concerns may be addressed through concrete revisions.

\paragraph{Annotation Objective.}
Benchmark curation aims to identify instances where a reviewer concern can be linked to a concrete author-side revision action. Such instances allow us to evaluate whether generated feedback is not only diagnostically relevant, but also practically useful for revision. Because multiple valid revision paths may exist for the same reviewer concern, the marked rebuttal action is treated as a grounded reference signal rather than the only correct answer.

\paragraph{Selection Criteria.}
Each candidate review--rebuttal instance is evaluated according to three criteria. These criteria are operationalized as annotator questions in the independent annotation protocol below.

\begin{itemize}[leftmargin=*]
    \item \textbf{Relevant and Substantive Response.}
    The rebuttal must directly engage with the reviewer-raised concern and provide a meaningful response. Instances are filtered out if the rebuttal shifts to unrelated claims, only repeats high-level justification, or mainly provides defensive argumentation without addressing the core issue.

    \item \textbf{Actionability and Specificity.}
    The rebuttal must contain a concrete revision-oriented action. Examples include adding an experiment, performing an ablation, clarifying assumptions, revising a method description, expanding limitations, modifying implementation details, or narrowing an empirical claim. Generic promises such as ``we will improve the discussion'' or ``we will clarify this point'' are not sufficient unless accompanied by a specific revision plan.

    \item \textbf{Discussion-Level Support.}
    When follow-up reviewer discussion is available, it is used as additional evidence for whether the rebuttal remains aligned with the original concern. Instances are preferred when the reviewer acknowledges that the proposed response or revision would address the concern. Cases are filtered out when the follow-up indicates that the main issue remains unresolved or that the rebuttal has drifted away from the original point.
\end{itemize}

\paragraph{Independent Annotation Protocol.}
We begin with a pool of candidate review--rebuttal instances sampled from automatically aligned weakness--response pairs. Each candidate instance is independently annotated by two annotators. Annotators are shown the reviewer concern, the candidate rebuttal span, and any available follow-up reviewer discussion, but they do not see each other's labels.

For each instance, annotators provide three criterion-level judgments and one final keep/filter decision. Specifically, they judge whether the rebuttal span:
(1) directly and substantively addresses the reviewer concern,
(2) contains a concrete revision-oriented action, and
(3) is supported by follow-up reviewer discussion when such discussion is available.
They then decide whether the instance should be retained in \bench under these criteria. Annotators also mark the rebuttal span that captures the author-side revision action.

We use a three-point scale for the criterion-level judgments: \texttt{2} for clear yes, \texttt{1} for partial or unclear, and \texttt{0} for no. For discussion-level support, annotators may also select \texttt{N/A} when no follow-up discussion is available. The intermediate label \texttt{1} is used to identify borderline cases during adjudication.

For benchmark retention, we apply a conservative rule: an instance is retained only if it directly and substantively addresses the reviewer concern (\texttt{Q1=2}), contains a concrete revision-oriented action (\texttt{Q2=2}), and has no follow-up evidence indicating that the concern remains unresolved (\texttt{Q3=2} or \texttt{N/A}). Instances with Q3=0 are always filtered. Instances with \texttt{Q1=1} or \texttt{Q2=1} are treated as borderline and are filtered by default unless adjudication determines that the revision signal is sufficiently specific and reliable.  Disagreements on the final keep/filter decision or on the relevant span are resolved by a third adjudicator. The final benchmark contains instances retained after adjudication.

\paragraph{Validation Metrics.}
We report human validation statistics for the benchmark filtering process in Table~\ref{tab:human_validation}. Precision, recall, and F1 are computed by comparing each annotator's initial keep/filter decisions against the adjudicated labels, and then averaging across annotators. Precision measures how often instances labeled as retained by an annotator are retained after adjudication. Recall measures how many adjudicated-retained instances are recovered by the annotator. F1 summarizes the two. We also report Cohen's $\kappa$ between the two independent annotators before adjudication.

\begin{figure*}[t]
    \centering
    \includegraphics[width=\textwidth]{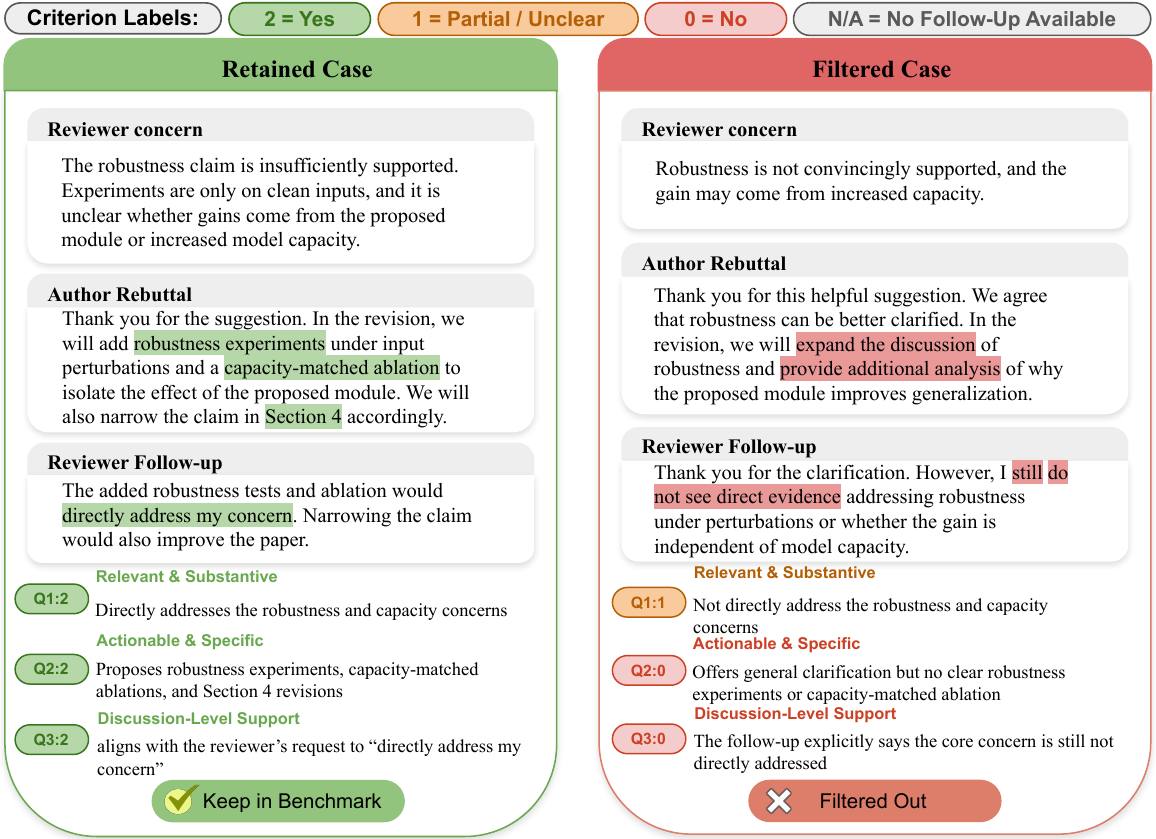}
    \caption{Illustration of benchmark filtering criteria.}
    \label{fig:benchmark_filter_example}
\end{figure*}

\paragraph{Retained vs.\ Filtered Cases.}
Figure~\ref{fig:benchmark_filter_example} shows representative retained and filtered cases under the annotation protocol. The retained example receives clear positive labels on all three criteria: the rebuttal directly addresses the reviewer concern (\texttt{Q1=2}), proposes concrete revisions including robustness experiments and a capacity-matched ablation (\texttt{Q2=2}), and is supported by reviewer follow-up (\texttt{Q3=2}). It is therefore kept in \bench. In contrast, the filtered example is only partially related to the reviewer concern (\texttt{Q1=1}), does not propose a concrete revision action (\texttt{Q2=0}), and is contradicted by reviewer follow-up indicating that the core concern remains unresolved (\texttt{Q3=0}). This distinction illustrates why \bench retains only cases where the rebuttal provides a specific and reliable revision signal, rather than merely fluent or persuasive author-response text.


\paragraph{Scope of the Benchmark.}
Because the benchmark requires a concrete author-side revision action, \bench focuses on \emph{rebuttal-resolvable} weaknesses: cases where reviewer concerns can be linked to plausible revision paths in the author response. This design supports evaluation of revision-oriented feedback, but it does not cover the full space of peer-review concerns. For example, unresolved novelty disputes, fundamental conceptual disagreements, or cases where reviewers and authors do not converge on a revision path are outside the main scope of this benchmark.

\begin{table*}[t]
\centering
\small
\setlength{\tabcolsep}{5pt}
\renewcommand{\arraystretch}{1.15}
\begin{tabularx}{\textwidth}{p{2.2cm} X}
\toprule
\textbf{Component} & \textbf{Case Study Content} \\
\midrule

\textbf{Input} &
\textbf{Paper:} \emph{Advancing Nearest Neighbor Explanation-by-Example with Critical Classification Regions} \\
& \textbf{Weakness observation:} The twin-system underlying the proposed method is not properly presented or illustrated, making it impossible to fully understand the paper without reading prior work describing the twin-system. The proposed Critical Classification Regions (CCR) method is not adequately discussed, and the paper does not explain how the method identifies critical regions or how these regions contribute to explanations.\\
& \textbf{Weakness label:} \emph{Unclear or Incomplete Method Description} \\
& \textbf{Metadata:} \textit{Abstract:} There is growing evidence that post-hoc explanation-by-example with nearest neighbors is a promising approach to XAI. \ldots\ A controlled user study on ImageNet ($N$=163) shows that CCRs improve people’s assessments of CNN predictions for difficult classifications caused by ambiguity. \textit{Keywords:} Explainable AI; Post-hoc Nearest Neighbor Explanation-by-Example; User Study; Case-based Reasoning; Convolutional Neural Network. \\
&  \textbf{Retrieval Chunks: [optional]}  \\
\midrule
\textbf{Task 1} &
\textbf{Claim generation 1:} The paper does not adequately explain the twin‑system architecture that underlies the proposed method, leaving readers unable to fully understand or reproduce the approach without consulting external prior work. \\&
\textbf{Claim generation 2: } The paper does not clearly articulate how Critical Classification Regions (CCRs) are identified or how these regions concretely contribute to the explanation-by-example mechanism. \\
\midrule
\textbf{Task 2} &
\textbf{Actionable suggestion for claim 1:} \\
& \textbf{What:} Provide a complete, standalone description of the twin‑system architecture, including its components, data flow, and training objective. \\
& \textbf{Where:} A dedicated subsection within the method section (e.g., immediately after the introduction of CCRs) or a clearly labeled appendix dedicated to architectural details. \\
& \textbf{How:} Add a concise but thorough textual explanation covering: (1) the architecture of the twin networks, (2) how inputs are processed in parallel, (3) how similarity or correspondence between image regions is computed, (4) how this structure integrates with the CCR extraction pipeline, and (5) any hyperparameters or constraints required for reproducing the system. Supplement the description with a schematic diagram illustrating the two network branches, the information passed between them, and their relation to the nearest‑neighbor retrieval process.\\
& \textbf{Expected Outcome:} Readers will be able to understand the twin‑system without consulting external work, ensuring the method is self‑contained and reproducible. The added clarity should also make it easier to evaluate whether CCRs genuinely depend on the architectural choices described. \\

\bottomrule
\end{tabularx}
\caption{Case study illustrating the two-task formulation of \textsc{ActionReview}. Task 1 maps paper context and a weakness label to a diagnostic claim, while Task 2 maps the same context and the generated claim to structured actionable guidance.}
\label{tab:task_definition_case}
\end{table*}

\section{Data Collection and Licensing}
\label{app:data_collection}

This appendix provides additional details on the source distribution, licensing, and example format of \textsc{ActReview-40K}. The dataset is constructed from publicly available review--rebuttal discussions and paper metadata, and is used to build structured supervision for diagnostic claim generation and actionable suggestion generation.

\subsection{Source Distribution}
\label{app:source_distribution}

Table~\ref{tab:source_distribution} reports the distribution of collected papers across venues and years. 
We first collect raw review--rebuttal threads from OpenReview and then filter for papers with usable reviewer concerns and author responses. 
\textbf{Raw} denotes all collected papers, \textbf{Act.} denotes papers retained after filtering for actionable review--rebuttal interactions, and \textbf{Ret.} denotes the corresponding retention rate.

\begin{table}[H]
\centering
\small
\setlength{\tabcolsep}{3pt}
\renewcommand{\arraystretch}{1.08}
\begin{tabular}{llrrr}
\toprule
\textbf{Venue} & \textbf{Year} & \textbf{Raw} & \textbf{Act.} & \textbf{Ret.} \\
\midrule
ICLR    & 2021 & 2,594 & 2,092 & 80.6\% \\
ICLR    & 2022 & 2,617 & 2,081 & 79.5\% \\
ICLR    & 2023 & 3,792 & 3,006 & 79.3\% \\
NeurIPS & 2021 & 2,768 & 2,004 & 72.4\% \\
NeurIPS & 2022 & 2,824 & 2,308 & 81.7\% \\
NeurIPS & 2023 & 3,395 & 2,844 & 83.8\% \\
EMNLP   & 2023 & 2,020 & 1,484 & 73.5\% \\
\midrule
\textbf{Total} & -- & 20,010 & 15,819 & 79.1\% \\
\bottomrule
\end{tabular}
\caption{
Source-paper distribution for \textsc{ActReview-40K}. 
\textbf{Raw} denotes all collected papers, \textbf{Act.} denotes papers retained after filtering for actionable review--rebuttal interactions, and \textbf{Ret.} denotes the retention rate.
}
\label{tab:source_distribution}
\end{table}

\subsection{Data Licensing and Attribution}
\label{app:data_licensing}

\textsc{ActReview-40K} is constructed from publicly available peer-review and paper sources. 
Review comments, rebuttals, discussion threads, and metadata are collected from OpenReview\footnote{\url{https://openreview.net}}, which distributes content under the Creative Commons Attribution 4.0 International (CC BY 4.0) license. 
Paper contents are accessed from publicly available repositories such as arXiv, where licenses vary across individual papers.

We do not claim ownership of the original papers, reviews, rebuttals, or discussion content. 
Paper content is used only for structured extraction, localized retrieval, and context construction. 
We preserve attribution to the original authors and sources, and use all data in accordance with the corresponding licenses. 
The released dataset will include derived structured fields and metadata needed for reproducibility, while respecting the licensing constraints of the original sources.
\subsection{Privacy and anonymization.}
Our source data are publicly available review--rebuttal threads from OpenReview. 
Before release, we remove or avoid distributing personally identifying information beyond what is already publicly available, and we do not release non-public reviewer identities or private metadata. 
The benchmark is distributed for research use, and users should not attempt to deanonymize reviewers, authors, or papers beyond the information made public by the source platform.
\subsection{Real vs. Enhanced Data Example}
\label{app:real_vs_enhanced}

Figure~\ref{fig:real_vs_enhanced_example} illustrates how a raw review--rebuttal instance is transformed into structured supervision. 
The real data consists of a reviewer weakness and the corresponding author rebuttal. 
The enhanced instance rewrites this signal into initial-review-style feedback: a diagnostic claim, evidence, and actionable revision guidance. 
This conversion makes the supervision more suitable for learning revision-oriented feedback, while avoiding direct imitation of the rebuttal text.
\begin{figure}[!t]
\centering
\captionsetup{justification=centering}
\scriptsize

\begin{tcolorbox}[
    enhanced,
    width=\linewidth,
    colback=blue!4,
    colframe=blue!55!black,
    title=\textbf{Real Data},
    fonttitle=\bfseries,
    boxrule=0.5pt,
    arc=1.2mm,
    left=0.6mm,right=0.6mm,top=0.6mm,bottom=0.6mm
]
\textbf{Consolidated Weakness}
\begin{tcolorbox}[colback=white,colframe=blue!30,boxrule=0.3pt,arc=0.8mm,
left=0.6mm,right=0.6mm,top=0.5mm,bottom=0.5mm]
The paper insufficiently explains why the proposed method improves self-supervised adversarial robustness compared to existing methods, and provides no clear justification for why SIR and AIR should be used simultaneously despite AIR appearing to help both clean performance and robustness.
\end{tcolorbox}

\vspace{0.15em}

\textbf{Rebuttal}
\begin{tcolorbox}[colback=white,colframe=blue!30,boxrule=0.3pt,arc=0.8mm,
left=0.6mm,right=0.6mm,top=0.5mm,bottom=0.5mm]
AIR is a simple plug-in that can help all existing ACL methods. The embedding should be exempted from nuisance style factors for better transferability. AIR regulates robust representations, while SIR regulates standard representations. Incorporating both can improve natural generalization and adversarial robustness. Experiments in the main paper and Appendix C.5 validate AIR on ACL, DynCAL, AdvCL, and A-InfoNCE.
\end{tcolorbox}
\end{tcolorbox}

\vspace{0.45em}

\begin{tcolorbox}[
    enhanced,
    width=\linewidth,
    colback=green!4,
    colframe=green!45!black,
    title=\textbf{Enhanced Data},
    fonttitle=\bfseries,
    boxrule=0.5pt,
    arc=1.2mm,
    left=0.6mm,right=0.6mm,top=0.6mm,bottom=0.6mm
]
\textbf{Claim}
\begin{tcolorbox}[colback=white,colframe=green!30,boxrule=0.3pt,arc=0.8mm,
left=0.6mm,right=0.6mm,top=0.5mm,bottom=0.5mm]
The paper does not clearly articulate or empirically validate why combining SIR and AIR improves adversarial robustness and clean performance compared to using either regularizer alone.
\end{tcolorbox}

\vspace{0.15em}

\textbf{Evidence}
\begin{tcolorbox}[colback=white,colframe=green!30,boxrule=0.3pt,arc=0.8mm,
left=0.6mm,right=0.6mm,top=0.5mm,bottom=0.5mm]
The methodology introduces SIR and AIR but does not explain their complementary roles, and the experiments do not isolate their individual and joint contributions. Thus, the reader cannot determine whether simultaneous use is necessary or incidental.
\end{tcolorbox}
\end{tcolorbox}

\caption{
\textbf{Comparison between real and enhanced training instances.}
Enhanced data decomposes a real review--rebuttal example into a claim--evidence pair.
}
\label{fig:real_vs_enhanced_example}
\end{figure}

\begin{figure}[!t]
\ContinuedFloat
\centering
\captionsetup{justification=centering}
\scriptsize

\begin{tcolorbox}[
    enhanced,
    width=\linewidth,
    colback=green!4,
    colframe=green!45!black,
    title=\textbf{Enhanced Data (continued)},
    fonttitle=\bfseries,
    boxrule=0.5pt,
    arc=1.2mm,
    left=0.6mm,right=0.6mm,top=0.6mm,bottom=0.6mm
]
\textbf{Actionable Suggestions}
\begin{tcolorbox}[colback=white,colframe=green!30,boxrule=0.3pt,arc=0.8mm,
left=0.6mm,right=0.6mm,top=0.5mm,bottom=0.5mm]

\textbf{Suggestion 1.}
\begin{itemize}[leftmargin=1.1em,itemsep=0.05em,topsep=0.05em]
    \item \textbf{What:} Add an ablation separating SIR-only, AIR-only, and SIR+AIR.
    \item \textbf{Where:} Experiments section.
    \item \textbf{How:} Compare clean and robust accuracy under identical settings.
\end{itemize}

\vspace{0.1em}

\textbf{Suggestion 2.}
\begin{itemize}[leftmargin=1.1em,itemsep=0.05em,topsep=0.05em]
    \item \textbf{What:} Explain the distinct roles of SIR and AIR.
    \item \textbf{Where:} End of Methodology.
    \item \textbf{How:} Clarify why the two regularizers are complementary rather than redundant.
\end{itemize}

\vspace{0.1em}

\textbf{Suggestion 3.}
\begin{itemize}[leftmargin=1.1em,itemsep=0.05em,topsep=0.05em]
    \item \textbf{What:} Evaluate AIR as a plug-in on multiple ACL variants.
    \item \textbf{Where:} Additional experiments table.
    \item \textbf{How:} Report performance with and without AIR under the same setup.
\end{itemize}

\end{tcolorbox}
\end{tcolorbox}

\caption{
\textbf{Comparison between real and enhanced training instances (continued).}
Enhanced data further specifies \textit{what} to improve, \textit{where} the revision should be made, and \textit{how} to implement it.
}
\end{figure}
\section{Rebuttal-Guided Data Construction}
\label{app:data_construction}
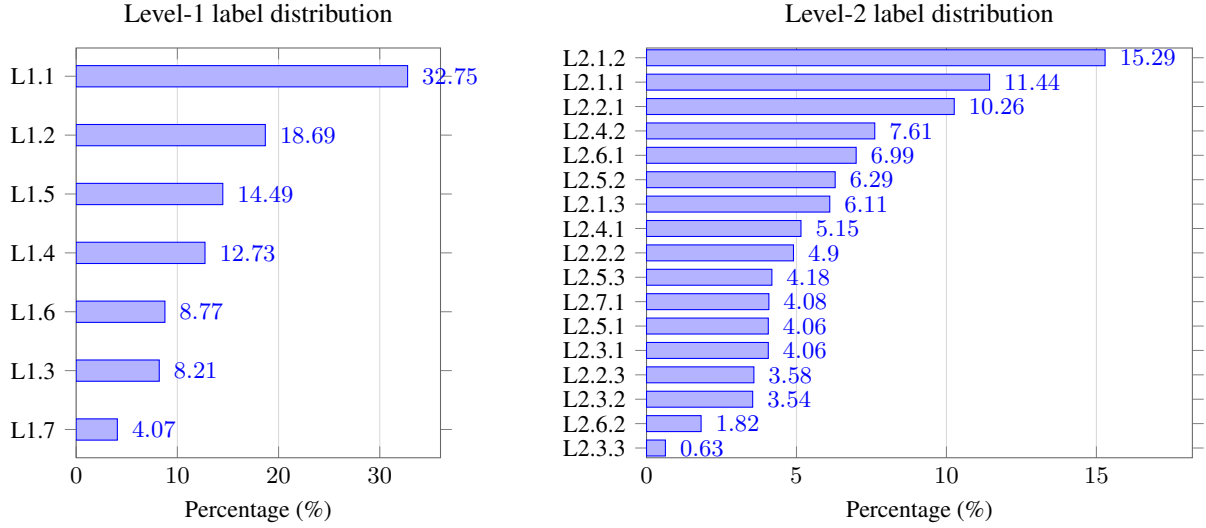
\begin{figure*}[t]
\centering

\begin{minipage}[t]{0.40\textwidth}
\centering
\begin{tikzpicture}
\begin{axis}[
    xbar,
    width=\linewidth,
    height=7.0cm,
    xmin=0, xmax=36,
    bar width=8pt,
    symbolic y coords={L1.7,L1.3,L1.6,L1.4,L1.5,L1.2,L1.1},
    ytick=data,
    xlabel={Percentage (\%)},
    ylabel={},
    nodes near coords,
    every node near coord/.append style={
        font=\small,
        anchor=west,
        xshift=2pt
    },
    point meta=x,
    nodes near coords={\pgfmathprintnumber[fixed,precision=2]{\pgfplotspointmeta}},
    enlarge y limits=0.08,
    xmajorgrids,
    ymajorgrids=false,
    grid style={gray!30},
    title={Level-1 label distribution},
    tick label style={font=\small},
    label style={font=\small},
    title style={font=\normalsize},
    axis line style={black!60},
]
\addplot coordinates {
    (4.07,L1.7)
    (8.21,L1.3)
    (8.77,L1.6)
    (12.73,L1.4)
    (14.49,L1.5)
    (18.69,L1.2)
    (32.75,L1.1)
};
\end{axis}
\end{tikzpicture}
\end{minipage}
\hfill
\begin{minipage}[t]{0.55\textwidth}
\centering
\begin{tikzpicture}
\begin{axis}[
    xbar,
    width=\linewidth,
    height=7.0cm,
    xmin=0, xmax=18.2,
    bar width=6.0pt,
    symbolic y coords={
        L2.3.3,
        L2.6.2,
        L2.3.2,
        L2.2.3,
        L2.3.1,
        L2.5.1,
        L2.7.1,
        L2.5.3,
        L2.2.2,
        L2.4.1,
        L2.1.3,
        L2.5.2,
        L2.6.1,
        L2.4.2,
        L2.2.1,
        L2.1.1,
        L2.1.2
    },
    ytick=data,
    xlabel={Percentage (\%)},
    ylabel={},
    nodes near coords,
    every node near coord/.append style={
        font=\small,
        anchor=west,
        xshift=2pt
    },
    point meta=x,
    nodes near coords={\pgfmathprintnumber[fixed,precision=2]{\pgfplotspointmeta}},
    enlarge y limits=0.025,
    xmajorgrids,
    ymajorgrids=false,
    grid style={gray!30},
    title={Level-2 label distribution},
    tick label style={font=\small},
    label style={font=\small},
    title style={font=\normalsize},
    axis line style={black!60},
]
\addplot coordinates {
    (0.63,L2.3.3)
    (1.82,L2.6.2)
    (3.54,L2.3.2)
    (3.58,L2.2.3)
    (4.06,L2.3.1)
    (4.06,L2.5.1)
    (4.08,L2.7.1)
    (4.18,L2.5.3)
    (4.90,L2.2.2)
    (5.15,L2.4.1)
    (6.11,L2.1.3)
    (6.29,L2.5.2)
    (6.99,L2.6.1)
    (7.61,L2.4.2)
    (10.26,L2.2.1)
    (11.44,L2.1.1)
    (15.29,L2.1.2)
};
\end{axis}
\end{tikzpicture}
\end{minipage}

\caption{Normalized distributions of Level-1 and Level-2 weakness labels.}
\label{fig:taxonomy_distribution}
\end{figure*}
This appendix provides additional details for the construction of \data. 
We describe the weakness taxonomy, weakness extraction and alignment prompts, alignment validation, rebuttal-guided enhancement, enhancement quality control, and localized evidence retrieval.

\subsection{Data-Driven Weakness Taxonomy}
\label{app:taxonomy}

We construct a two-level weakness taxonomy through a data-driven, human-in-the-loop process. 
We first sample approximately 500 weakness statements from peer reviews across venues and years, and use iterative LLM-assisted pattern discovery to induce candidate fine-grained categories with definitions, keywords, and representative examples. 
The candidate categories are then consolidated through semantic merging and human review. 
Annotators decide whether proposed category groups should be merged, partially merged, or kept separate. 
The resulting Level-2 categories are organized into 7 higher-level themes under constraints of broad coverage, non-overlapping primary assignment, and balanced granularity.

We validate the taxonomy on a stratified sample of 100 weakness statements, each independently labeled by three annotators. 
Gold labels are obtained by majority vote. 
The taxonomy achieves substantial agreement, with Fleiss' $\kappa=0.79$ at Level~2 and $\kappa=0.88$ at Level~1, suggesting that the categories are interpretable and consistently applicable.

The final taxonomy contains 7 Level-1 categories and 17 Level-2 categories. 
Each category includes a definition, representative keywords, frequency statistics, and example weaknesses. 
Figure~\ref{fig:taxonomy_distribution} shows the normalized label distributions, and the full taxonomy is reported in Tables~\ref{tab:weakness_taxonomy_part1}--\ref{tab:weakness_taxonomy_part3}.


\begin{table*}[t]
\centering
\footnotesize
\setlength{\tabcolsep}{4pt}
\renewcommand{\arraystretch}{1.12}

\begin{tabular}{>{\centering\arraybackslash}m{2.6cm}
                >{\raggedright\arraybackslash}p{3.3cm}
                >{\raggedright\arraybackslash}p{9.0cm}}
\toprule
\textbf{Level 1 Category} & \textbf{Level 2 Category} & \textbf{Definition} \\
\midrule

\multirow[c]{3}{=}{\makecell[c]{Experimental Design\\and Empirical\\Validation\\Weaknesses}}
& Insufficient or Narrow Experimental Evaluation
& The experimental evaluation is too limited, simplistic, or narrow in scope to support the paper's claims. This includes reliance on too few datasets, tasks, domains, or problem settings; overly simplistic or toy benchmarks; weak evidence of generalization; and experimental setups that fail to convincingly demonstrate robustness, scalability, or real-world relevance. \\
\cdashline{2-3}

& Missing or Inadequate Comparative and Component Analysis
& The paper fails to adequately justify performance claims due to missing or weak baselines, absent or insufficient ablation studies, and lack of analysis isolating the contributions of individual components, hyperparameters, or design choices. As a result, it is unclear what drives improvements or how the method compares to existing work. \\
\cdashline{2-3}

& Weak, Unreliable, or Flawed Empirical Evidence
& The reported empirical results are unconvincing, unreliable, or methodologically flawed. This includes marginal or inconsistent gains, lack of statistical rigor, inappropriate or unjustified evaluation metrics, unfair or invalid comparisons, and contradictory or unexplained results that undermine confidence in the conclusions. \\

\midrule

\multirow[c]{3}{=}{\makecell[c]{Methodological\\Clarity and\\Reproducibility\\Issues}}
& Unclear or Incomplete Method Description
& The proposed method, algorithm, or pipeline is poorly explained, underspecified, or confusing, including missing explanations of components, training or inference procedures, equations, or interactions. This prevents reviewers from clearly understanding how the method works or assessing its correctness. \\
\cdashline{2-3}

& Missing or Insufficient Experimental and Reproducibility Details
& Critical experimental, implementation, or evaluation details are missing or insufficiently specified, making results hard to interpret or reproduce. This includes missing hyperparameters, training setups, datasets, hardware, code availability, or unexplained sensitivity to tuning choices. \\
\cdashline{2-3}

& Unclear Problem Definition, Assumptions, or Scope
& The paper does not clearly define the problem, objectives, assumptions, or scope of applicability. Core concepts, variables, or conditions are ambiguous, making it unclear what is being solved and under what assumptions the results hold. \\

\bottomrule
\end{tabular}

\caption{A taxonomy of common paper weaknesses, organized by high-level categories and corresponding subtypes.}
\label{tab:weakness_taxonomy_part1}
\end{table*}

\begin{table*}[t]
\centering
\footnotesize
\setlength{\tabcolsep}{4pt}
\renewcommand{\arraystretch}{1.12}

\begin{tabular}{>{\centering\arraybackslash}m{2.6cm}
                >{\raggedright\arraybackslash}p{3.3cm}
                >{\raggedright\arraybackslash}p{9.0cm}}
\toprule
\textbf{Level 1 Category} & \textbf{Level 2 Category} & \textbf{Definition} \\
\midrule

\multirow[c]{3}{=}{\makecell[c]{Theoretical\\Soundness and\\Justification Gaps}}
& Missing or Insufficient Theoretical Justification
& The work lacks adequate theoretical analysis or justification for its claims, methods, or design choices. This includes missing proofs, absent guarantees, unclear or weakly motivated theoretical arguments, and a general failure to formally support correctness, convergence, or complexity claims. \\
\cdashline{2-3}

& Flawed or Unjustified Theoretical Assumptions
& Theoretical results rely on assumptions that are unrealistic, overly strong, poorly justified, or violated in practice. This includes questionable modeling choices, weak soundness due to assumption gaps, and theoretical arguments that collapse when assumptions are examined critically. \\
\cdashline{2-3}

& Theory--Practice Misalignment
& There is a clear disconnect between the theoretical analysis and the empirical evaluation, where theoretical claims are not reflected in experiments, assumptions do not hold in practice, or experimental results fail to validate the stated theory. \\

\midrule

\multirow[c]{2}{=}{\makecell[c]{Novelty, Contribution,\\and Positioning\\Limitations}}
& Insufficient Positioning and Related Work Coverage
& The paper fails to adequately situate its contributions within existing literature, including missing, incomplete, or unclear discussion of relevant prior work, weak comparisons, or poor explanation of how the approach differs from existing methods. This undermines claims of novelty and proper positioning. \\
\cdashline{2-3}

& Weak, Incremental, or Overstated Novelty
& The contribution is perceived as offering limited or unclear novelty, often being incremental, derivative, or a repackaging of known ideas. Reviewers may find the novelty unclear, trivial, or exaggerated relative to prior work, questioning whether the paper meaningfully advances the state of the art. \\

\midrule

\multirow[c]{3}{=}{\makecell[c]{Motivation, Claims,\\and Practical\\Relevance Issues}}
& Weak or Unclear Motivation and Framing
& The paper fails to clearly motivate the problem, task, or method, or provides insufficient intuition and framing for why the work matters, why design choices are made, or why the setting is meaningful or realistic. \\
\cdashline{2-3}

& Unsupported, Overstated, or Incorrect Claims
& The paper makes claims that are not adequately supported by theory or experiments, are overstated relative to the evidence, misleadingly phrased, or technically incorrect. \\
\cdashline{2-3}

& Limited Practical Relevance or Real-World Applicability
& The proposed approach is questioned for being impractical, unrealistic, or unlikely to have real-world impact due to assumptions, scalability, cost, deployment constraints, or overclaimed practical impact. \\

\bottomrule
\end{tabular}

\caption{A taxonomy of common paper weaknesses (continued), organized by high-level categories and corresponding subtypes.}
\label{tab:weakness_taxonomy_part2}
\end{table*}

\begin{table*}[t]
\centering
\footnotesize
\setlength{\tabcolsep}{4pt}
\renewcommand{\arraystretch}{1.18}

\begin{tabular}{>{\centering\arraybackslash}m{3.0cm}
                >{\raggedright\arraybackslash}p{3.6cm}
                >{\raggedright\arraybackslash}p{8.4cm}}
\toprule
\textbf{Level 1 Category} & \textbf{Level 2 Category} & \textbf{Definition} \\
\midrule

\multirow[c]{2}{=}{\makecell[c]{Writing, Presentation,\\and Communication\\Problems}}
& Unclear Writing, Organization, or Notation
& The manuscript is difficult to understand due to poor writing quality, weak organization, confusing exposition, inconsistent presentation, or unclear or overloaded notation that impedes comprehension of the ideas or methods. \\
\cdashline{2-3}

& Formatting, Figures, or Submission Issues
& Problems with formatting, templates, figures, diagrams, or visual presentation that reduce readability, violate submission guidelines, or fail to adequately support the text. \\

\midrule

\makecell[c]{Scalability, Efficiency,\\and Resource\\Considerations}
& Missing Computational Cost, Runtime, and Scalability Analysis
& The paper makes a claim, contribution, or assumption whose validity or applicability depends on computational cost, runtime, memory usage, or scalability (e.g., an efficiency claim, or a claim of real-time or large-scale applicability), but fails to adequately analyze or report the corresponding characteristics. This label does not apply to papers that make no such claim. \\[1.2ex]

\bottomrule
\end{tabular}

\caption{A taxonomy of common paper weaknesses (continued), organized by high-level categories and corresponding subtypes.}
\label{tab:weakness_taxonomy_part3}
\end{table*}
\FloatBarrier

\paragraph{Weakness taxonomy discovery prompt.}
We use the following prompt Figure ~\ref{fig:taxonomy_prompt} to induce candidate weakness categories from raw reviewer weakness statements.

\begin{figure*}[t]
\centering
\setlength{\fboxsep}{0pt}

\begin{tcolorbox}[
    colback=gray!8,
    colframe=yellow!50!black,
    boxrule=0.6pt,
    arc=1pt,
    left=8pt,
    right=8pt,
    top=8pt,
    bottom=8pt,
    width=0.95\textwidth
]
\footnotesize
\textbf{Prompt for Weakness Taxonomy Discovery}

\vspace{0.5em}
You are given:
\begin{enumerate}[leftmargin=1.5em,itemsep=0.2em,topsep=0.2em]
    \item A batch of reviewer weakness statements extracted from peer reviews,
    \item Metadata such as venue, year, and decision,
    \item Previously discovered categories, if any.
\end{enumerate}

Your task is to identify \textbf{recurring weakness patterns} and organize them into a set of \textbf{semantic categories}.

\vspace{0.6em}
\textbf{Core objective:}
Induce a set of weakness categories that are:
\begin{itemize}[leftmargin=1.5em,itemsep=0.2em,topsep=0.2em]
    \item \textbf{semantically coherent}: each category captures a distinct type of weakness,
    \item \textbf{generalizable}: applicable across papers and venues,
    \item \textbf{interpretable}: clearly understandable by humans,
    \item \textbf{non-overlapping}: categories should be distinct whenever possible.
\end{itemize}

\vspace{0.6em}
\textbf{Important constraints:}
\begin{itemize}[leftmargin=1.5em,itemsep=0.2em,topsep=0.2em]
    \item Group weaknesses based on the \textbf{underlying issue}, not surface wording.
    \item Avoid creating overly specific categories; prefer recurring general patterns.
    \item Avoid merging categories that represent \textbf{different evaluation dimensions}.
    \item Each weakness should be explainable by exactly one primary category.
\end{itemize}

\vspace{0.6em}
\textbf{Required output format:}
For each category, provide:
\begin{itemize}[leftmargin=1.5em,itemsep=0.2em,topsep=0.2em]
    \item \textbf{Category name}
    \item \textbf{Definition}: one or two sentences
    \item \textbf{Keywords}: three to six indicative terms
    \item \textbf{Representative examples}: examples from the input weaknesses
\end{itemize}

\vspace{0.6em}
\textbf{Desired properties:}
\begin{itemize}[leftmargin=1.5em,itemsep=0.2em,topsep=0.2em]
    \item Capture common failure modes in academic papers,
    \item Balance coverage and granularity,
    \item Facilitate downstream classification and evaluation.
\end{itemize}

\vspace{0.6em}
\textbf{Example behavior:}
If multiple weaknesses refer to missing ablation studies, insufficient baseline comparisons, or lack of component analysis, they should be grouped into a coherent experimental-insufficiency category rather than treated separately.
\end{tcolorbox}

\caption{
Prompt template used for data-driven weakness taxonomy discovery.
The LLM groups raw reviewer weaknesses into semantically coherent categories with definitions, keywords, and examples.
}
\label{fig:taxonomy_prompt}
\end{figure*}

\subsection{Weakness Extraction and Alignment}
\label{app:alignment_prompts}

Reviews often contain multiple concerns in a single paragraph. 
We therefore decompose each review into atomic weakness units before aligning them with author responses. 
This decomposition reduces entanglement among concerns and enables fine-grained supervision from rebuttal spans.

\paragraph{Step 1: Review segmentation.}
GPT-5.4 is prompted to decompose each full review into atomic weakness units.

\paragraph{Step 2: Weakness--rebuttal span mapping.}
Each atomic weakness is then aligned to the corresponding author rebuttal span. The prompt showns in Figure ~\ref{fig:segmentation_mapping_prompts}

\begin{figure*}[t]
\centering
\setlength{\fboxsep}{0pt}

\begin{tcolorbox}[
    colback=gray!8,
    colframe=yellow!50!black,
    boxrule=0.6pt,
    arc=1pt,
    left=8pt,
    right=8pt,
    top=8pt,
    bottom=8pt,
    width=0.95\textwidth
]
\footnotesize
\textbf{Prompt for Weakness Segmentation}

\vspace{0.5em}
You are analyzing a peer review. Segment the review into individual, independent weakness points.

\vspace{0.6em}
\textbf{Task:}
Identify all distinct weakness points, concerns, or questions raised by the reviewer.
Each point should be a complete, self-contained critique.
Number the points sequentially as Point 1, Point 2, etc.

\vspace{0.6em}
\textbf{Output format:}

\vspace{0.2em}
\noindent
Point 1: [First weakness]

\noindent
Point 2: [Second weakness]

\vspace{0.6em}
\textbf{Guidelines:}
\begin{itemize}[leftmargin=1.5em,itemsep=0.2em,topsep=0.2em]
    \item Merge closely related sub-points into one atomic weakness.
    \item Each point should be substantial and technically meaningful.
    \item Preserve the reviewer's original wording whenever possible.
    \item Prefer content from sections such as ``Weaknesses'', ``Concerns'', or ``Questions'' when present.
    \item Extract only technical concerns, questions, or requests for improvement.
    \item Do not extract attitude statements, summary statements, recommendation statements, or score announcements.
    \item Do not extract any text that references a rebuttal, revision, or author response.
\end{itemize}

\vspace{0.8em}
\hrule
\vspace{0.8em}

\textbf{Prompt for Weakness--Rebuttal Span Mapping}

\vspace{0.5em}
You are analyzing a peer-review discussion.
Map each weakness point to the specific part of the author rebuttal that directly addresses it.

\vspace{0.6em}
\textbf{Task:}
For each atomic weakness, identify the minimal rebuttal span that provides a substantive author response.
The span should directly answer, clarify, justify, or provide evidence for the corresponding weakness.

\vspace{0.6em}
\textbf{Output format:}

\vspace{0.2em}
\noindent
W1 $\rightarrow$ [Exact rebuttal text addressing W1] (Confidence: 0.95)

\noindent
W2 $\rightarrow$ No Response (Confidence: 1.00)

\vspace{0.6em}
\textbf{Critical rules:}
\begin{itemize}[leftmargin=1.5em,itemsep=0.2em,topsep=0.2em]
    \item Extract exact text from the rebuttal; do not paraphrase.
    \item Map each weakness to its own most specific rebuttal segment.
    \item Prefer the shortest span that fully addresses the weakness.
    \item If a weakness is not substantively addressed, write ``No Response''.
    \item If the rebuttal quotes or restates the reviewer's weakness before answering, extract only the authors' reply.
    \item If the extracted segment is identical or nearly identical to the weakness text, write ``No Response''.
    \item Assign a confidence score reflecting how directly the rebuttal span addresses the weakness.
\end{itemize}
\end{tcolorbox}

\caption{
Prompt templates used for atomic weakness segmentation and weakness--rebuttal span mapping.
The first prompt decomposes full reviews into self-contained weakness units, while the second prompt aligns each weakness to the minimal author-response span that addresses it.
}
\label{fig:segmentation_mapping_prompts}
\end{figure*}

\paragraph{Post-processing filters.}
We apply three post-processing filters after LLM mapping. 
First, \emph{quote-only detection} removes spans where the rebuttal mainly restates the
reviewer weakness without adding author-side content. We compute bidirectional token
overlap between the weakness and the candidate rebuttal span, and discard spans when
more than 55\% of the weakness appears in the rebuttal span and more than 45\% of the
candidate span is copied from the weakness. These thresholds were chosen conservatively
from a small development audit to remove quote-only mappings while retaining spans that
quote a concern before answering it. 
Second, \emph{cross-reference resolution} replaces outputs such as ``Same segment as
W$k$'' with the actual referenced text. 
Third, \emph{shared-preamble deduplication} discards segments already assigned to a
previous weakness, as these usually correspond to generic opening statements rather than
specific responses.

\subsection{Alignment Quality Audit}
\label{app:alignment_quality}

A key step in constructing \data is mapping each atomic reviewer weakness to the rebuttal span that addresses it. 
Since these mappings provide latent supervision for rebuttal-guided enhancement, we validate the retained weakness--rebuttal span mappings through a human audit.

\paragraph{Sampling.}
We sample 50 papers to cover different venues, years, and weakness categories. 
From these papers, we include all atomic weakness segments that pass our automatic alignment filters and confidence threshold, yielding 794 mapped review segments. 
This audit evaluates the quality of the retained filtered set rather than all possible review--rebuttal pairs.

\paragraph{Annotation protocol.}
Two trained annotators independently map each atomic weakness to the most specific rebuttal span that addresses it. 
Annotators are shown the paper metadata, the atomic weakness, the full author rebuttal, and surrounding discussion when available. 
If no rebuttal span substantively addresses the weakness, annotators mark \textsc{No Response}. 
Annotators are instructed to select the minimal span that captures the concrete author response, following the same one-to-one mapping constraint and span granularity used by our pipeline.

\paragraph{Adjudication and correctness.}
After independent annotation, disagreements are adjudicated to produce a gold mapping. 
We compare each automatically predicted rebuttal span against the adjudicated gold span. 
A predicted mapping is counted as correct if it matches the gold \textsc{No Response} label, or if the predicted and gold spans overlap with token-level intersection-over-union (IoU) of at least 0.5:
\begin{equation}
\mathrm{IoU}(s_{\mathrm{pred}}, s_{\mathrm{gold}})
=
\frac{|s_{\mathrm{pred}} \cap s_{\mathrm{gold}}|}
{|s_{\mathrm{pred}} \cup s_{\mathrm{gold}}|}.
\end{equation}

\paragraph{Metrics and discussion.}
We compute span-level precision, recall, and F1 for the automatic mappings against the adjudicated gold standard. 
Precision measures the fraction of predicted mappings that match the gold span, while recall measures the fraction of gold response mappings recovered by the pipeline. 
We also report Cohen's $\kappa$ between the two annotators before adjudication.

The retained mappings achieve high span-level accuracy, suggesting that the alignment pipeline provides reliable supervision for downstream data construction. 
Most residual errors arise from long rebuttal paragraphs that address multiple reviewer concerns, implicit responses that use different terminology from the review, or generic author responses that acknowledge a concern without specifying a concrete revision. 
These cases motivate our benchmark filtering criteria and our treatment of rebuttal-derived actions as grounded reference signals rather than unique gold outputs.

\subsection{Constructing Enhanced Instances}
\label{app:construct_enhancement}

Raw reviewer comments often identify a problem without specifying a concrete revision path. 
For example, a reviewer may state that an ablation study is insufficient, but not indicate which components should be ablated, what comparison should be added, or what claim the ablation should support. 
Author rebuttals provide useful signals for this missing information because they often describe how authors attempted to address the concern, such as adding experiments, clarifying assumptions, revising claims, or expanding limitations.

However, rebuttals cannot be used directly as training targets for reviewer feedback generation. 
They are written after the review, often adopt a defensive or retrospective tone, and may mix concrete revisions with justification, acknowledgments, or generic promises. 
We therefore treat rebuttals as \emph{latent supervision}: they reveal plausible revision actions, but the generated feedback must be written as if it were part of the original review.

\paragraph{Initial-review perspective.}
The enhancement model is explicitly instructed to write from the perspective of an initial reviewer. 
It may use the aligned rebuttal span to infer what revision would address the concern, but it must not mention the rebuttal, the author response, any revised version of the paper, or any post-submission change. 
This prevents leakage from the rebuttal into the generated feedback and ensures that the resulting target resembles proactive reviewer guidance rather than retrospective commentary.

\paragraph{Enhanced output structure.}
For each aligned weakness--rebuttal pair, we prompt GPT-5.4 to produce two structured fields. 
The first field is a \emph{diagnostic claim}, a concise reviewer-side statement identifying the paper-specific deficiency. 
The second field contains \emph{revision suggestions}, which specify what should be revised, where the revision should appear, how it can be carried out, and what outcome the revision is expected to achieve. 
The output is designed to supervise both Task~1 and Task~2.

\paragraph{Post-processing filters.}
We apply automatic filters to reduce artifacts introduced by the enhancement step. 
First, we remove outputs with \emph{rebuttal leakage}, such as references to ``the authors' response,'' ``the rebuttal,'' or ``the revised paper.'' 
Second, we filter \emph{retrospective rewrites}, where the output describes what the authors already did rather than what an initial reviewer should request. 
Third, we remove outputs that \emph{near-copy} the aligned rebuttal span, since these often preserve author-side phrasing rather than reviewer-side guidance. 
Fourth, we discard \emph{underspecified suggestions}, such as generic requests to clarify the method or add experiments without specifying what should be clarified, where the revision should appear, or how the authors could implement it. 
Instances that fail these filters are regenerated once and removed if they still fail.

\paragraph{Rebuttal-guided enhancement prompt.}
The following prompt template Figure ~\ref{fig:enhancement_prompt} is used for rebuttal-guided enhancement. 
Placeholders in braces are filled with instance-specific content.

\begin{figure*}[t]
\centering
\setlength{\fboxsep}{0pt}

\begin{tcolorbox}[
    colback=gray!8,
    colframe=yellow!50!black,
    boxrule=0.6pt,
    arc=1pt,
    left=8pt,
    right=8pt,
    top=8pt,
    bottom=8pt,
    width=0.95\textwidth
]
\footnotesize
\textbf{Prompt for Rebuttal-Guided Feedback Enhancement}

\vspace{0.5em}
You are rewriting peer-review feedback into structured initial-review-style feedback.

\vspace{0.6em}
\textbf{Input:}
You are given:
\begin{enumerate}[leftmargin=1.5em,itemsep=0.2em,topsep=0.2em]
    \item Paper metadata, including title, abstract, and keywords.
    \item The target weakness category.
    \item The original reviewer weakness.
    \item The aligned author rebuttal span.
    \item Relevant paper context chunks.
\end{enumerate}

\vspace{0.6em}
\textbf{Goal:}
Write feedback as if it appeared in the original review, before the author response was written.

Use the rebuttal span only as background evidence for what revision action may address the weakness.
Do not mention the rebuttal, the author response, ``the authors later added'', ``in the revised paper'', or any post-submission change.

\vspace{0.6em}
\textbf{Required output fields:}

\begin{itemize}[leftmargin=1.5em,itemsep=0.2em,topsep=0.2em]
    \item \textbf{Diagnostic Claim}: One sentence identifying the core paper-specific deficiency.
    \item \textbf{Actionable Suggestions}: Concrete revision guidance that specifies:
    \begin{itemize}[leftmargin=1.5em,itemsep=0.15em,topsep=0.15em]
        \item \textbf{What} should be revised or added.
        \item \textbf{Where} the revision should appear, such as a section, experiment, figure, table, claim, or method description.
        \item \textbf{How} the authors can implement the revision.
        \item \textbf{Expected Outcome}: what the revision would clarify, validate, or improve.
    \end{itemize}
\end{itemize}

\vspace{0.6em}
\textbf{Quality requirements:}
\begin{itemize}[leftmargin=1.5em,itemsep=0.2em,topsep=0.2em]
    \item The feedback should be specific and grounded in the provided paper context.
    \item The suggestions should be useful for revision rather than merely descriptive.
    \item Avoid generic advice such as ``add more details'' unless it is accompanied by concrete implementation guidance.
    \item Do not introduce unsupported claims that are not grounded in the weakness, rebuttal span, or paper context.
    \item Return only the structured feedback.
\end{itemize}
\end{tcolorbox}

\caption{
Prompt template used for rebuttal-guided feedback enhancement.
The prompt converts an original reviewer weakness and its aligned rebuttal span into initial-review-style diagnostic claims and actionable revision suggestions.
}
\label{fig:enhancement_prompt}
\end{figure*}

\begin{figure*}[t]
\centering
\includegraphics[width=0.95\textwidth]{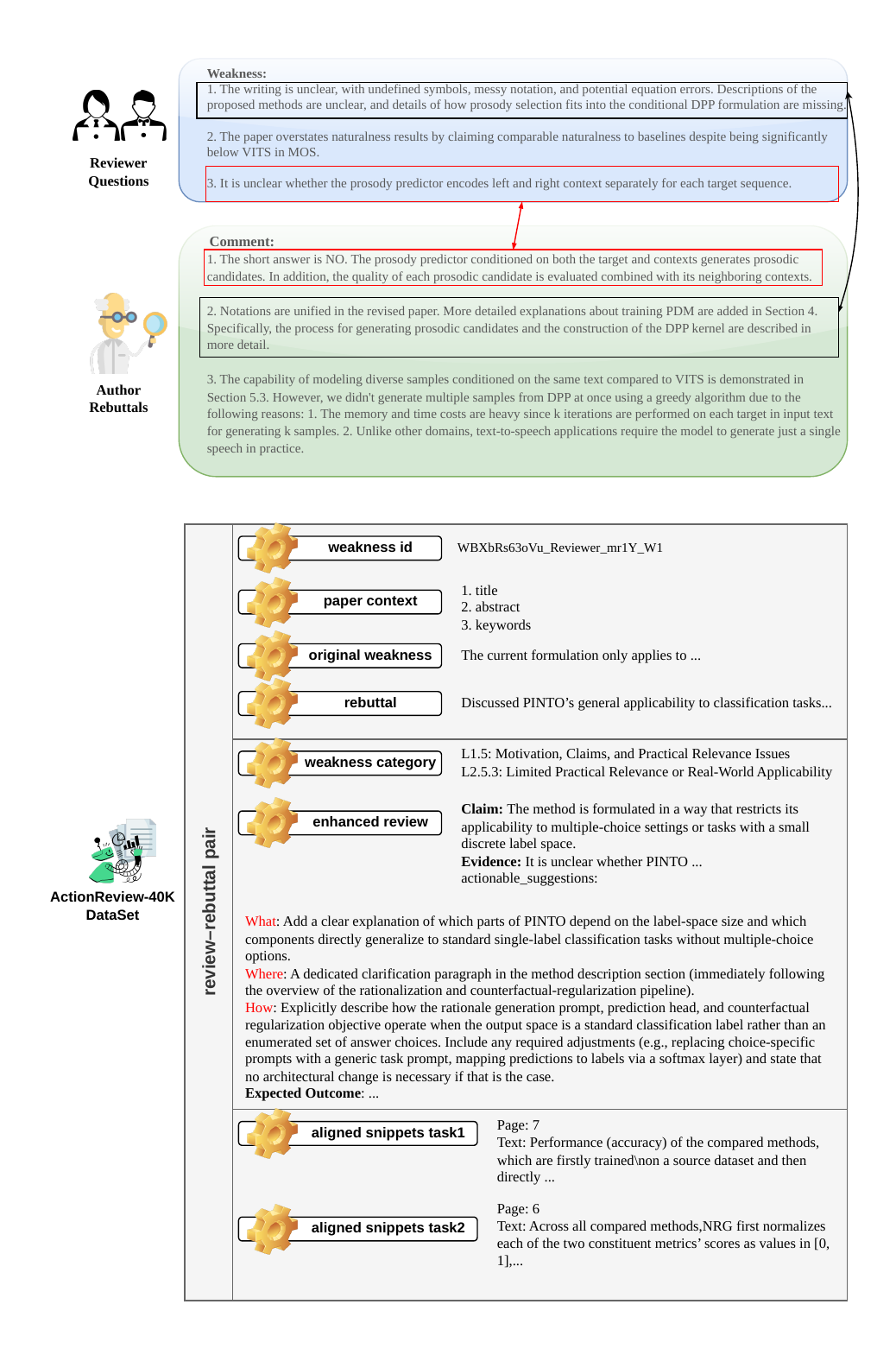}
\caption{
Example of a constructed review--rebuttal instance in \data.
The instance includes the original weakness, aligned rebuttal signal, taxonomy label, enhanced diagnostic claim and revision suggestions, and retrieved paper chunks.
}
\label{fig:data_example}
\end{figure*}

\subsection{Enhancement Quality Audit}
\label{app:enhancement_quality}

We conduct a human audit to evaluate whether rebuttal-guided enhancement produces reliable training targets beyond satisfying automatic format checks. 
The audit examines whether each enhanced instance preserves the original reviewer concern, uses the aligned rebuttal span as valid revision evidence, grounds the suggested revision in a plausible paper location, and avoids leaking post-rebuttal information into reviewer-side feedback.

\paragraph{Sampling.}
We sample 500 enhanced instances from the training-data construction pipeline, stratified by venue, year, and Level-2 weakness category. 
For each instance, annotators are shown the atomic reviewer weakness, weakness label, aligned rebuttal span, available paper context, and enhanced output. 
Each enhanced output contains a diagnostic claim and structured revision suggestions with \texttt{what}, \texttt{where}, \texttt{how}, and \texttt{expected\_outcome} fields.

\paragraph{Annotation criteria.}
Each instance is evaluated along four dimensions:
\begin{itemize}[leftmargin=*]
    \item \textbf{Concern preservation}: the diagnostic claim preserves the core meaning of the original reviewer weakness without introducing a different or substantially stronger concern.
    \item \textbf{Rebuttal-supported revision guidance}: the suggested revision is supported by the aligned rebuttal span, such as adding an experiment, clarifying an assumption, narrowing a claim, revising a method description, or reporting missing details.
    \item \textbf{Location validity}: the \texttt{where} field identifies a plausible and concrete revision location, preferably matching an observed section title or close lexical variant when section information is available.
    \item \textbf{No rebuttal leakage}: the output is written from the perspective of an initial reviewer and does not mention the rebuttal, author response, revised paper, or post-submission changes.
\end{itemize}

\paragraph{Annotation protocol and metrics.}
Two annotators independently score each criterion on a three-point scale: 2 for satisfied, 1 for partially satisfied or unclear, and 0 for not satisfied. 
An instance is considered acceptable if all four criteria are satisfied after adjudication. 
Borderline cases are retained only when the issue does not affect the instance's suitability as a training target, and disagreements on final acceptability are resolved by a third adjudicator. 
We report criterion-level pass rates, overall acceptability, and Cohen's $\kappa$ for the binary acceptable/not acceptable decision before adjudication.

\begin{wraptable}{r}{0.46\textwidth}
\centering
\small
\setlength{\tabcolsep}{4pt}
\renewcommand{\arraystretch}{1.12}
\resizebox{\linewidth}{!}{%
\begin{tabular}{lc}
\toprule
\textbf{Criterion} & \textbf{Pass Rate} \\
\midrule
Concern preservation & 93.2\% \\
Rebuttal-supported revision guidance & 86.4\% \\
Location validity & 88.0\% \\
No rebuttal leakage & 97.6\% \\
\midrule
Overall acceptability & 84.8\% \\
\bottomrule
\end{tabular}
}
\caption{Human audit results for rebuttal-guided feedback enhancement. Pass rates are computed after adjudication. Inter-annotator agreement for the binary acceptability decision is $\kappa=0.78$.}
\label{tab:enhancement_quality}
\end{wraptable}

\paragraph{Results and error analysis.}
As shown in Table~\ref{tab:enhancement_quality}, the enhanced targets achieve high overall quality, with 84.8\% of audited instances judged acceptable after adjudication. 
Concern preservation is strong at 93.2\%, indicating that the enhancement step usually retains the intent of the original reviewer weakness rather than rewriting it into a different critique. 
The highest pass rate is observed for no rebuttal leakage at 97.6\%, suggesting that the pipeline effectively converts rebuttal-derived information into initial-review-style guidance without explicitly exposing post-rebuttal context.

Most remaining errors arise from two sources. 
First, rebuttal-supported revision guidance has the lowest pass rate at 86.4\%. 
These failures typically occur when the aligned rebuttal span provides only indirect evidence, causing the enhanced output to infer a revision action that is plausible but not clearly supported. 
Second, location validity fails when the available paper context is incomplete or when the model proposes an overly broad location, such as a general method or experiment section, instead of a concrete revision site. 
In contrast, rebuttal leakage is rare, suggesting that the main limitation of the enhancement step is not stylistic contamination from rebuttals, but the difficulty of converting partial author responses into precise, well-grounded revision guidance.

\subsection{Localized Evidence Retrieval}
\label{app:retrieval}

To construct informative yet efficient model inputs, we replace full-paper conditioning with localized evidence retrieval. 
This design is motivated by the observation that different weakness categories require different regions of the paper. 
Experimental-design concerns often require result tables, ablations, baselines, datasets, or evaluation protocols, whereas theoretical or methodological concerns may require assumptions, derivations, model definitions, equations, or appendix content.

\paragraph{PDF preprocessing and chunk construction.}
For each enhanced-review record, we retrieve the corresponding paper PDF from the stored paper URL. 
When the PDF URL is unavailable or fails, we attempt OpenReview-based fallback URLs derived from the paper identifier. 
We then extract page-level text using a PDF parser and segment each page into paragraph-level chunks. 
Chunks are assigned page numbers and sequential chunk identifiers. 
Very short chunks are discarded, and each remaining chunk is assigned a coarse chunk type based on lexical and structural cues, such as \textit{method}, \textit{implementation}, \textit{protocol}, \textit{analysis}, \textit{table\_result}, or \textit{generic}. 
This chunk typing is used only as a retrieval signal rather than as a supervised label.

\paragraph{Task-specific retrieval queries.}
Retrieval is performed offline and is task-specific. 
For Task~1, the retrieval query is constructed from the enhanced diagnostic claim, evidence statement, original reviewer weakness, follow-up discussion when available, and paper metadata such as title, weakness labels, and keywords. 
The goal is to retrieve chunks that support weakness diagnosis, namely chunks that indicate whether the paper contains, omits, or insufficiently supports the content described in the weakness.

For Task~2, we use two complementary retrieval channels. 
The first channel retrieves \textit{evidence-support} chunks using the diagnostic claim, evidence statement, original weakness, follow-up discussion, rebuttal-derived construction fields, and paper metadata. 
The second channel retrieves \textit{revision-support} chunks using the actionable suggestion fields, including \textit{what} should be changed, \textit{where} the revision should be made, \textit{how} it should be implemented, and the expected outcome. 
This separation allows Task~2 inputs to include both evidence for the criticism and localized context useful for formulating concrete revision guidance.

\paragraph{Scoring-based chunk ranking.}
Rather than using the full paper as model input, we rank candidate chunks with a deterministic scoring function. 
The score combines lexical overlap, phrase overlap, section-hint matches, page-hint matches, structural-anchor matches, and task-specific bias terms. 
Structural anchors include references to equations, algorithms, figures, tables, sections, appendices, lemmas, theorems, named methods, and other paper-specific markers. 
The ranking also uses category-sensitive biases: empirical weaknesses increase the weight of result tables, protocols, baselines, and implementation details; theory or novelty weaknesses increase the weight of proofs, propositions, related-work discussion, and method sections; ablation-related weaknesses increase the weight of ablation analyses and appendix results. 
Chunks from irrelevant sections such as references, acknowledgments, or generic societal-impact statements are penalized.

\paragraph{Neighborhood expansion and final selection.}
After an initial ranking pass, we expand around high-scoring chunks by including nearby chunks from the same PDF neighborhood. 
This helps recover useful surrounding context when relevant information is split across adjacent paragraphs. 
The expanded candidates are rescored and deduplicated by normalized text prefix. 
In the default setting, Task~1 retains the top-2 diagnostic-evidence chunks. 
For Task~2, we separately rank evidence-support chunks and revision-support chunks, then merge the two lists into a balanced final context with top-5 chunks. 
This merged context is designed to support both claim grounding and concrete suggestion generation.

\paragraph{Retrieval and rebuttal separation.}
During data construction, rebuttal spans are used to identify revision actions and construct enhanced feedback.
However, the generation model does not receive rebuttal text during training or evaluation.
During training, it receives only paper metadata, the target weakness label, and the retrieved paper chunks; at \bench{} evaluation time, retrieved chunks are replaced with full-paper context, as described in Section~\ref{sec:retrieval}.
This separation prevents the model from relying on author responses at inference time and ensures that generated feedback is conditioned only on information available to an initial reviewer.

\paragraph{Retrieval example.}
Table~\ref{tab:retrieval_example} illustrates task-specific retrieval for a theoretical-justification weakness. 
The example shows that Task~1 retrieval focuses on evidence for diagnosing the concern, while Task~2 retrieval additionally selects chunks that help identify where and how a revision can be made.

\begin{table*}[t]
\centering
\small
\setlength{\tabcolsep}{4pt}
\renewcommand{\arraystretch}{1.2}
\begin{tabularx}{\textwidth}{p{2.6cm} X}
\toprule
\textbf{Component} & \textbf{Content} \\
\midrule

\textbf{Paper} &
\emph{Towards simple time-to-event modeling: optimizing neural networks via rank regression} \\

\midrule
\textbf{Weakness} &
The rank-based loss from Jin et al.\ (2003) was derived for linear models, and there is no guarantee that in the non-linear neural network setting the optimal parameters correspond to a zero rank statistic as intended. \\

\midrule
\textbf{Task 1 Retrieved Chunks} &
\textbf{Chunk 1 (Page 5, Method):}  
Derivation of the rank-based loss from Gehan's statistic, highlighting its theoretical grounding in linear model settings. \\

& \textbf{Chunk 2 (Page 5, Method):}  
Discussion noting that the asymptotic properties of the rank statistic are established for linear predictors and are not explicitly justified for non-linear models. \\

& \textbf{Chunk 3 (Page 4, Analysis):}  
Description of combining non-linear neural representations with a rank-based objective, without clarifying whether the original statistical guarantees still hold. \\

\midrule
\textbf{Task 2 Evidence Chunks} &
\textbf{Chunk 1 (Page 5):}  
Explicit statement that theoretical guarantees derived for Gehan's rank statistic may not directly generalize to non-linear predictor functions. \\

& \textbf{Chunk 2 (Page 4):}  
Explanation of the rank-based loss formulation used in the neural network setting, showing the direct transfer from linear models. \\

& \textbf{Chunk 3 (Page 2):}  
Motivation for adapting statistical objectives to neural networks, without providing formal justification of consistency or optimality. \\

\midrule
\textbf{Task 2 Support Chunks} &
\textbf{Chunk 1 (Page 5, Method):}  
Formal definition of the rank-based loss and its connection to Gehan's statistic, indicating where theoretical assumptions are introduced. \\

& \textbf{Chunk 2 (Page 5, Discussion):}  
Text discussing the applicability limitations of rank-based estimators when extended beyond linear settings. \\

& \textbf{Chunk 3 (Page 4, Method):}  
Model formulation combining neural networks with rank-based loss, identifying where clarification or justification can be added. \\

& \textbf{Chunk 4 (Appendix / Theory):}  
Supplementary discussion or lack of proof regarding consistency of the objective in the non-linear setting. \\

\midrule
\textbf{Final Input} &
Paper metadata (title, abstract, keywords), weakness statement, and top retrieved chunks. We use top-2 chunks for Task~1 and top-5 merged evidence/support chunks for Task~2. Rebuttal text is not included in the model input. \\

\bottomrule
\end{tabularx}
\caption{
Example of task-specific context retrieval for a theoretical-justification weakness. 
Task~1 retrieval focuses on evidence for diagnosing the weakness, while Task~2 retrieval additionally includes chunks that help localize and justify concrete revision guidance.
}
\label{tab:retrieval_example}
\end{table*}

\paragraph{Retrieval ablation and chunk-size sensitivity.}
\label{retri_sen}
We analyze retrieval in two ways. 
First, we compare localized top-$k$ retrieval with full-paper conditioning. 
Retrieval remains competitive with full-text inputs while using substantially shorter contexts, supporting the design choice of task-specific context selection. 
Full-paper inputs provide marginal gains on a small subset of metrics, but these gains are limited and come with much higher input cost. 
We therefore interpret retrieval primarily as an efficiency and focus improvement: it filters out irrelevant sections and concentrates the model input around evidence and revision locations relevant to the target weakness.

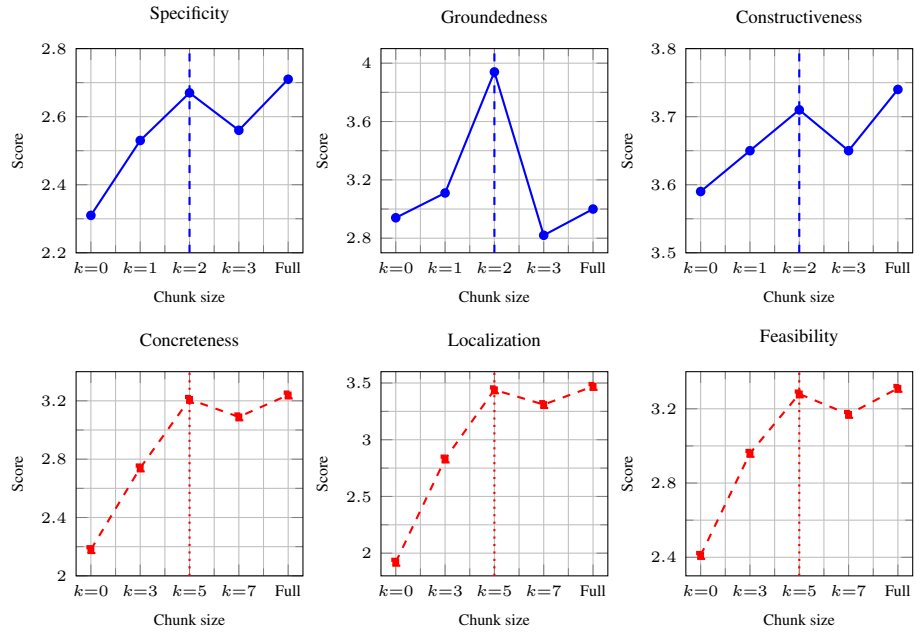
\begin{figure*}[t]
\centering

\begin{subfigure}[t]{0.235\textwidth}
\centering
\begin{tikzpicture}
\begin{axis}[
    width=3.0cm,
    height=2.7cm,
    scale only axis,
    xmin=0.7, xmax=5.3,
    ymin=2.2, ymax=2.8,
    xtick={1,2,3,4,5},
    xticklabels={$k{=}0$,$k{=}1$,$k{=}2$,$k{=}3$,Full},
    ytick={2.2,2.4,2.6,2.8},
    xlabel={Chunk size},
    ylabel={Score},
    title={Specificity},
    grid=both,
    minor tick num=1,
    tick label style={font=\tiny},
    label style={font=\tiny},
    title style={font=\scriptsize},
    every axis plot/.append style={thick, mark size=1.4pt},
]
\addplot[blue, mark=*] coordinates {
    (1,2.31) (2,2.53) (3,2.67) (4,2.56) (5,2.71)
};
\addplot[blue, dashed] coordinates {(3,2.2) (3,2.8)};
\end{axis}
\end{tikzpicture}
\end{subfigure}
\hspace{0.006\textwidth}
\begin{subfigure}[t]{0.235\textwidth}
\centering
\begin{tikzpicture}
\begin{axis}[
    width=3.0cm,
    height=2.7cm,
    scale only axis,
    xmin=0.7, xmax=5.3,
    ymin=2.7, ymax=4.1,
    xtick={1,2,3,4,5},
    xticklabels={$k{=}0$,$k{=}1$,$k{=}2$,$k{=}3$,Full},
    ytick={2.8,3.2,3.6,4.0},
    xlabel={Chunk size},
    ylabel={Score},
    title={Groundedness},
    grid=both,
    minor tick num=1,
    tick label style={font=\tiny},
    label style={font=\tiny},
    title style={font=\scriptsize},
    every axis plot/.append style={thick, mark size=1.4pt},
]
\addplot[blue, mark=*] coordinates {
    (1,2.94) (2,3.11) (3,3.94) (4,2.82) (5,3.00)
};
\addplot[blue, dashed] coordinates {(3,2.7) (3,4.1)};
\end{axis}
\end{tikzpicture}
\end{subfigure}
\hspace{0.006\textwidth}
\begin{subfigure}[t]{0.235\textwidth}
\centering
\begin{tikzpicture}
\begin{axis}[
    width=3.0cm,
    height=2.7cm,
    scale only axis,
    xmin=0.7, xmax=5.3,
    ymin=3.5, ymax=3.8,
    xtick={1,2,3,4,5},
    xticklabels={$k{=}0$,$k{=}1$,$k{=}2$,$k{=}3$,Full},
    ytick={3.5,3.6,3.7,3.8},
    xlabel={Chunk size},
    ylabel={Score},
    title={Constructiveness},
    grid=both,
    minor tick num=1,
    tick label style={font=\tiny},
    label style={font=\tiny},
    title style={font=\scriptsize},
    every axis plot/.append style={thick, mark size=1.4pt},
]
\addplot[blue, mark=*] coordinates {
    (1,3.59) (2,3.65) (3,3.71) (4,3.65) (5,3.74)
};
\addplot[blue, dashed] coordinates {(3,3.5) (3,3.8)};
\end{axis}
\end{tikzpicture}
\end{subfigure}

\vspace{0.25em}

\begin{subfigure}[t]{0.235\textwidth}
\centering
\begin{tikzpicture}
\begin{axis}[
    width=3.0cm,
    height=2.7cm,
    scale only axis,
    xmin=0.7, xmax=5.3,
    ymin=2.0, ymax=3.4,
    xtick={1,2,3,4,5},
    xticklabels={$k{=}0$,$k{=}3$,$k{=}5$,$k{=}7$,Full},
    ytick={2.0,2.4,2.8,3.2},
    xlabel={Chunk size},
    ylabel={Score},
    title={Concreteness},
    grid=both,
    minor tick num=1,
    tick label style={font=\tiny},
    label style={font=\tiny},
    title style={font=\scriptsize},
    every axis plot/.append style={thick, mark size=1.3pt},
]
\addplot[red, dashed, mark=square*] coordinates {
    (1,2.18) (2,2.74) (3,3.21) (4,3.09) (5,3.24)
};
\addplot[red, dotted] coordinates {(3,2.0) (3,3.4)};
\end{axis}
\end{tikzpicture}
\end{subfigure}
\hspace{0.006\textwidth}
\begin{subfigure}[t]{0.235\textwidth}
\centering
\begin{tikzpicture}
\begin{axis}[
    width=3.0cm,
    height=2.7cm,
    scale only axis,
    xmin=0.7, xmax=5.3,
    ymin=1.8, ymax=3.6,
    xtick={1,2,3,4,5},
    xticklabels={$k{=}0$,$k{=}3$,$k{=}5$,$k{=}7$,Full},
    ytick={2.0,2.5,3.0,3.5},
    xlabel={Chunk size},
    ylabel={Score},
    title={Localization},
    grid=both,
    minor tick num=1,
    tick label style={font=\tiny},
    label style={font=\tiny},
    title style={font=\scriptsize},
    every axis plot/.append style={thick, mark size=1.3pt},
]
\addplot[red, dashed, mark=square*] coordinates {
    (1,1.92) (2,2.83) (3,3.44) (4,3.31) (5,3.47)
};
\addplot[red, dotted] coordinates {(3,1.8) (3,3.6)};
\end{axis}
\end{tikzpicture}
\end{subfigure}
\hspace{0.006\textwidth}
\begin{subfigure}[t]{0.235\textwidth}
\centering
\begin{tikzpicture}
\begin{axis}[
    width=3.0cm,
    height=2.7cm,
    scale only axis,
    xmin=0.7, xmax=5.3,
    ymin=2.3, ymax=3.4,
    xtick={1,2,3,4,5},
    xticklabels={$k{=}0$,$k{=}3$,$k{=}5$,$k{=}7$,Full},
    ytick={2.4,2.8,3.2},
    xlabel={Chunk size},
    ylabel={Score},
    title={Feasibility},
    grid=both,
    minor tick num=1,
    tick label style={font=\tiny},
    label style={font=\tiny},
    title style={font=\scriptsize},
    every axis plot/.append style={thick, mark size=1.3pt},
]
\addplot[red, dashed, mark=square*] coordinates {
    (1,2.41) (2,2.96) (3,3.28) (4,3.17) (5,3.31)
};
\addplot[red, dotted] coordinates {(3,2.3) (3,3.4)};
\end{axis}
\end{tikzpicture}
\end{subfigure}

\caption{
Effect of retrieved chunk size on generation quality across Task~1 and Task~2, as judged by GPT-5.4. 
Blue solid lines denote Task~1 and red dashed lines denote Task~2. 
Vertical reference lines mark the default chunk size for each task: $k{=}2$ for Task~1 and $k{=}5$ for Task~2.
}
\label{fig:retrieval_k_both_tasks}
\end{figure*}

Second, we evaluate the effect of retrieved chunk size in Figure~\ref{fig:retrieval_k_both_tasks}. 
Task~1 is relatively insensitive to chunk size: small localized contexts are usually sufficient for diagnostic claim generation. 
Task~2 is more retrieval-sensitive because actionable suggestions require both evidence for the weakness and localization cues for revision. 
Performance improves with additional context and generally plateaus around $k{=}5$, supporting our default setting of top-2 chunks for Task~1 and top-5 chunks for Task~2.

\paragraph{Retrieval quality validation.}
We validate retrieval quality with two complementary checks: GPT-5.4 screening and human audit. 
Both checks evaluate the retrieved context itself, rather than the generated model output. 
For each sampled instance, the evaluator is shown the paper metadata, target weakness label, reviewer weakness, and retrieved chunks, but not the rebuttal. 
The evaluator then judges whether the retrieved chunks satisfy four criteria: 
(1) \textit{Weakness-relevant evidence}, whether the chunks contain paper content related to the target weakness; 
(2) \textit{Specific grounding support}, whether the chunks are specific enough to support grounded feedback rather than generic criticism; 
(3) \textit{Revision localization cues}, whether the chunks indicate where or how the paper could be revised, such as a method section, experiment, table, figure, claim, or appendix; and 
(4) \textit{Overall sufficient context}, whether the retrieved context is sufficient for the corresponding generation task. 
For Task~1, evaluators focus on whether the chunks support weakness diagnosis; for Task~2, they additionally assess whether the chunks provide localization cues for actionable revision guidance. 
Each criterion is judged as pass or fail, and Table~\ref{tab:retrieval_quality} reports pass rates on a 0--100 scale, where higher values indicate better retrieval quality. 
GPT-5.4 screening is used only for analysis and not as a training signal for the generator. 
Human annotators follow the same criteria on a stratified subset of retrieved contexts. 
As shown in Table~\ref{tab:retrieval_quality}, both GPT-5.4 and human annotators find that most retrieved contexts contain weakness-relevant evidence and provide sufficient grounding support. 
Task~2 localization is more challenging, but the dual-channel retrieval design still yields strong localization support, with human pass rate of 78.6\%. The prompt is provided in Figure ~\ref{fig:retrieval_quality_prompt}.

\begin{table*}[t]
\centering
\small
\setlength{\tabcolsep}{5pt}
\renewcommand{\arraystretch}{1.15}
\begin{tabular}{lcccc}
\toprule
\multirow{2}{*}{\textbf{Retrieval Quality Criterion}}
& \multicolumn{2}{c}{\textbf{Task 1: Diagnostic Claims}}
& \multicolumn{2}{c}{\textbf{Task 2: Actionable Suggestions}} \\
\cmidrule(lr){2-3} \cmidrule(lr){4-5}
& \textbf{LLM} & \textbf{Human}
& \textbf{LLM} & \textbf{Human} \\
\midrule
Weakness-relevant evidence
& 90.4 & 87.2 & 88.6 & 85.5 \\
Specific grounding support
& 86.7 & 83.9 & 85.1 & 82.4 \\
Revision localization cues
& N/A & N/A & 82.8 & 78.6 \\
Overall sufficient context
& 85.9 & 82.7 & 84.3 & 80.8 \\
\bottomrule
\end{tabular}
\caption{
Retrieval quality validation on a 200 stratified subset of retrieved contexts. 
Values report pass rates judged by GPT-5.4 screening and human annotators. 
Task~1 focuses on whether retrieved chunks support weakness diagnosis, while Task~2 additionally evaluates whether chunks provide localization cues for actionable revision guidance.
}
\label{tab:retrieval_quality}
\end{table*}

\begin{figure*}[t]
\centering
\begin{tcolorbox}[
  colback=gray!5,
  colframe=gray!45,
  width=0.96\textwidth,
  fontupper=\small,
  title={Retrieval Quality Screening Prompt},
  breakable
]
You are evaluating the quality of retrieved paper chunks for actionable peer-review generation.

You are given:
\begin{enumerate}[leftmargin=*,nosep]
    \item Paper metadata, including title and abstract.
    \item A target weakness label.
    \item The reviewer-identified weakness.
    \item The task type: Task 1 diagnostic claim generation or Task 2 actionable suggestion generation.
    \item The retrieved paper chunks.
\end{enumerate}

Your task is to judge whether the retrieved chunks provide sufficient context for the given task. 
Do not evaluate any generated model output. 
Do not use or assume access to the author rebuttal. 
Judge only from the provided paper metadata, weakness, task definition, and retrieved chunks.

\textbf{Evaluation criteria.}
For each criterion, output \texttt{pass} or \texttt{fail}, with a short explanation.

\begin{itemize}[leftmargin=*]
    \item \textbf{Weakness-relevant evidence}: Do the retrieved chunks contain paper content related to the target weakness?
    \item \textbf{Specific grounding support}: Are the chunks specific enough to support grounded feedback, rather than only generic or superficial keyword matches?
    \item \textbf{Revision localization cues}: Do the chunks indicate where or how the paper could be revised, such as a method section, experiment, table, figure, claim, equation, or appendix? 
    For Task 1, this criterion may be judged based on whether the chunks localize the diagnosed issue. 
    For Task 2, this criterion should judge whether the chunks support concrete revision guidance.
    \item \textbf{Overall sufficient context}: Overall, are the retrieved chunks sufficient for the target task?
\end{itemize}

\textbf{Task-specific interpretation.}
For Task 1, focus on whether the chunks support diagnosing the target weakness. 
For Task 2, focus on whether the chunks support both the weakness diagnosis and actionable revision guidance.

Return only valid JSON in the following format:
\begin{verbatim}
{
  "weakness_relevant_evidence": {
    "decision": "pass/fail",
    "reason": "short explanation"
  },
  "specific_grounding_support": {
    "decision": "pass/fail",
    "reason": "short explanation"
  },
  "revision_localization_cues": {
    "decision": "pass/fail",
    "reason": "short explanation"
  },
  "overall_sufficient_context": {
    "decision": "pass/fail",
    "reason": "short explanation"
  }
}
\end{verbatim}
\end{tcolorbox}
\caption{Prompt used for GPT-5.4 retrieval quality screening. The prompt evaluates retrieved chunks only and does not use rebuttal text or generated model outputs.}
\label{fig:retrieval_quality_prompt}
\end{figure*}

\subsection{Implementation Details and Compute Budget}
\label{app:implementation_details}

We use Qwen3-8B-Base as the backbone model for both supervised fine-tuning and reinforcement learning. 
For comparison, we also evaluate larger or specialized baselines, including Qwen3-32B, DeepReviewer-14B, OpenReviewer-8B, and RbtAct-8B when available. 
For \ours{} training, the model uses paper metadata, the target weakness label, and retrieved paper chunks as input. At \bench{} evaluation time, all evaluated systems receive the same paper metadata, target weakness label, and full-paper context, and no rebuttal text is provided during evaluation.

For supervised fine-tuning, we train a single multi-task model jointly on Task~1 and Task~2 using the enhanced instances in \data.
We use full-parameter fine-tuning with the standard autoregressive language modeling objective. 
For reinforcement learning, we initialize from the corresponding SFT model and apply GRPO with $K=8$ sampled responses per prompt. 
Rewards are computed from frozen instance-specific rubrics, using deterministic checks for format-related criteria and GPT-5.4-based scoring for semantic criteria. 
KL regularization is applied toward the SFT reference model.

Experiments were conducted on four NVIDIA H200 GPUs with 143GB memory per GPU. 
The main SFT and GRPO runs were performed using distributed training across the four GPUs. 
We used fixed training configurations for the main model variants rather than extensive hyperparameter search. 
The main computational cost comes from full-parameter SFT, GRPO sampling and scoring, candidate generation for rubric construction, and LLM-based evaluation. 
Detailed hyperparameters, decoding settings, and prompt templates are provided in the released code to support reproducibility.
\section{Rubric Construction and Reward Design}
\label{app:rubric}

This appendix provides details on the rubric construction and reward design used in the RL stage of \ours. 
We first describe our candidate-aware rubric construction procedure, then provide the scoring prompt and two reward-ablation prompts: direct judge reward and fixed task-level rubric reward. 
Finally, we describe reward normalization and provide qualitative examples comparing candidate-aware rubrics with generic rubrics.

\subsection{Candidate-Aware Rubric Construction}
\label{app:candidate_aware_rubric}

The goal of rubric construction is to provide an instance-specific reward specification for open-ended peer-review feedback generation. 
A generic reward prompt can ask whether an output is accurate, specific, or actionable, but it cannot determine which experiment, assumption, paper section, or revision action is relevant for a particular paper--weakness pair. 
For example, a generic rubric may reward an output for recommending ``additional experiments,'' but it cannot distinguish whether the relevant missing experiment is a robustness test, an ablation, a baseline comparison, or a controlled synthetic analysis. 
This motivates our candidate-aware rubric design.

For each RL training instance, we construct a candidate pool consisting of:
\begin{itemize}[leftmargin=*]
    \item multiple outputs sampled from the SFT model under different decoding temperatures;
    \item the human-written or reference-style enhanced output;
    \item a strong LLM-generated reference output;
\end{itemize}

The diverse SFT candidates expose realistic model failure modes, such as generic advice, unsupported claims, missing localization, repeated claims, or shallow suggestions. 
The reference outputs provide positive examples of paper-grounded and revision-oriented feedback. 
Given this candidate pool, GPT-5.4 synthesizes a rubric containing two types of criteria:
\begin{enumerate}[leftmargin=*]
    \item \textbf{Hard constraints}, which specify invalid or severely flawed outputs, such as hallucinating paper content, referring to the rebuttal, addressing the wrong weakness label, or failing to follow the required task format.
    \item \textbf{Weighted soft requirements}, which describe desirable instance-specific properties. Each soft requirement is assigned a weight, and the weights sum to 1.0.
\end{enumerate}

The rubric is constructed offline before RL training and then frozen. 
During GRPO, the policy does not update the rubric, and reference claims or suggestions are not used as direct supervision. 
They are only used offline to help define the reward criteria.
\paragraph{Candidate-aware rubric construction prompt.}
Figure~\ref{fig:candidate_aware_rubric_prompt} summarizes the prompt template used to construct candidate-aware rubrics.
\begin{figure*}[t]
\centering
\setlength{\fboxsep}{0pt}

\begin{tcolorbox}[
    colback=gray!8,
    colframe=yellow!50!black,
    boxrule=0.6pt,
    arc=1pt,
    left=8pt,
    right=8pt,
    top=8pt,
    bottom=8pt,
    width=0.95\textwidth
]
\footnotesize
\textbf{Prompt for Candidate-Aware Rubric Construction}

\vspace{0.5em}
You are constructing an instance-specific reward rubric for peer-review feedback generation.

\vspace{0.6em}
\textbf{Input:}
You are given:
\begin{enumerate}[leftmargin=1.5em,itemsep=0.2em,topsep=0.2em]
    \item The task type, either diagnostic-claim generation or actionable-suggestion generation.
    \item Paper metadata, including title, abstract, keywords, and weakness labels.
    \item The original reviewer weakness.
    \item Retrieved paper context chunks.
    \item A pool of candidate outputs sampled from the SFT model.
    \item A human-written or reference-style enhanced output.
    \item A strong LLM-generated reference output.
\end{enumerate}

\vspace{0.6em}
\textbf{Goal:}
Construct a rubric that can distinguish high-quality, paper-grounded feedback from weak, generic, unsupported, or incorrectly formatted outputs.

The rubric should be specific to this paper--weakness instance.
It should reward outputs that correctly identify the target deficiency, ground the feedback in the provided paper context, and provide concrete revision guidance when required by the task.

\vspace{0.6em}
\textbf{Hard constraints:}
List invalid or severely flawed behaviors.
Examples include:
\begin{itemize}[leftmargin=1.5em,itemsep=0.2em,topsep=0.2em]
    \item The output addresses the wrong weakness or changes the target weakness label.
    \item The output hallucinates paper content not supported by the retrieved context.
    \item The output mentions the rebuttal, author response, revised paper, or any post-submission information.
    \item The output fails to follow the required task format.
    \item The output gives only generic advice without paper-specific grounding.
\end{itemize}

\vspace{0.6em}
\textbf{Weighted soft requirements:}
Create a set of instance-specific soft requirements.
Each requirement should describe one desirable property of a high-quality answer.
Assign each requirement a weight, and make sure the weights sum to 1.0.

Soft requirements should cover dimensions such as:
\begin{itemize}[leftmargin=1.5em,itemsep=0.2em,topsep=0.2em]
    \item correctness of the diagnostic claim;
    \item grounding in retrieved paper evidence;
    \item specificity to the target weakness and paper context;
    \item localization of the relevant section, experiment, table, figure, method component, or claim;
    \item actionability of the proposed revision;
    \item completeness of the required output fields;
    \item avoidance of unsupported or overly broad suggestions.
\end{itemize}

\vspace{0.6em}
\textbf{Candidate-aware calibration:}
Use the candidate pool to identify common failure modes.
The rubric should penalize failures observed in weak candidates and reward properties demonstrated by strong candidates.
Do not simply copy the reference output.
Instead, abstract the qualities that make the reference output useful, grounded, and actionable.

\vspace{0.6em}
\textbf{Required output format:}
Return a structured rubric with:
\begin{itemize}[leftmargin=1.5em,itemsep=0.2em,topsep=0.2em]
    \item \textbf{Hard constraints}: a list of invalid or severely flawed output behaviors.
    \item \textbf{Soft requirements}: a list of weighted criteria, each with a name, definition, weight, and scoring guidance.
    \item \textbf{Notes}: brief instance-specific reminders about what the model should focus on.
\end{itemize}
\end{tcolorbox}

\caption{
Prompt template for candidate-aware rubric construction.
The rubric is constructed offline from SFT candidates, reference-style enhanced outputs, and strong LLM-generated outputs, then frozen before reinforcement learning.
}
\label{fig:candidate_aware_rubric_prompt}
\end{figure*}

\subsection{Rubric Scoring Prompt}
\label{app:rubric_scoring_prompt}

After constructing a frozen rubric for each RL instance, we use an LLM judge to score sampled model outputs against the rubric. 
For each weighted soft requirement, the judge receives the requirement and the model output, and returns an integer score from 1 to 5. 
The requirement-level scores are normalized to $[0,1]$ and combined according to the rubric weights. (prompt shown in Figure ~\ref{fig:rubric_scoring_prompt})
\begin{figure*}[t]
\centering
\setlength{\fboxsep}{0pt}

\begin{tcolorbox}[
    colback=gray!8,
    colframe=yellow!50!black,
    boxrule=0.6pt,
    arc=1pt,
    left=8pt,
    right=8pt,
    top=8pt,
    bottom=8pt,
    width=0.95\textwidth
]
\footnotesize
\textbf{Prompt for Rubric Requirement Scoring}

\vspace{0.5em}
You are a strict but fair grader evaluating the quality of an academic peer review output.

You will be given a Task 1 output and a single grading requirement.

\vspace{0.6em}
\textbf{Task:}
Score how well the output satisfies the requirement using one integer from 1 to 5.

\vspace{0.6em}
\textbf{Scale:}
\begin{itemize}[leftmargin=1.5em,itemsep=0.2em,topsep=0.2em]
    \item \textbf{5}: The requirement is fully and clearly satisfied.
    \item \textbf{4}: The requirement is mostly satisfied, with only minor gaps.
    \item \textbf{3}: The requirement is partially satisfied.
    \item \textbf{2}: The requirement is barely satisfied.
    \item \textbf{1}: The requirement is not satisfied.
\end{itemize}

\vspace{0.6em}
\textbf{Important rules:}
\begin{itemize}[leftmargin=1.5em,itemsep=0.2em,topsep=0.2em]
    \item Use the full 1--5 range when appropriate.
    \item Do not default to 3 unless the evidence is genuinely mixed.
    \item Return only a single integer: 1, 2, 3, 4, or 5.
    \item Do not provide any explanation.
\end{itemize}

\vspace{0.6em}
\textbf{Input format:}

\vspace{0.2em}
\noindent
\texttt{\#\# Task Type}

\noindent
Task 1 (Weakness Claim Discovery)

\vspace{0.4em}
\noindent
\texttt{\#\# Grading Requirement}

\noindent
\texttt{[Requirement from the frozen rubric]}

\vspace{0.4em}
\noindent
\texttt{\#\# Review Output to Grade}

\noindent
\texttt{[Model-generated output]}

\vspace{0.4em}
\noindent
\texttt{Score (1--5):}
\end{tcolorbox}

\caption{
Prompt used to score each weighted soft requirement in the frozen rubric.
The judge returns a single integer score from 1 to 5 for each requirement, which is then normalized and combined using the rubric weights.
}
\label{fig:rubric_scoring_prompt}
\end{figure*}

\subsection{Direct Judge Reward Prompt}
\label{app:direct_judge_reward_prompt}

To isolate the effect of rubric design, we compare our candidate-aware rubric reward with a simpler direct-judge reward.
In this ablation, the LLM judge scores each sampled output using fixed task-level dimensions rather than an instance-specific rubric.
This baseline tests whether applying GRPO with a generic LLM judge is sufficient, or whether explicit candidate-aware rubric construction provides additional benefit. (prompt shown in Figure ~\ref{fig:direct_judge_reward_prompt})

\begin{figure*}[t]
\centering
\setlength{\fboxsep}{0pt}

\begin{tcolorbox}[
    colback=gray!8,
    colframe=yellow!50!black,
    boxrule=0.6pt,
    arc=1pt,
    left=8pt,
    right=8pt,
    top=8pt,
    bottom=8pt,
    width=0.95\textwidth
]
\footnotesize
\textbf{Prompt for Direct Judge Reward}

\vspace{0.5em}
You are an expert evaluator for academic peer-review feedback.

\vspace{0.6em}
\textbf{Input:}
You are given:
\begin{enumerate}[leftmargin=1.5em,itemsep=0.2em,topsep=0.2em]
    \item Paper metadata and retrieved paper context.
    \item A target weakness label.
    \item The task type: Task~1 or Task~2.
    \item A model-generated output.
\end{enumerate}

\vspace{0.6em}
\textbf{Goal:}
Evaluate the model-generated output directly.

Do not use any instance-specific rubric.
Do not compare the output to a reference answer.
Judge the output only based on the paper context, the target weakness label, and the task definition.

\vspace{0.6em}
\textbf{Task definitions:}
\begin{itemize}[leftmargin=1.5em,itemsep=0.2em,topsep=0.2em]
    \item \textbf{Task~1}: The output should contain diagnostic claims that identify concrete deficiencies in the paper under the target weakness label.
    \item \textbf{Task~2}: The output should contain actionable suggestions that explain what should be revised, where the revision should appear, how it can be implemented, and what outcome the revision should achieve.
\end{itemize}

\vspace{0.6em}
\textbf{Scoring dimensions:}
Score each applicable dimension from 0 to 5.

\begin{itemize}[leftmargin=1.5em,itemsep=0.2em,topsep=0.2em]
    \item \textbf{Task~1 Technical Accuracy}: For Task~1 only, is the diagnostic claim factually correct, paper-grounded, and supported by the retrieved context? Penalize hallucinated paper content, unsupported criticisms, or claims that address the wrong weakness.
    \item \textbf{Task~1 Overall}: For Task~1 only, how useful is the output for helping authors understand the core weakness in the paper?
    \item \textbf{Task~2 Actionability}: For Task~2 only, does the output provide concrete \textit{what}, \textit{where}, \textit{how}, and expected-outcome guidance that an author can follow?
    \item \textbf{Task~2 Overall}: For Task~2 only, how useful is the output for helping authors revise the paper to address the target weakness?
\end{itemize}

\vspace{0.6em}
\textbf{Special case:}
If the target weakness is not supported by the provided paper context, an output of \texttt{None} may receive a high Task~1 technical-accuracy score.
However, \texttt{None} should not receive a high overall or actionability score unless the lack of support is clearly justified.

\vspace{0.6em}
\textbf{Required output format:}
Return only valid JSON.
For Task~1, use:
\begin{verbatim}
{
  "task1_technical_accuracy": 0,
  "task1_overall": 0,
  "brief_reason": "short explanation"
}
\end{verbatim}

For Task~2, use:
\begin{verbatim}
{
  "task2_actionability": 0,
  "task2_overall": 0,
  "brief_reason": "short explanation"
}
\end{verbatim}
\end{tcolorbox}

\caption{
Prompt used for the direct-judge reward ablation.
Unlike our candidate-aware rubric reward, this baseline evaluates outputs using fixed task-level dimensions and does not receive an instance-specific rubric.
}
\label{fig:direct_judge_reward_prompt}
\end{figure*}

\subsection{Fixed Task-Level Rubric Prompt}
\label{app:fixed_task_rubric_prompt}

We also evaluate a fixed task-level rubric setting.
Unlike candidate-aware rubrics, the fixed rubric is shared across all instances of the same task.
It captures general task requirements, but does not adapt to the target weakness, paper evidence, or model-specific failure modes.
This ablation tests whether improvements come merely from using a rubric, or from constructing an instance-specific candidate-aware rubric. 

\begin{tcolorbox}[
    enhanced,
    breakable,
    colback=gray!8,
    colframe=yellow!50!black,
    boxrule=0.6pt,
    arc=1pt,
    left=6pt,
    right=6pt,
    top=6pt,
    bottom=6pt,
    width=\linewidth,
    fontupper=\scriptsize,
    title={\scriptsize Prompt for Fixed Task-Level Rubric Reward},
    fonttitle=\bfseries
]
You are an expert evaluator for academic peer-review feedback.

\vspace{0.4em}
\textbf{Input:}
You are given:
\begin{enumerate}[leftmargin=1.2em,itemsep=0.1em,topsep=0.1em]
    \item Paper metadata and retrieved paper context.
    \item A target weakness label.
    \item The task type: Task~1 or Task~2.
    \item A model-generated output.
    \item A fixed task-level rubric.
\end{enumerate}

\vspace{0.4em}
\textbf{Goal:}
Use the fixed task-level rubric to score the model-generated output.
The rubric is shared across all instances of the same task.
It is not customized to the specific paper, weakness, retrieved evidence, or candidate outputs.
Do not construct a new rubric for this instance.
Do not compare the output to a reference answer.
Judge the output only based on the provided paper context, the target weakness label, the task definition, and the fixed rubric below.

\vspace{0.5em}
\textbf{Fixed rubric for Task~1: Diagnostic Claim Generation.}
For Task~1, the output should contain one or more diagnostic claims that identify concrete deficiencies in the paper under the target weakness label.

Score the following dimensions from 0 to 5:
\begin{itemize}[leftmargin=1.2em,itemsep=0.1em,topsep=0.1em]
    \item \textbf{Technical Accuracy}: Is the diagnostic claim factually correct and supported by the retrieved paper context? Penalize hallucinated paper content, unsupported criticisms, or claims that address the wrong weakness.
    \item \textbf{Specificity and Grounding}: Does the claim refer to concrete paper elements, such as methods, experiments, datasets, assumptions, figures, tables, equations, or claims?
    \item \textbf{Depth and Constructiveness}: Does the claim explain why the weakness matters, rather than merely restating a generic concern?
    \item \textbf{Granularity and Format}: Does the output avoid redundant, speculative, overly broad, or excessive claims, and does it follow the required Task~1 output format?
    \item \textbf{Overall Quality}: Overall, how useful is the output for helping authors understand the core weakness in the paper?
\end{itemize}

\vspace{0.5em}
\textbf{Fixed rubric for Task~2: Actionable Suggestion Generation.}
For Task~2, the output should contain actionable suggestions that explain what should be revised, where the revision should appear, how it can be implemented, and what outcome the revision should achieve.

Score the following dimensions from 0 to 5:
\begin{itemize}[leftmargin=1.2em,itemsep=0.1em,topsep=0.1em]
    \item \textbf{Technical Accuracy}: Is the suggestion technically sound and supported by the retrieved paper context? Penalize hallucinated paper content, unsupported claims, or infeasible suggestions.
    \item \textbf{Actionability}: Does the suggestion provide concrete \textit{what}, \textit{where}, \textit{how}, and expected-outcome guidance that an author can follow?
    \item \textbf{Grounding and Localization}: Is the suggestion tied to concrete paper content and localized to relevant sections, experiments, figures, tables, methods, claims, or appendix material?
    \item \textbf{Depth and Constructiveness}: Does the suggestion address the underlying weakness rather than merely restating the problem or asking for vague clarification?
    \item \textbf{Overall Quality}: Overall, how useful is the output as practical revision guidance for addressing the target weakness?
\end{itemize}

\vspace{0.5em}
\textbf{Hard constraints:}
Assign a very low score if the output:
\begin{itemize}[leftmargin=1.2em,itemsep=0.1em,topsep=0.1em]
    \item mentions rebuttals, author responses, revised papers, or post-submission changes;
    \item invents paper content that is not supported by the retrieved context;
    \item addresses a weakness category unrelated to the target label;
    \item gives only generic advice without paper-specific grounding;
    \item fails to follow the required output format for the corresponding task.
\end{itemize}

\vspace{0.4em}
\textbf{Special case:}
If the target weakness is not supported by the provided paper context, an output of \texttt{None} may receive a high technical-accuracy score.
However, \texttt{None} should not receive a high overall-quality score unless the lack of support is clearly justified.

\vspace{0.4em}
\textbf{Required output format:}
Return only valid JSON.

For Task~1, use:
\begin{verbatim}
{
  "technical_accuracy": 0,
  "specificity_grounding": 0,
  "depth_constructiveness": 0,
  "granularity_format": 0,
  "overall_quality": 0,
  "hard_constraint_violation": false,
  "brief_reason": "short explanation"
}
\end{verbatim}

For Task~2, use:
\begin{verbatim}
{
  "technical_accuracy": 0,
  "actionability": 0,
  "grounding_localization": 0,
  "depth_constructiveness": 0,
  "overall_quality": 0,
  "hard_constraint_violation": false,
  "brief_reason": "short explanation"
}
\end{verbatim}
\end{tcolorbox}

\subsection{Rubric Examples}
\label{app:rubric_examples}

Tables~\ref{tab:case_task1_generic_vs_candidate_aware} and~\ref{tab:case_task2_generic_vs_candidate_aware} show examples comparing candidate-aware rubric construction with generic rubric construction. 
The generic rubrics capture broad quality dimensions, such as whether the response asks for more experiments or stronger empirical support. 
In contrast, the candidate-aware rubrics specify the concrete missing evidence, relevant paper setting, expected revision location, and failure modes observed in candidate outputs. 
This additional specificity makes the reward more discriminative for the target paper--weakness instance.

\begin{table*}[t]
\centering
\small
\setlength{\tabcolsep}{5pt}
\renewcommand{\arraystretch}{1.2}
\begin{tabularx}{\textwidth}{p{0.47\textwidth} p{0.47\textwidth}}
\toprule
\textbf{Candidate-aware rubric (example-specific)} & \textbf{Generic rubric} \\
\midrule

\textbf{Soft requirements}
\begin{itemize}[leftmargin=*]
    \item Identifies that the toy/synthetic point-cloud experiment is evaluated only qualitatively, without quantitative metrics.
    \item Notes the absence of analysis on how the metric behaves under controlled geometric properties such as \textit{curvature} or \textit{sharpness}.
    \item Identifies the lack of robustness evaluation under realistic point-cloud corruptions such as \textit{noise}, \textit{occlusion}, missing points, or outliers.
    \item Rewards claims grounded in the actual paper setting, namely a statistical/Riemannian framework for point clouds.
    \item Penalizes hallucinated concerns such as generic classification-baseline or ablation complaints that are not supported by the paper context.
    \item Rewards coverage of distinct failure modes rather than repeated paraphrases of the same generic issue.
\end{itemize}

\textbf{Hard constraints}
\begin{itemize}[leftmargin=*]
    \item Output must be formatted as numbered diagnostic claims or \texttt{None}.
    \item Must not reference rebuttal or post-submission content.
    \item Must not introduce unsupported tasks, datasets, or baselines.
\end{itemize}
&
\textbf{Soft requirements}
\begin{itemize}[leftmargin=*]
    \item The response should identify one or more weaknesses showing that the experimental evaluation is limited or insufficient.
    \item The response should mention that the paper would benefit from more quantitative results or broader empirical validation.
    \item The response should note that additional experiments, benchmarks, or comparisons could strengthen the paper.
    \item The response should remain relevant to the weakness label and avoid obviously unrelated criticism.
    \item The response should present concise and non-duplicative claims.
\end{itemize}

\textbf{Hard constraints}
\begin{itemize}[leftmargin=*]
    \item Output must be formatted as numbered diagnostic claims or \texttt{None}.
    \item Must not reference rebuttal or post-submission content.
\end{itemize}
\\

\bottomrule
\end{tabularx}
\caption{Task~1 rubric example comparing candidate-aware rubric construction with generic rubric construction. The candidate-aware rubric specifies concrete missing evidence and paper-specific failure modes, while the generic rubric mainly rewards broad empirical-evaluation concerns.}
\label{tab:case_task1_generic_vs_candidate_aware}
\end{table*}

\begin{table*}[t]
\centering
\small
\setlength{\tabcolsep}{5pt}
\renewcommand{\arraystretch}{1.2}
\begin{tabularx}{\textwidth}{p{0.47\textwidth} p{0.47\textwidth}}
\toprule
\textbf{Candidate-aware rubric (example-specific)} & \textbf{Generic rubric} \\
\midrule

\textbf{Soft requirements}
\begin{itemize}[leftmargin=*]
    \item In \textbf{Evidence}, explicitly states that the toy/synthetic dataset is currently evaluated only qualitatively, with no numerical metrics reported.
    \item In \textbf{Suggestion}, proposes a controlled synthetic experiment that varies geometric properties such as curvature or sharp edges, rather than vaguely asking for ``more experiments.''
    \item Requires at least one concrete metric family, such as reconstruction error, geodesic/distance distortion, or pairwise distance preservation.
    \item Requires a specific placement for the revision, such as immediately after the toy-dataset discussion or in a dedicated appendix subsection on synthetic quantitative results.
    \item Requires the \textbf{Expected Outcome} to explain the diagnostic purpose: revealing how the proposed Riemannian metric responds to controlled geometric changes, thereby supporting or clarifying the qualitative claims.
    \item Penalizes suggestions that are only generic requests for broader evaluation without explaining what should be measured or where the revision should be added.
\end{itemize}

\textbf{Hard constraints}
\begin{itemize}[leftmargin=*]
    \item Output must include diagnostic claim and structured revision guidance.
    \item Revision guidance must include what to revise, where to revise, how to implement the revision, and the expected outcome.
    \item Must not reference rebuttal or post-submission content.
\end{itemize}
&
\textbf{Soft requirements}
\begin{itemize}[leftmargin=*]
    \item The evidence should explain why the current experimental support is limited.
    \item The suggestions should recommend adding stronger empirical validation.
    \item The suggestions should be actionable and mention possible quantitative analysis or comparisons.
    \item The expected outcome should explain that the added experiments would strengthen the paper.
    \item The severity assessment should reflect whether the missing evaluation weakens the paper's empirical support.
\end{itemize}

\textbf{Hard constraints}
\begin{itemize}[leftmargin=*]
    \item Output must include the required structured sections.
    \item Must not reference rebuttal or post-submission content.
\end{itemize}
\\

\bottomrule
\end{tabularx}
\caption{Task~2 rubric example comparing candidate-aware rubric construction with generic rubric construction. The candidate-aware rubric specifies the concrete experiment, metrics, revision location, and diagnostic purpose required for a high-quality actionable suggestion.}
\label{tab:case_task2_generic_vs_candidate_aware}
\end{table*}
\section{Evaluation Details}
\label{app:evaluation}

This appendix provides additional details for the evaluation protocol used in Section~\ref{sec:exp}. 
We include baseline prompts, schema-controlled prompts, the LLM-as-a-Judge prompt, human annotation instructions, automatic metric definitions, claim alignment, and bootstrap confidence intervals.

\subsection{LLM-as-a-Judge Prompt}
\label{app:judge_prompts}

We use GPT-5.4 as a pairwise judge. 
For each instance, \ours-RL is compared against one baseline output under the same task input. 
The judge chooses \ours-RL, the baseline, or tie for each evaluation dimension. 
To reduce position bias, each pair is judged twice with output order swapped, and the two judgments are averaged before computing the final adjusted win rate. (prompts shown in Figure ~\ref{fig:judge-prompt-task1} and Figure ~\ref{fig:judge-prompt-task2})

Both tasks are evaluated on \textit{Technical Accuracy}, \textit{Depth \& Constructiveness}, and \textit{Overall}. 
The task-specific dimension is \textit{Specificity \& Grounding} for Task~1 and \textit{Actionability} for Task~2.


\begin{figure*}[t]
\centering
\setlength{\fboxsep}{0pt}

\begin{tcolorbox}[
    colback=gray!8,
    colframe=yellow!50!black,
    boxrule=0.6pt,
    arc=1pt,
    left=8pt,
    right=8pt,
    top=8pt,
    bottom=8pt,
    width=0.95\textwidth
]
\footnotesize
\textbf{Prompt for Task~1 Pairwise Judge}

\vspace{0.5em}
You are a neutral expert evaluator comparing two AI-generated diagnostic claims for academic peer review.

\vspace{0.6em}
\textbf{Input:}
You are given:
\begin{enumerate}[leftmargin=1.5em,itemsep=0.2em,topsep=0.2em]
    \item Paper metadata, including title and abstract.
    \item A target weakness label.
    \item The reference claim count.
    \item Assistant~A's generated diagnostic claims.
    \item Assistant~B's generated diagnostic claims.
\end{enumerate}

\vspace{0.6em}
\textbf{Goal:}
Determine which assistant provides better diagnostic claims for the given paper and weakness label.

The primary question is: which set of claims would better help authors understand the concrete weakness in the paper?

\vspace{0.6em}
\textbf{Scoring dimensions:}
Score each assistant on each dimension using an integer from 0 to 5.

\begin{itemize}[leftmargin=1.5em,itemsep=0.2em,topsep=0.2em]
    \item \textbf{Technical Accuracy}: Are the claims factually correct and supported by the paper metadata and context? Penalize hallucinated paper content, unsupported criticisms, or claims that address the wrong weakness.
    \item \textbf{Depth and Constructiveness}: Do the claims explain why the weakness matters rather than merely restating a generic concern?
    \item \textbf{Specificity and Grounding}: Do the claims refer to concrete paper elements, such as methods, experiments, datasets, assumptions, figures, tables, equations, or claims?
    \item \textbf{Overall}: Overall, how useful are the claims for helping authors understand the target weakness?
\end{itemize}

\vspace{0.6em}
\textbf{Scoring guide:}
\begin{itemize}[leftmargin=1.5em,itemsep=0.2em,topsep=0.2em]
    \item \textbf{5}: Excellent; accurate, specific, well-grounded, and highly useful.
    \item \textbf{4}: Strong; mostly accurate and useful, with only minor gaps.
    \item \textbf{3}: Adequate; partially useful but missing important specificity or depth.
    \item \textbf{2}: Weak; only loosely relevant or mostly generic.
    \item \textbf{1}: Very poor; largely unsupported, shallow, or misaligned.
    \item \textbf{0}: Fundamentally wrong, hallucinated, or not responsive to the task.
\end{itemize}

\vspace{0.6em}
\textbf{Key rules:}
\begin{itemize}[leftmargin=1.5em,itemsep=0.2em,topsep=0.2em]
    \item Judge semantic quality, not surface wording.
    \item Do not reward a longer answer unless it adds accurate and useful content.
    \item Penalize unsupported claims, hallucinated paper content, or claims unrelated to the target weakness label.
    \item The \textbf{Overall} score should reflect holistic reviewer usefulness, not simply the average of the other dimensions.
\end{itemize}

\vspace{0.6em}
\textbf{Output format:}
Return only valid JSON with integer scores from 0 to 5. Do not include reasoning.

\begin{tcolorbox}[
    colback=white,
    colframe=gray!35,
    boxrule=0.35pt,
    arc=1mm,
    left=3pt,
    right=3pt,
    top=3pt,
    bottom=3pt
]
\ttfamily\scriptsize
\{
  "technical\_accuracy": \{"score\_a": 0, "score\_b": 0\},\newline
  "depth\_and\_constructiveness": \{"score\_a": 0, "score\_b": 0\},\newline
  "specificity\_and\_grounding": \{"score\_a": 0, "score\_b": 0\},\newline
  "overall": \{"score\_a": 0, "score\_b": 0\}\newline
\}
\end{tcolorbox}

\vspace{0.6em}
\textbf{User template:}

\begin{tcolorbox}[
    colback=white,
    colframe=gray!35,
    boxrule=0.35pt,
    arc=1mm,
    left=3pt,
    right=3pt,
    top=3pt,
    bottom=3pt
]
\textbf{Paper Title:} \textit{\{title\}}

\textbf{Abstract:} \textit{\{abstract\}}

\textbf{Weakness Label:} \textit{\{weakness\_label\}}

\textbf{Reference Claim Count:} \textit{\{gt\_count\}} claims

\textbf{Assistant A's Claims:} \textit{\{claims\_a\}}

\textbf{Assistant B's Claims:} \textit{\{claims\_b\}}
\end{tcolorbox}
\end{tcolorbox}

\caption{
LLM-as-a-judge prompt template for Task~1 pairwise evaluation.
We apply the swap trick by exchanging Assistant~A and Assistant~B across two rounds, then average scores before computing the final verdict.
}
\label{fig:judge-prompt-task1}
\end{figure*}


\begin{figure*}[t]
\centering
\setlength{\fboxsep}{0pt}

\begin{tcolorbox}[
    colback=gray!8,
    colframe=yellow!50!black,
    boxrule=0.6pt,
    arc=1pt,
    left=8pt,
    right=8pt,
    top=8pt,
    bottom=8pt,
    width=0.95\textwidth
]
\footnotesize
\textbf{Prompt for Task~2 Pairwise Judge}

\vspace{0.5em}
You are a neutral expert evaluator comparing two AI-generated actionable suggestions for academic peer-review revision.

\vspace{0.6em}
\textbf{Input:}
You are given:
\begin{enumerate}[leftmargin=1.5em,itemsep=0.2em,topsep=0.2em]
    \item Paper metadata, including title and abstract.
    \item A diagnostic claim that the suggestions should address.
    \item Assistant~A's actionable suggestion.
    \item Assistant~B's actionable suggestion.
\end{enumerate}

\vspace{0.6em}
\textbf{Goal:}
Determine which assistant provides more useful revision guidance for addressing the diagnostic claim.

The primary question is: which suggestion would a real author find more useful for revising the paper?

\vspace{0.6em}
\textbf{Scoring dimensions:}
Score each assistant on each dimension using an integer from 0 to 5.

\begin{itemize}[leftmargin=1.5em,itemsep=0.2em,topsep=0.2em]
    \item \textbf{Technical Accuracy}: Is the suggestion technically sound and free of hallucinated paper content? Penalize unsupported claims, infeasible suggestions, or references to non-existent sections, figures, tables, or experiments.
    \item \textbf{Depth and Constructiveness}: Does the suggestion address the underlying weakness rather than merely restating the problem?
    \item \textbf{Actionability}: Does the suggestion provide concrete \textit{what}, \textit{where}, \textit{how}, and expected-outcome guidance that an author can directly follow?
    \item \textbf{Overall}: Overall, how useful is the suggestion as practical revision guidance?
\end{itemize}

\vspace{0.6em}
\textbf{Scoring guide:}
\begin{itemize}[leftmargin=1.5em,itemsep=0.2em,topsep=0.2em]
    \item \textbf{5}: Excellent; accurate, concrete, localized, and directly actionable.
    \item \textbf{4}: Strong; useful and mostly concrete, with only minor gaps.
    \item \textbf{3}: Adequate; partially actionable but missing important specificity.
    \item \textbf{2}: Weak; mostly generic or difficult for authors to implement.
    \item \textbf{1}: Very poor; vague, unsupported, or largely unhelpful.
    \item \textbf{0}: Fundamentally wrong, hallucinated, or not responsive to the task.
\end{itemize}

\vspace{0.6em}
\textbf{Key rules:}
\begin{itemize}[leftmargin=1.5em,itemsep=0.2em,topsep=0.2em]
    \item Specific \textit{what}, exact \textit{where}, and step-by-step \textit{how} should be preferred over longer but vague suggestions.
    \item Penalize suggestions that invent paper content or refer to unavailable evidence.
    \item Penalize suggestions that only say ``add more details'' or ``clarify the method'' without concrete implementation guidance.
    \item The \textbf{Overall} score should reflect holistic author usefulness, not simply the average of the other dimensions.
\end{itemize}

\vspace{0.6em}
\textbf{Output format:}
Return only valid JSON with integer scores from 0 to 5. Do not include reasoning.

\begin{tcolorbox}[
    colback=white,
    colframe=gray!35,
    boxrule=0.35pt,
    arc=1mm,
    left=3pt,
    right=3pt,
    top=3pt,
    bottom=3pt
]
\ttfamily\scriptsize
\{
  "technical\_accuracy": \{"score\_a": 0, "score\_b": 0\},\newline
  "depth\_and\_constructiveness": \{"score\_a": 0, "score\_b": 0\},\newline
  "actionability": \{"score\_a": 0, "score\_b": 0\},\newline
  "overall": \{"score\_a": 0, "score\_b": 0\}\newline
\}
\end{tcolorbox}

\vspace{0.6em}
\textbf{User template:}

\begin{tcolorbox}[
    colback=white,
    colframe=gray!35,
    boxrule=0.35pt,
    arc=1mm,
    left=3pt,
    right=3pt,
    top=3pt,
    bottom=3pt
]
\textbf{Paper Title:} \textit{\{title\}}

\textbf{Abstract:} \textit{\{abstract\}}

\textbf{Diagnostic Claim Being Addressed:} \textit{\{claim\}}

\textbf{Assistant A's Actionable Suggestion:} \textit{\{suggestion\_a\}}

\textbf{Assistant B's Actionable Suggestion:} \textit{\{suggestion\_b\}}
\end{tcolorbox}
\end{tcolorbox}

\caption{
LLM-as-a-judge prompt template for Task~2 pairwise evaluation.
The swap trick and verdict computation follow the same protocol as Task~1.
}
\label{fig:judge-prompt-task2}
\end{figure*}

\paragraph{Adjusted win rate.}
For each pairwise judgment, we map the decision to a score from the perspective of \ours-RL:
\[
s_i =
\begin{cases}
1, & \text{if \ours-RL wins},\\
0.5, & \text{if the comparison is a tie},\\
0, & \text{if the baseline wins}.
\end{cases}
\]
The adjusted win rate is computed as:
\[
\mathrm{AdjWinRate}
=
100 \times \frac{1}{N}\sum_{i=1}^{N} s_i.
\]
Thus, 50 indicates parity, values above 50 indicate preference for \ours-RL, and values below 50 indicate preference for the baseline.

\subsection{Human Evaluation Instructions}
\label{app:human_eval}

We conduct human evaluation using the same pairwise protocol as the GPT-5.4 judge. 
Human evaluation is performed on a 200-instance subset of \bench. 
Each comparison is independently annotated by two graduate-level annotators with peer-review experience. 
Annotators are shown the same paper context, target weakness label, task input, and two anonymized model outputs. 
Model identities are hidden, and output order is randomized.

\paragraph{Annotation task.}
For each comparison, annotators choose \ours-RL, the baseline, or tie for each evaluation dimension. 
Both tasks are evaluated on Technical Accuracy, Depth \& Constructiveness, and Overall. 
The task-specific dimension is Specificity \& Grounding for Task~1 and Actionability for Task~2.

\paragraph{Scoring.}
We map each judgment to a pairwise score from the perspective of \ours-RL:
\[
\begin{array}{rcl}
\ours\text{-RL win} &=& 1,\\
\text{tie} &=& 0.5,\quad \text{baseline win}=0.
\end{array}
\]
For each comparison, we first average the two annotators' scores. 
We then average across all comparisons and multiply by 100 to report the human adjusted win rate. 
This makes human evaluation directly comparable to the LLM-as-a-Judge results.

\paragraph{Agreement.}
The two annotators make identical categorical choices in 93\% of pairwise judgments, indicating high raw agreement. 
When annotators disagree, we retain the disagreement by averaging their pairwise scores rather than forcing a single adjudicated winner. 
For example, if one annotator selects \ours-RL and the other selects the baseline, the instance contributes 0.5, reflecting no clear preference.

\paragraph{Human annotation instructions.}
Annotators compare two anonymized model outputs for the same paper-review task. 
For each instance, annotators are given the paper metadata, retrieved paper context, target weakness category, task input, and two anonymized outputs, denoted as Output~A and Output~B. 
Annotators are asked to choose which output is better for each evaluation dimension, with three possible choices: Output~A, Output~B, or Tie.

For \textbf{Technical Accuracy}, annotators select the output that is more factually correct and better supported by the provided paper context. 
They are instructed to penalize hallucinated paper content, unsupported criticism, and infeasible suggestions. 
For \textbf{Depth and Constructiveness}, annotators select the output that provides more substantive and constructive feedback rather than shallow or generic comments. 
For \textbf{Specificity and Grounding} in Task~1, annotators select the output that better identifies concrete paper-specific weaknesses grounded in the provided context. 
For \textbf{Actionability} in Task~2, annotators select the output that better explains what should be revised, where the revision should be made, how the revision can be implemented, and what outcome the revision would achieve. 
For \textbf{Overall}, annotators select the output that would be more useful to authors.

Annotators are explicitly instructed not to prefer an output merely because it is longer, more fluent, or more formal. 
Instead, they are asked to focus on correctness, paper grounding, specificity, and usefulness for revision.

\subsection{Automatic Metrics.}
\label{app:metric_defs}

We use lightweight rule-based metrics to quantify different aspects in generated outputs.

\paragraph{Claim count.}
For a Task~1 output $y$, we define $n$-claims$(y)$ as the number of generated claim headers matching the pattern \texttt{Claim $k$:}. We report \textbf{$n$ claims} as the average of this quantity over the evaluation set.

\paragraph{Specificity Score.}
For a generated text $y$, we compute a heuristic specificity score based on three observable signals: references to localized paper content, technical terminology, and quantitative details. Let $L(y)$ denote the number of explicit paper-location references in $y$, including mentions such as \texttt{Section 3}, \texttt{Figure 2}, \texttt{Table 1}, \texttt{Algorithm 1}, \texttt{Appendix}, and \texttt{Equation 4}. Let $T(y)$ denote the number of unique all-caps technical terms or acronyms matched by the regular expression \texttt{[A-Z]\{2,\}}. Let $Q(y)$ denote the number of numeric expressions, including percentages. The specificity score is defined as
\begin{equation}
\mathrm{Spec}(y) = L(y) + 0.5\,T(y) + 0.3\,Q(y).
\end{equation}
For Task~1, we compute this score for each generated claim and report the average across claims in an instance, then average over the evaluation set. For Task~2 evidence, we apply the same formula to the extracted evidence section.

\paragraph{Suggestion similarity.}
For Task~2, let $\hat{S} = \{\hat{s}_1, \dots, \hat{s}_m\}$ denote the set of generated suggestion blocks and let $S = \{s_1, \dots, s_n\}$ denote the set of reference suggestion blocks. We compute sentence embeddings for each suggestion using the same sentence-transformer encoder as in our semantic similarity metric, and use cosine similarity as the base similarity function $\mathrm{sim}(\cdot,\cdot)$. We define suggestion similarity as
\begin{equation}
\mathrm{Sugg.sim}(y,\tilde{y}) =
\frac{1}{m}\sum_{i=1}^{m}\max_{1 \leq j \leq n} \mathrm{sim}(\hat{s}_i, s_j),
\end{equation}
where $y$ and $\tilde{y}$ are the generated and reference outputs, respectively. This metric measures whether each generated suggestion is semantically aligned with at least one reference action.

\paragraph{Evidence specificity.}
For Task~2, let $e(y)$ denote the evidence section extracted from generated output $y$. We compute evidence specificity by applying the same specificity function used for Task~1 to the extracted evidence section:
\begin{equation}
\mathrm{Evid.spec}(y) = \mathrm{Spec}(e(y)).
\end{equation}
Thus, \textbf{Evid.spec} and \textbf{Spec.} share the same underlying formula, but differ in the text unit to which they are applied: \textbf{Spec.} is computed at the claim level in Task~1, whereas \textbf{Evid.spec} is computed on the evidence section in Task~2.

\paragraph{Example-specific Rubric Score.}
In addition to generic automatic metrics and pairwise LLM-as-a-Judge comparison, we report an \emph{example-specific rubric score} (\textbf{Rubric}). For each test instance, we first generate a dedicated rubric offline based on its task type and instance context, and then use GPT-5.4 to score a model output against that rubric. The rubric consists of \emph{hard constraints} and \emph{weighted soft requirements}. If any hard constraint is violated, the score is set to zero; otherwise, GPT-5.4 scores each soft requirement on a 1--5 scale, which is linearly mapped to $[0,1]$ and aggregated by a weighted average to produce the final scalar score. In our implementation, hard constraints are checked deterministically, while semantic requirements are judged by GPT-5.4; the resulting score is stored per instance and averaged over the evaluation set. This design allows the rubric metric to capture instance-specific criteria that are not well reflected by global overlap-based metrics.

\subsection{Claim Alignment for Automatic Metrics}
\label{app:claim_alignment}

Task~1 outputs are sets of diagnostic claims, and different systems may generate different numbers of claims for the same instance. 
For example, a reference may contain two claims, while a model may generate one, three, or none. 
Directly concatenating all claims and computing ROUGE-L or semantic similarity would be misleading, because a system could be rewarded for writing extra claims or penalized for using a different claim order. 
We therefore align generated claims to reference claims before computing claim-level metrics.

Let $\hat{\mathcal{C}}=\{\hat{c}_1,\ldots,\hat{c}_m\}$ be the generated claims and $\mathcal{C}^{*}=\{c^{*}_1,\ldots,c^{*}_n\}$ be the reference claims. 
We compute a semantic similarity score $S_{ij}$ for every generated--reference pair $(\hat{c}_i,c^{*}_j)$. 
We then find the one-to-one matching that maximizes the total similarity:
\[
M^{*}=\arg\max_{M}\sum_{(i,j)\in M}S_{ij}.
\]
Each generated claim and each reference claim can appear in at most one matched pair. 
ROUGE-L, BLEU, and semantic similarity are computed over the matched pairs and then averaged. 
Generated claims that are not matched to any reference claim are treated as over-generation, while reference claims that are not matched by any generated claim are treated as under-generation. 
We report the average number of generated claims separately to make this calibration tradeoff visible.

\paragraph{Handling \texttt{None}.}
If both the generated output and reference contain no claim, the instance is counted as a correct abstention. 
If the generated output contains claims but the reference contains no claim, the output is treated as false-positive over-generation. 
If the reference contains claims but the model outputs \texttt{None}, the output is treated as under-generation.

\subsection{Bootstrap Confidence Intervals}
\label{app:ci}

We report 95\% bootstrap confidence intervals for main automatic metrics, pairwise adjusted win rates, and ablation results. 
Bootstrap resampling is performed over evaluation instances.

\paragraph{Automatic metrics.}
For automatic metrics, we sample evaluation instances with replacement and recompute the metric on each bootstrap sample. 
The 2.5th and 97.5th percentiles of the bootstrap distribution are used as the confidence interval.

\paragraph{Pairwise LLM-as-a-Judge and human evaluation.}
For pairwise evaluation, each instance contributes a score of 1, 0.5, or 0 from the perspective of \ours-RL. 
We resample instances with replacement and recompute the adjusted win rate:
\[
\mathrm{AdjWinRate}^{(b)}
=
100 \times \frac{1}{N}\sum_{i=1}^{N}s_i^{(b)}.
\]
The 95\% confidence interval is obtained from the empirical percentiles of the bootstrap distribution.

\paragraph{Interpretation.}
We avoid making strong claims from small differences whose confidence intervals overlap. 
In particular, differences below two percentage points are treated as trends rather than reliable improvements unless supported by consistent human, judge-based, and task-specific evidence.

\begin{figure*}[t]
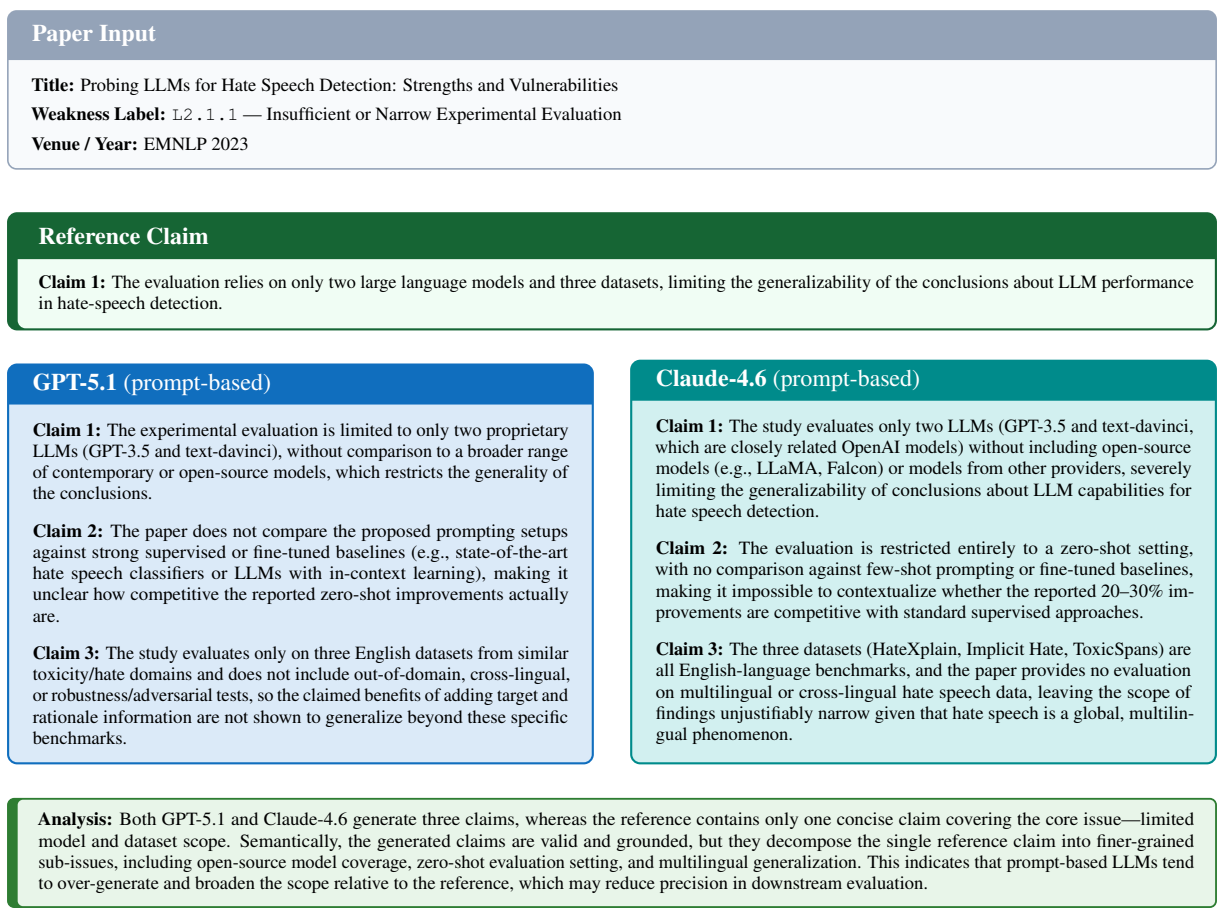

\centering

\begin{tcolorbox}[
  enhanced,
  colback=inputbg,
  colframe=inputborder,
  colbacktitle=inputborder,
  coltitle=white,
  boxrule=0.6pt,
  arc=3pt,
  left=6pt, right=6pt, top=4pt, bottom=4pt,
  fonttitle=\bfseries\footnotesize,
  title={Paper Input},
  toptitle=2pt, bottomtitle=2pt,
]
  \scriptsize
  \textbf{Title:} Probing LLMs for Hate Speech Detection: Strengths and
  Vulnerabilities\\[3pt]
  \textbf{Weakness Label:} \texttt{L2.1.1} --- Insufficient or Narrow
  Experimental Evaluation\\[3pt]
  \textbf{Venue / Year:} EMNLP 2023
\end{tcolorbox}

\vspace{4pt}

\begin{tcolorbox}[
  enhanced,
  colback=refgreenbg,
  colframe=refgreen,
  colbacktitle=refgreen,
  coltitle=white,
  boxrule=0.8pt,
  leftrule=4pt,
  arc=3pt,
  left=5pt, right=5pt, top=3pt, bottom=3pt,
  fonttitle=\bfseries\footnotesize,
  title={Reference Claim},
  toptitle=2pt, bottomtitle=2pt,
]
  \scriptsize
  \textbf{Claim 1:} The evaluation relies on only two large language models
  and three datasets, limiting the generalizability of the conclusions about
  LLM performance in hate-speech detection.
\end{tcolorbox}

\vspace{4pt}

\begin{minipage}[t]{0.485\textwidth}
  \begin{tcolorbox}[
    enhanced,
    colback=gptbluebg,
    colframe=gptblue,
    colbacktitle=gptblue,
    coltitle=white,
    boxrule=0.8pt,
    arc=3pt,
    left=6pt, right=6pt, top=4pt, bottom=4pt,
    fonttitle=\bfseries\footnotesize,
    title={GPT-5.1 \normalfont\footnotesize(prompt-based)}
  ]
    \scriptsize
    \textbf{Claim 1:} The experimental evaluation is limited to only two
    proprietary LLMs (GPT-3.5 and text-davinci), without comparison to a
    broader range of contemporary or open-source models, which restricts
    the generality of the conclusions.

    \medskip
    \textbf{Claim 2:} The paper does not compare the proposed prompting
    setups against strong supervised or fine-tuned baselines (e.g.,
    state-of-the-art hate speech classifiers or LLMs with in-context
    learning), making it unclear how competitive the reported zero-shot
    improvements actually are.

    \medskip
    \textbf{Claim 3:} The study evaluates only on three English datasets
    from similar toxicity/hate domains and does not include out-of-domain,
    cross-lingual, or robustness/adversarial tests, so the claimed benefits
    of adding target and rationale information are not shown to generalize
    beyond these specific benchmarks.
  \end{tcolorbox}
\end{minipage}%
\hfill%
\begin{minipage}[t]{0.485\textwidth}
  \begin{tcolorbox}[
    enhanced,
    colback=claudetealbg,
    colframe=claudeteal,
    colbacktitle=claudeteal,
    coltitle=white,
    boxrule=0.8pt,
    arc=3pt,
    left=6pt, right=6pt, top=4pt, bottom=4pt,
    fonttitle=\bfseries\footnotesize,
    title={Claude-4.6 \normalfont\footnotesize(prompt-based)}
  ]
    \scriptsize
    \textbf{Claim 1:} The study evaluates only two LLMs (GPT-3.5 and
    text-davinci, which are closely related OpenAI models) without
    including open-source models (e.g., LLaMA, Falcon) or models from
    other providers, severely limiting the generalizability of conclusions
    about LLM capabilities for hate speech detection.

    \medskip
    \textbf{Claim 2:} The evaluation is restricted entirely to a zero-shot
    setting, with no comparison against few-shot prompting or fine-tuned
    baselines, making it impossible to contextualize whether the reported
    20--30\% improvements are competitive with standard supervised approaches.

    \medskip
    \textbf{Claim 3:} The three datasets (HateXplain, Implicit Hate,
    ToxicSpans) are all English-language benchmarks, and the paper provides
    no evaluation on multilingual or cross-lingual hate speech data, leaving
    the scope of findings unjustifiably narrow given that hate speech is a
    global, multilingual phenomenon.
  \end{tcolorbox}
\end{minipage}

\vspace{3pt}
\begin{tcolorbox}[
  enhanced,
  colback=analysisbg,
  colframe=analysisborder,
  boxrule=0.7pt,
  leftrule=4pt,
  arc=2pt,
  left=5pt, right=5pt, top=2pt, bottom=2pt,
]
  \scriptsize
  \textbf{Analysis:} Both GPT-5.1 and Claude-4.6 generate three claims,
  whereas the reference contains only one concise claim covering the core
  issue---limited model and dataset scope. Semantically, the generated claims
  are valid and grounded, but they decompose the single reference claim into
  finer-grained sub-issues, including open-source model coverage, zero-shot
  evaluation setting, and multilingual generalization. This indicates that
  prompt-based LLMs tend to over-generate and broaden the scope relative to
  the reference, which may reduce precision in downstream evaluation.
\end{tcolorbox}

\caption{Task~1 diagnostic claim generation: GPT-5.1 and Claude-4.6 outputs
compared against the reference claim. The gray box shows the paper input;
the green box shows the rebuttal-derived reference claim.}
\label{fig:gpt_claude_task1}
\end{figure*}

\begin{figure*}[t]
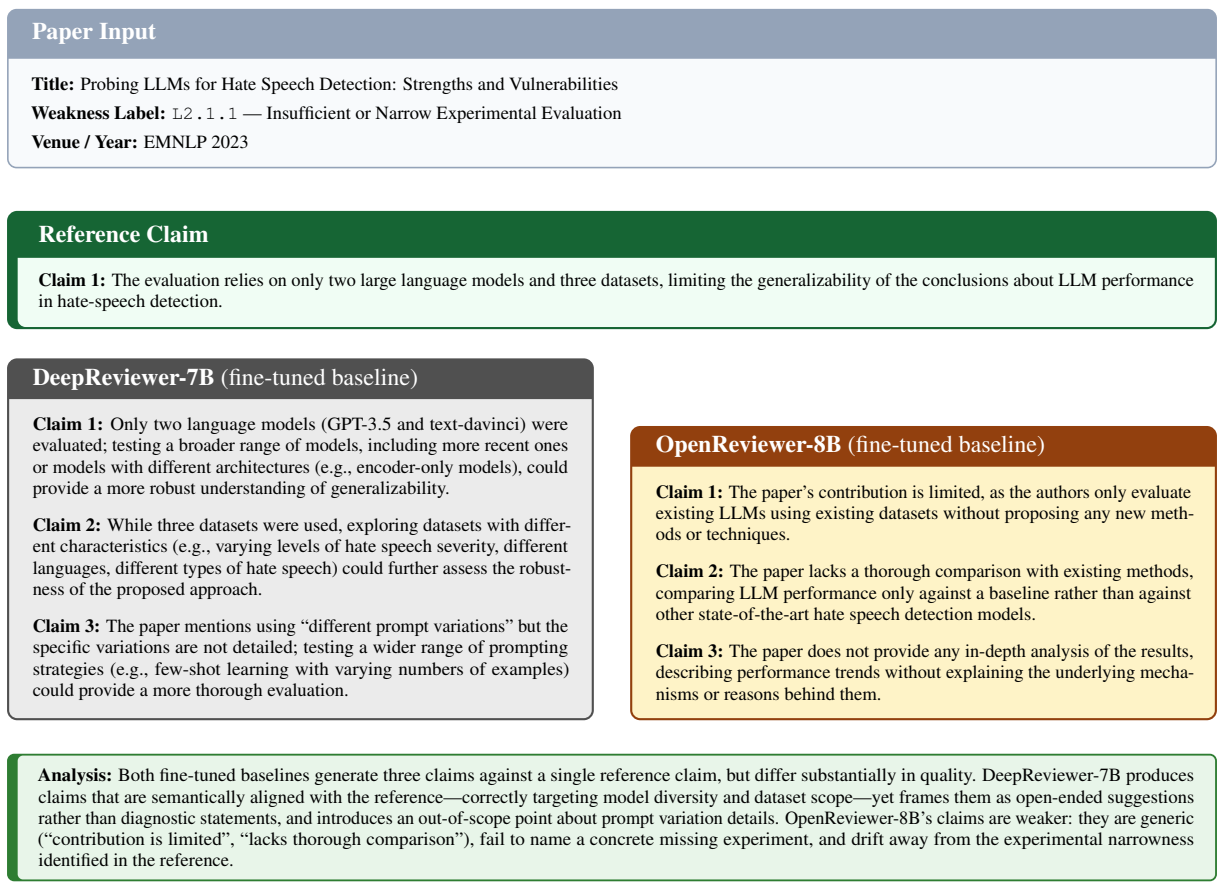

\centering

\begin{tcolorbox}[
  enhanced,
  colback=inputbg,
  colframe=inputborder,
  colbacktitle=inputborder,
  coltitle=white,
  boxrule=0.6pt,
  arc=3pt,
  left=6pt, right=6pt, top=4pt, bottom=4pt,
  fonttitle=\bfseries\footnotesize,
  title={Paper Input},
  toptitle=2pt, bottomtitle=2pt,
]
  \scriptsize
  \textbf{Title:} Probing LLMs for Hate Speech Detection: Strengths and
  Vulnerabilities\\[3pt]
  \textbf{Weakness Label:} \texttt{L2.1.1} --- Insufficient or Narrow
  Experimental Evaluation\\[3pt]
  \textbf{Venue / Year:} EMNLP 2023
\end{tcolorbox}

\vspace{4pt}

\begin{tcolorbox}[
  enhanced,
  colback=refgreenbg,
  colframe=refgreen,
  colbacktitle=refgreen,
  coltitle=white,
  boxrule=0.8pt,
  leftrule=4pt,
  arc=3pt,
  left=5pt, right=5pt, top=3pt, bottom=3pt,
  fonttitle=\bfseries\footnotesize,
  title={Reference Claim},
  toptitle=2pt, bottomtitle=2pt,
]
  \scriptsize
  \textbf{Claim 1:} The evaluation relies on only two large language models
  and three datasets, limiting the generalizability of the conclusions about
  LLM performance in hate-speech detection.
\end{tcolorbox}

\vspace{4pt}

\begin{minipage}[t]{0.485\textwidth}
  \begin{tcolorbox}[
    enhanced,
    colback=deepgraybg,
    colframe=deepgray,
    colbacktitle=deepgray,
    coltitle=white,
    boxrule=0.8pt,
    arc=3pt,
    left=6pt, right=6pt, top=4pt, bottom=4pt,
    fonttitle=\bfseries\footnotesize,
    title={DeepReviewer-7B \normalfont\footnotesize(fine-tuned baseline)}
  ]
    \scriptsize
    \textbf{Claim 1:} Only two language models (GPT-3.5 and text-davinci)
    were evaluated; testing a broader range of models, including more
    recent ones or models with different architectures (e.g., encoder-only
    models), could provide a more robust understanding of generalizability.

    \medskip
    \textbf{Claim 2:} While three datasets were used, exploring datasets
    with different characteristics (e.g., varying levels of hate speech
    severity, different languages, different types of hate speech) could
    further assess the robustness of the proposed approach.

    \medskip
    \textbf{Claim 3:} The paper mentions using ``different prompt
    variations'' but the specific variations are not detailed; testing a
    wider range of prompting strategies (e.g., few-shot learning with
    varying numbers of examples) could provide a more thorough evaluation.
  \end{tcolorbox}
\end{minipage}%
\hfill%
\begin{minipage}[t]{0.485\textwidth}
  \begin{tcolorbox}[
    enhanced,
    colback=openbrownbg,
    colframe=openbrown,
    colbacktitle=openbrown,
    coltitle=white,
    boxrule=0.8pt,
    arc=3pt,
    left=6pt, right=6pt, top=4pt, bottom=4pt,
    fonttitle=\bfseries\footnotesize,
    title={OpenReviewer-8B \normalfont\footnotesize(fine-tuned baseline)}
  ]
    \scriptsize
    \textbf{Claim 1:} The paper's contribution is limited, as the authors
    only evaluate existing LLMs using existing datasets without proposing
    any new methods or techniques.

    \medskip
    \textbf{Claim 2:} The paper lacks a thorough comparison with existing
    methods, comparing LLM performance only against a baseline rather than
    against other state-of-the-art hate speech detection models.

    \medskip
    \textbf{Claim 3:} The paper does not provide any in-depth analysis of
    the results, describing performance trends without explaining the
    underlying mechanisms or reasons behind them.
  \end{tcolorbox}
\end{minipage}

\vspace{3pt}
\begin{tcolorbox}[
  enhanced,
  colback=analysisbg,
  colframe=analysisborder,
  boxrule=0.7pt,
  leftrule=4pt,
  arc=2pt,
  left=5pt, right=5pt, top=2pt, bottom=2pt,
]
  \scriptsize
  \textbf{Analysis:} Both fine-tuned baselines generate three claims
  against a single reference claim, but differ substantially in quality.
  DeepReviewer-7B produces claims that are semantically aligned with the
  reference---correctly targeting model diversity and dataset scope---yet
  frames them as open-ended suggestions rather than diagnostic statements,
  and introduces an out-of-scope point about prompt variation details.
  OpenReviewer-8B's claims are weaker: they are generic (``contribution is
  limited'', ``lacks thorough comparison''), fail to name a concrete missing
  experiment, and drift away from the experimental narrowness identified in
  the reference.
\end{tcolorbox}

\caption{Task~1 diagnostic claim generation: DeepReviewer-7B and
OpenReviewer-8B outputs compared against the reference claim. The gray box
shows the paper input; the green box shows the rebuttal-derived reference claim.}
\label{fig:deep_open_task1}
\end{figure*}
\section{Additional Evaluations and Analyses}
\label{app:additional_analysis}

This section provides additional experiments examining the effect of the
two-task formulation, generalization beyond rebuttal-derived evaluation
references, robustness to independent judges, scientific validity and
technical failures, the limitations of localized context, and model behavior
when the original rebuttal provides an incomplete resolution.

\subsection{End-to-End versus Two-Task Formulation}
\label{app:formulation_ablation}

To isolate the effect of separating weakness diagnosis from revision
planning, we compare our two-task SFT formulation against an end-to-end SFT
variant. Both variants use the same Qwen3-8B-Base backbone, training
instances, optimization settings, and evaluation protocol. The end-to-end
variant jointly generates a diagnostic claim and its actionable suggestion
in a single output. The two-task variant first generates a diagnostic claim
and then conditions suggestion generation on that generated claim.

We evaluate both variants before rubric-based reinforcement learning so that
the comparison isolates the effect of task decomposition rather than the
candidate-aware reward design. Outputs are scored using the same GPT-5.4
rubric on Task~1 Specificity and Grounding, Task~1 Overall quality,
Task~2 Actionability, Task~2 Overall quality, and Claim--Suggestion
Alignment. The final metric assesses whether the proposed revision directly
and coherently addresses the diagnosed weakness.



As shown in Table~\ref{tab:formulation_ablation}, the two-task formulation
improves all five evaluation dimensions. The largest improvement occurs in
Task~2 Overall quality, which increases by 0.42 points, while
Claim--Suggestion Alignment increases by 0.35 points. These results indicate
that explicitly conditioning revision planning on a separately generated
diagnostic claim strengthens the connection between the identified weakness
and the proposed action, rather than merely changing the output format.

\subsection{Independent Evaluation on 2025--2026 Papers}
\label{app:independent_eval}

ActReview is trained on review--rebuttal interactions and the original
\bench{} evaluation references are also grounded in author-side resolution
signals. To test whether the learned behavior transfers beyond
rebuttal-derived references and to examine temporal generalization, we
construct an independent held-out evaluation set from OpenReview papers
submitted in 2025--2026.

The evaluation set contains 175 instances from 50 papers that are not used in
ActReview training or in the construction of the original benchmark.
Independent reviewers inspect each paper without access to its original
reviews, rebuttals, or discussion threads. They identify a paper-supported
weakness, assign the corresponding Level-2 weakness label, and specify the
requirements that a sufficient resolution should satisfy. This procedure
prevents the evaluation targets from inheriting the content or phrasing of
the original author rebuttals.

For a controlled comparison, all evaluated systems receive the same paper
context, metadata, and weakness label. Each model first generates its Task~1
diagnostic claim, which is then provided as the input claim for Task~2.
The paired outputs are evaluated jointly as a complete reviewer comment,
thereby capturing both diagnostic quality and error propagation from Task~1
to Task~2.

Two graduate-level annotators independently rate each anonymized output
without access to model identity or the original review--rebuttal discussion.
The outputs are scored on a three-point scale, where 0 indicates that the
criterion is not satisfied, 1 indicates partial satisfaction, and 2 indicates
full satisfaction. The four criteria are:

\begin{itemize}
    \item \textbf{Weakness Validity}: whether the diagnostic claim identifies
    a genuine weakness supported by the paper;
    \item \textbf{Technical Correctness}: whether the diagnosis and proposed
    revision are scientifically and technically sound;
    \item \textbf{Resolution Sufficiency}: whether the recommendation
    adequately resolves the identified weakness; and
    \item \textbf{Reviewer Usefulness}: whether the complete comment would
    provide useful guidance to an author revising the paper.
\end{itemize}

The mean quadratic-weighted Cohen's $\kappa$ across the four dimensions is
0.71, indicating substantial inter-annotator agreement.

\begin{table*}[t]
\centering
\small
\setlength{\tabcolsep}{5pt}
\renewcommand{\arraystretch}{1.15}
\begin{tabular}{lcccc}
\toprule
\multirow{2}{*}{\textbf{Model}}
& \multicolumn{2}{c}{\textbf{Diagnostic Quality}}
& \multicolumn{2}{c}{\textbf{Revision Quality}} \\
\cmidrule(lr){2-3} \cmidrule(lr){4-5}
& \textbf{Weakness Validity}
& \textbf{Technical Correctness}
& \textbf{Resolution Sufficiency}
& \textbf{Reviewer Usefulness} \\
\midrule
GPT-5.1$_{\mathrm{schema}}$
& 1.67 & 1.64 & 1.23 & 1.34 \\

Gemini$_{\mathrm{schema}}$
& 1.51 & 1.62 & 1.39 & 1.29 \\

\textsc{\ours-SFT}
& 1.67 & 1.61 & 1.51 & 1.61 \\

\textsc{\ours-RL}
& \textbf{1.68}
& \textbf{1.66}
& \textbf{1.70}
& \textbf{1.68} \\
\bottomrule
\end{tabular}
\caption{
Independent human evaluation on 175 instances from 50 held-out
2025--2026 OpenReview papers constructed without rebuttal guidance.
Task~1 and Task~2 outputs are evaluated jointly as complete reviewer
comments. Values are mean ratings on a 0--2 scale; higher is better.
}
\label{tab:independent_recent_eval}
\end{table*}

Table~\ref{tab:independent_recent_eval} shows that
\textsc{\ours-RL} remains comparable to
GPT-5.1$_{\mathrm{schema}}$ in Weakness Validity
(1.68 versus 1.67) and Technical Correctness
(1.66 versus 1.64). Its larger gains occur in revision-oriented dimensions:
Resolution Sufficiency increases from 1.23 to 1.70, and Reviewer Usefulness
increases from 1.34 to 1.68. \textsc{\ours-SFT} also outperforms the
proprietary baselines on these two dimensions.

These results suggest that rebuttal-guided supervision transfers beyond
evaluation against rebuttal-derived references. ActReview remains competitive
in independently assessed diagnostic and technical quality while generating
more sufficient and useful revision guidance. This experiment provides
evidence of temporal and supervision-source generalization, but it does not
establish generalization across venues, disciplines, or substantially
different peer-review norms.

\subsection{Cross-Judge Robustness}
\label{app:cross_judge}

GPT-5.4 is involved in data enhancement, rubric construction, semantic reward
scoring, and scalable pairwise evaluation. This repeated use raises the
possibility that \textsc{\ours-RL} is optimized toward evaluator-specific
preferences. We therefore conduct an additional cross-judge analysis using
Claude, which is not involved in data construction, training, reward scoring,
or rubric generation.

We use the same 200-instance subset employed for the original human
evaluation and compare \textsc{\ours-RL} against RbtAct,
DeepReviewer-14B, and Gemini$_{\mathrm{schema}}$. GPT-5.4, Claude, and
human annotators independently evaluate the same output pairs. Model
identities are hidden, output order is randomized, and each judge selects
\textsc{\ours-RL}, the baseline, or a tie. We report adjusted win rates,
where a win contributes 1, a tie contributes 0.5, and a loss contributes 0.

\begin{table*}[t]
\centering
\small
\setlength{\tabcolsep}{4.5pt}
\renewcommand{\arraystretch}{1.12}
\begin{tabular}{llcccc}
\toprule
\multirow{2}{*}{\textbf{Baseline}}
& \multirow{2}{*}{\textbf{Judge}}
& \multicolumn{4}{c}{\textbf{Task 1: Diagnostic Claims}} \\
\cmidrule(lr){3-6}
&
& \textbf{Tech. Acc.}
& \textbf{Depth}
& \textbf{Spec./Gnd.}
& \textbf{Overall} \\
\midrule
\multirow{3}{*}{RbtAct}
& GPT-5.4 & 65.3 & 56.2 & 60.3 & 59.1 \\
& Claude  & 64.1 & 61.8 & 61.0 & 59.7 \\
& Human   & 64.4 & 57.8 & 58.7 & 62.0 \\
\midrule
\multirow{3}{*}{DeepReviewer-14B}
& GPT-5.4 & 76.4 & 70.2 & 70.7 & 73.6 \\
& Claude  & 75.3 & 71.5 & 68.3 & 72.9 \\
& Human   & 76.2 & 71.9 & 66.1 & 72.8 \\
\midrule
\multirow{3}{*}{Gemini$_{\mathrm{schema}}$}
& GPT-5.4 & 54.3 & 59.9 & 61.1 & 60.5 \\
& Claude  & 56.3 & 60.2 & 64.2 & 62.3 \\
& Human   & 54.9 & 58.1 & 63.7 & 59.9 \\
\bottomrule
\end{tabular}
\caption{
Cross-judge pairwise evaluation for Task~1 on the same 200-instance
human-evaluation subset. Values are adjusted win rates of
\textsc{\ours-RL} against each baseline; 50 indicates parity.
}
\label{tab:cross_judge_task1}
\end{table*}

\begin{table*}[t]
\centering
\small
\setlength{\tabcolsep}{4.5pt}
\renewcommand{\arraystretch}{1.12}
\begin{tabular}{llcccc}
\toprule
\multirow{2}{*}{\textbf{Baseline}}
& \multirow{2}{*}{\textbf{Judge}}
& \multicolumn{4}{c}{\textbf{Task 2: Actionable Suggestions}} \\
\cmidrule(lr){3-6}
&
& \textbf{Tech. Acc.}
& \textbf{Depth}
& \textbf{Action.}
& \textbf{Overall} \\
\midrule
\multirow{3}{*}{RbtAct}
& GPT-5.4 & 58.2 & 71.7 & 54.8 & 59.4 \\
& Claude  & 60.6 & 70.6 & 56.5 & 58.2 \\
& Human   & 58.3 & 67.5 & 54.5 & 59.8 \\
\midrule
\multirow{3}{*}{DeepReviewer-14B}
& GPT-5.4 & 70.1 & 68.7 & 83.6 & 68.5 \\
& Claude  & 73.5 & 69.5 & 78.0 & 71.4 \\
& Human   & 69.4 & 70.9 & 77.3 & 71.9 \\
\midrule
\multirow{3}{*}{Gemini$_{\mathrm{schema}}$}
& GPT-5.4 & 54.2 & 56.8 & 63.6 & 54.3 \\
& Claude  & 60.9 & 55.1 & 64.1 & 55.6 \\
& Human   & 56.1 & 59.2 & 64.9 & 54.7 \\
\bottomrule
\end{tabular}
\caption{
Cross-judge pairwise evaluation for Task~2 on the same 200-instance
human-evaluation subset. Values are adjusted win rates of
\textsc{\ours-RL} against each baseline; 50 indicates parity.
}
\label{tab:cross_judge_task2}
\end{table*}

Across all three baselines and both tasks, GPT-5.4, Claude, and human
evaluation assign \textsc{\ours-RL} adjusted win rates above 50.
The direction and magnitude of the comparisons are also similar across the
three judges. Across the reported aggregate baseline--dimension comparisons,
GPT-5.4 and Claude achieve Spearman correlations of
$\rho=0.943$ and $\rho=0.899$ with human evaluation, respectively.
Their mean absolute deviations from human win rates are similarly close:
1.81 points for GPT-5.4 and 1.88 points for Claude.

We additionally measure instance-level categorical agreement on Overall
judgments after mapping each decision to an \textsc{\ours-RL} win, tie,
or baseline win.

\begin{table}[t]
\centering
\small
\setlength{\tabcolsep}{5.5pt}
\renewcommand{\arraystretch}{1.15}
\begin{tabular}{lcccc}
\toprule
\textbf{Judge Pair}
& \textbf{T1 Agr.}
& \textbf{T1 $\kappa$}
& \textbf{T2 Agr.}
& \textbf{T2 $\kappa$} \\
\midrule
GPT-5.4 vs.\ Human
& 80.97\% & 0.63 & 82.67\% & 0.65 \\

Claude vs.\ Human
& 81.68\% & 0.63 & 79.74\% & 0.62 \\

GPT-5.4 vs.\ Claude
& 86.49\% & 0.73 & 83.21\% & 0.68 \\
\bottomrule
\end{tabular}
\caption{
Instance-level categorical agreement on Overall pairwise judgments across
the same 200-instance subset and three representative baselines.
Agr.\ denotes exact agreement, and Cohen's $\kappa$ adjusts for agreement
expected by chance.
}
\label{tab:cross_judge_agreement}
\end{table}

As shown in Table~\ref{tab:cross_judge_agreement}, GPT-5.4 and Claude
achieve similar levels of agreement with human judgments. Claude is slightly
closer to humans on Task~1, whereas GPT-5.4 is slightly closer on Task~2.
There is therefore no systematic advantage suggesting that GPT-5.4 uniquely
favors \textsc{\ours-RL} because of its role in the training pipeline.

This analysis cannot completely eliminate training-side coupling because
GPT-5.4 remains involved in enhancement and reward construction. However,
the consistency of GPT-5.4, Claude, and human judgments reduces the likelihood
that the reported evaluation gains are primarily an artifact of
GPT-5.4-specific scoring preferences.

\subsection{Scientific Validity and Technical Error Analysis}
\label{app:technical_errors}

We conduct two complementary analyses to determine whether ActReview improves
substantive scientific quality or primarily improves the presentation and
executability of revision guidance. The first separates scientific validity
from revision utility. The second categorizes the remaining technical
failures at a finer level.

\subsubsection{Scientific Validity versus Revision Utility}
\label{app:scientific_validity}

We sample a stratified subset of 100 \bench{} instances and compare
anonymized outputs from GPT-5.1, GPT-5.1$_{\mathrm{schema}}$,
\textsc{\ours-SFT}, and \textsc{\ours-RL}. Output order is randomized,
and two annotators familiar with machine-learning peer review independently
rate each output on a 0--2 scale.

\textbf{Diagnostic Validity} measures whether the identified weakness is
genuinely present and supported by the paper. \textbf{Scientific Validity}
measures whether the proposed resolution is technically sound and does not
rely on invalid scientific assumptions. Exact inter-annotator agreement is
81\% for Diagnostic Validity and 76\% for Scientific Validity, with
quadratic-weighted Cohen's $\kappa$ values of 0.74 and 0.68, respectively.
Disagreements are resolved through adjudication.

We additionally record whether the output includes a concrete and executable
revision step. From these annotations, we derive two rates:

\begin{itemize}
    \item \textbf{Valid + Actionable}: the percentage of outputs that receive
    a Scientific Validity score of 2 and provide a concrete, executable
    revision; and
    \item \textbf{Actionable but Invalid}: the percentage of actionable
    outputs that receive a Scientific Validity score of 0.
\end{itemize}

\begin{table*}[t]
\centering
\small
\setlength{\tabcolsep}{5pt}
\renewcommand{\arraystretch}{1.15}
\begin{tabular}{lcccc}
\toprule
\multirow{2}{*}{\textbf{Model}}
& \multicolumn{2}{c}{\textbf{Scientific Validity}}
& \multicolumn{2}{c}{\textbf{Revision Utility}} \\
\cmidrule(lr){2-3} \cmidrule(lr){4-5}
& \textbf{Diagnostic Validity}
& \textbf{Scientific Validity}
& \textbf{Valid + Actionable (\%)}
& \textbf{Actionable but Invalid (\%)} \\
\midrule
GPT-5.1
& 1.68 & 1.70 & 55 & 18 \\

GPT-5.1$_{\mathrm{schema}}$
& 1.66 & 1.64 & 55 & 11 \\

\textsc{\ours-SFT}
& 1.71 & 1.67 & 61 & 20 \\

\textsc{\ours-RL}
& \textbf{1.72}
& \textbf{1.71}
& \textbf{74}
& \textbf{10} \\
\bottomrule
\end{tabular}
\caption{
Blinded human evaluation of scientific validity and revision utility on
100 stratified \bench{} instances.
Diagnostic Validity and Scientific Validity are mean ratings on a 0--2
scale. Valid + Actionable reports the percentage of outputs that are
scientifically valid and provide a concrete, executable revision.
Actionable but Invalid reports the percentage of actionable outputs that
contain a technically invalid recommendation. Higher is better except for
Actionable but Invalid.
}
\label{tab:scientific_validity_utility}
\end{table*}

As shown in Table~\ref{tab:scientific_validity_utility},
\textsc{\ours-RL} does not exhibit a large advantage over GPT-5.1 in
Scientific Validity (1.71 versus 1.70). We therefore do not interpret the
results as evidence of universally superior deep scientific judgment.
The larger benefit lies in the reliable conversion of valid diagnoses into
technically sound and executable revisions. \textsc{\ours-RL} produces the
highest Valid + Actionable rate at 74\%, compared with 55\% for GPT-5.1
and 61\% for \textsc{\ours-SFT}. It also produces the lowest
Actionable but Invalid rate at 10\%.

\subsubsection{Fine-Grained Technical Failures}
\label{app:fine_grained_errors}

We further analyze technical failures on the 175-instance independent
evaluation set described in Appendix~\ref{app:independent_eval}.
For each model, annotators assign non-exclusive failure labels to every
output with a Technical Correctness score below 2. Because labels are
non-exclusive, one output may exhibit multiple failure types.

The failure categories are defined as follows:

\begin{itemize}
    \item \textbf{Unsupported weakness}: the diagnosed concern is not
    substantiated by the paper;
    \item \textbf{Method misunderstanding}: the output mischaracterizes the
    proposed method, assumptions, or mechanism;
    \item \textbf{Invalid experimental recommendation}: the proposed
    experiment is inappropriate, confounded, or unable to test the stated
    claim;
    \item \textbf{Incorrect theoretical reasoning}: the output contains a
    substantively incorrect theoretical statement or derivation;
    \item \textbf{Context omission}: the output overlooks evidence,
    qualifications, or conditions already present in the paper;
    \item \textbf{Incomplete technical specification}: the proposed revision
    omits information required for correct execution;
    \item \textbf{Overstated necessity}: the output presents one plausible
    revision as strictly required despite the availability of alternative
    resolutions.
\end{itemize}

\textbf{Any technical error} denotes an output with a Technical Correctness
score below 2. \textbf{Severe technical error} denotes an output receiving
a Technical Correctness score of 0.

\begin{table*}[t]
\centering
\small
\setlength{\tabcolsep}{4.2pt}
\renewcommand{\arraystretch}{1.12}
\begin{tabular}{lcccc}
\toprule
\textbf{Technical Failure Type}
& \textbf{GPT-5.1$_{\mathrm{schema}}$}
& \textbf{Gemini$_{\mathrm{schema}}$}
& \textbf{\ours-SFT}
& \textbf{\ours-RL} \\
\midrule
Unsupported weakness
& 4.57\% & 10.29\% & 4.00\% & \textbf{2.86\%} \\

Method misunderstanding
& 8.00\% & 9.71\% & 8.57\% & \textbf{7.43\%} \\

Invalid experimental recommendation
& \textbf{2.86\%} & 4.00\% & 5.71\% & 3.43\% \\

Incorrect theoretical reasoning
& 3.43\% & 2.86\% & \textbf{2.29\%} & \textbf{2.29\%} \\

Context omission
& 9.71\% & 8.57\% & 9.14\% & \textbf{7.43\%} \\

Incomplete technical specification
& \textbf{15.43\%} & 16.57\% & 17.14\% & \textbf{15.43\%} \\

Overstated necessity
& \textbf{5.71\%} & 10.29\% & 19.43\% & 17.14\% \\
\midrule
Any technical error
& 29.14\% & \textbf{28.00\%} & 29.14\% & \textbf{28.00\%} \\

Severe technical error
& 6.86\% & 10.29\% & 9.71\% & \textbf{6.29\%} \\
\bottomrule
\end{tabular}
\caption{
Technical-failure rates on 175 independent evaluation instances per model.
Failure categories are non-exclusive, so one output may exhibit multiple
failure types. Any technical error denotes an output with a Technical
Correctness score below 2, whereas Severe technical error denotes a score
of 0. Lower is better.
}
\label{tab:technical_error_analysis}
\end{table*}

Table~\ref{tab:technical_error_analysis} shows that
\textsc{\ours-RL} has an overall technical-error rate of 28.00\%, matching
Gemini$_{\mathrm{schema}}$ and slightly improving over
GPT-5.1$_{\mathrm{schema}}$ and \textsc{\ours-SFT}, both at 29.14\%.
It also produces the lowest severe-error rate at 6.29\%.

Relative to \textsc{\ours-SFT}, rubric-based reinforcement learning reduces
unsupported weaknesses, method misunderstandings, invalid experimental
recommendations, context omissions, and severe technical errors. These
results indicate that the improvements in actionability and resolution
sufficiency are not accompanied by a higher overall rate of technical
failure.

The primary remaining failure mode is Overstated Necessity.
\textsc{\ours-RL} presents a particular revision as mandatory in 17.14\%
of cases, compared with 5.71\% for
GPT-5.1$_{\mathrm{schema}}$ and 10.29\% for
Gemini$_{\mathrm{schema}}$. Although reinforcement learning reduces this
rate relative to \textsc{\ours-SFT}, the revision-oriented supervision may
still encourage overly prescriptive recommendations. Future rubric designs
should explicitly reward conditional language and recognition of multiple
valid resolution paths.

\subsection{Global-Context Analysis}
\label{app:global_context}

Localized context can reduce irrelevant input and improve weakness-specific
grounding, but some review concerns require evidence distributed across
multiple sections of a paper. To examine this limitation, we select 50
instances requiring global paper understanding. These instances include
motivation--experiment mismatches, method--evaluation inconsistencies,
unsupported conclusions, and pipeline-level logic gaps.

We compare two SFT variants trained with either localized retrieved chunks
or full-paper text. All other model, data, and optimization settings are
held fixed. Importantly, both variants are evaluated under the same
full-paper inference protocol, so the comparison measures the effect of
training-time context construction rather than differences in evaluation-time
information.

Human annotators rate each output on a 0--2 scale for Weakness Validity,
Cross-Section Reasoning, Technical Correctness, and Resolution Sufficiency.
We also report Unsupported Claim Rate, which measures the percentage of
outputs containing a substantive claim that is not supported by the paper.

\begin{table*}[t]
\centering
\scriptsize
\setlength{\tabcolsep}{3pt}
\renewcommand{\arraystretch}{1.12}
\begin{tabular*}{\textwidth}{
@{\extracolsep{\fill}}
lccccc
@{}
}
\toprule
\textbf{Training Context}
& \shortstack{\textbf{Weakness}\\\textbf{Validity}}
& \shortstack{\textbf{Cross-Section}\\\textbf{Reasoning}}
& \shortstack{\textbf{Technical}\\\textbf{Correctness}}
& \shortstack{\textbf{Resolution}\\\textbf{Sufficiency}}
& \shortstack{\textbf{Unsupported}\\\textbf{Claim Rate} $\downarrow$} \\
\midrule
Retrieved chunks
& 1.39
& 1.17
& 1.60
& \textbf{1.51}
& 22\% \\

Full-paper text
& \textbf{1.41}
& \textbf{1.49}
& \textbf{1.62}
& 1.47
& \textbf{14\%} \\
\bottomrule
\end{tabular*}
\caption{
Human evaluation on 50 instances requiring global paper understanding.
Models are trained with localized retrieved chunks or full-paper text and
evaluated under the same full-paper inference protocol.
Ratings use a 0--2 scale; higher is better except for Unsupported Claim Rate.
}
\label{tab:global_context_analysis}
\end{table*}

As shown in Table~\ref{tab:global_context_analysis}, full-paper training
provides a clear advantage in Cross-Section Reasoning, increasing the mean
score from 1.17 to 1.49. It also reduces Unsupported Claim Rate from 22\%
to 14\%. Weakness Validity and Technical Correctness remain comparable
between the two settings, while retrieved-chunk training obtains slightly
higher Resolution Sufficiency.

These findings indicate that localized and global training contexts provide
complementary benefits. Localized context remains effective for
weakness-specific concerns by focusing supervision on relevant evidence,
whereas full-paper context is more suitable when a valid diagnosis requires
integrating information across sections. Localized retrieval should therefore
not be interpreted as universally preferable to full-paper conditioning.

\subsection{Hard Rebuttal-Incomplete Cases}
\label{app:hard_cases}

The main benchmark focuses on cases with an evidence-supported revision path,
but the aligned review--rebuttal pool also contains more difficult cases in
which the original author response is incomplete, only partially addresses
the concern, or proposes an inadequate resolution. We analyze a targeted set
of 50 such cases to determine whether \textsc{\ours-RL} merely reproduces
the author response or can assess and improve upon it.

Each generated response is assigned to one of four mutually exclusive
behavior categories:

\begin{itemize}
    \item \textbf{Follows a sufficient rebuttal}: the original response
    already contains a sufficient resolution and the model proposes a
    consistent revision path;
    \item \textbf{Extends an incomplete rebuttal}: the model preserves the
    valid part of the author response while adding information or actions
    required to resolve the concern;
    \item \textbf{Rejects an inadequate rebuttal path}: the model avoids an
    insufficient author-side response and proposes a more appropriate
    resolution; and
    \item \textbf{Introduces an unsupported resolution}: the model proposes
    a revision that is not supported by the reviewer concern or paper
    content.
\end{itemize}

\begin{table}[t]
\centering
\small
\setlength{\tabcolsep}{6pt}
\renewcommand{\arraystretch}{1.15}
\begin{tabular}{lc}
\toprule
\textbf{Model Behavior}
& \textbf{Rate (\%)} \\
\midrule
Follows a sufficient rebuttal
& 28 \\
Extends an incomplete rebuttal
& 46 \\
Rejects an inadequate rebuttal path
& 20 \\
Introduces an unsupported resolution
& 6 \\
\bottomrule
\end{tabular}
\caption{
Behavior analysis of \textsc{\ours-RL} on 50 hard cases in which the
original rebuttal provides an incomplete or insufficient resolution.
Each output is categorized according to whether it follows, extends, rejects,
or introduces a resolution path.
}
\label{tab:hard_case_analysis}
\end{table}

Table~\ref{tab:hard_case_analysis} shows that
\textsc{\ours-RL} does not simply reproduce author responses.
In 46\% of cases, it extends an incomplete rebuttal, and in another 20\%,
it rejects an inadequate resolution path. Thus, in 66\% of the hard cases,
the model improves upon or departs from the original author-side response.
It introduces an unsupported resolution in 6\% of cases.

\paragraph{Representative case.}
One example concerns a paper proposing a density-based uncertainty
categorization framework. The reviewer questions both the novelty of the
method relative to existing feature-space density approaches and the
practical usefulness of the proposed Bnd and IDM categories. The original
rebuttal states only that the introduction, figure, and method sections will
be revised to clarify the motivation. This response does not establish
empirical novelty or demonstrate that the proposed categories provide
distinct practical information.

Rather than reproducing this clarification-only response,
\textsc{\ours-RL} recommends a controlled comparison with a standard
feature-space density baseline, an ablation isolating the contribution of
the confusion-based component, and an evaluation testing whether the Bnd and
IDM categories provide complementary information. The resulting suggestion
extends the original rebuttal by proposing concrete evidence needed to
substantiate both novelty and practical utility while remaining grounded in
the reviewer concern and paper content.

This analysis does not establish performance on fundamentally
non-resolvable concerns such as irreparable methodological flaws or purely
subjective novelty disputes. Evaluating such concerns would require a
different benchmark design in which abstention, rejection of the method, or
multiple incompatible expert judgments can be treated as valid outcomes;
we take a first step in this direction in Appendix~\ref{app:abstention}.

\section{Held-out Evaluation of Zero-Shot Abstention}
\label{app:abstention}

\subsection{Motivation and Relation to Reviewer Concerns}
\label{app:abstention_motivation}

The task formulation in Section~\ref{sec:problem_formulation} requires
abstention: when the paper context does not support the target weakness
label, the model must return no claim ($k{=}0$) rather than fabricate a
weakness. Because every instance in \data and \bench is derived from a
weakness that a real reviewer raised, neither resource contains
non-supported (weakness, paper) pairs. We therefore do \emph{not} train
on negative instances. Instead we evaluate abstention as a held-out,
zero-shot behavior: does the weakness-conditioned instruction, together
with the grounding pressure introduced during training, induce correct
abstention on labels the paper does not exhibit?

Several properties make this a meaningful test rather than an accident of
prompting. First, judging whether a body of text contains evidence for a
specific claim is a capability the Qwen3-8B-Base backbone already
possesses from pretraining; abstention does not require a dedicated
training signal. Second, the instruction ``if the paper has no weakness
under this label, output exactly \texttt{None}'' appears in every SFT and
RL prompt, so the model is repeatedly exposed to the rule even though it
never sees a \texttt{None} target. Third, both training stages push the
model toward asserting only what the retrieved context supports: SFT
targets are tightly grounded in retrieved chunks, and the RL rubric
assigns zero reward to claims that hallucinate paper content or address
the wrong label (Appendix~\ref{app:rubric}). Abstention is the natural
endpoint of that constraint when no supporting evidence exists. The
opposing force is a prior toward always answering, induced by training
exclusively on supported instances; whether grounding pressure overcomes
that prior is an empirical question, which this evaluation answers.

\subsection{Constructing the Non-Supported Split}
\label{app:abstention_construction}

\paragraph{Distractor labels.}
We start from the \bench papers. For each paper we take $W_{\text{pos}}$,
the set of Level-2 weakness labels that reviewers raised for that paper.
We then sample \emph{distractor} labels from the complement
$\mathcal{W}\setminus W_{\text{pos}}$ over the 17-label taxonomy,
preferring labels whose Level-1 parent category does not appear in
$W_{\text{pos}}$, since a weakness type from an entirely unraised
category is more likely to be genuinely absent. We target a balanced
distribution of roughly 20 instances per Level-2 label.

\paragraph{Full-paper inference.}
For each non-supported paper--label pair, we use the same full-paper
inference protocol as in the main \bench evaluation. The only difference
between supported and non-supported instances is whether the queried
weakness label is supported by the paper.

\paragraph{Verification that the label is absent.}
A distractor label is only a valid negative if the paper truly does not
exhibit that weakness. We use a label-absence screening prompt adapted
from the retrieval-quality protocol in
Figure~\ref{fig:retrieval_quality_prompt}. Given the same full-paper
context used for inference and the target label definition, GPT-5.4
classifies each pair as \texttt{present}, \texttt{absent}, or
\texttt{uncertain}; only pairs classified as \texttt{absent} are
retained. For conditional labels such as L2.7.1 (Missing Computational
Cost, Runtime, and Scalability Analysis), whose taxonomy definition
applies only when the paper makes a claim that depends on the missing
content (Appendix~\ref{app:taxonomy}), the prompt explicitly checks both
this materiality condition and whether the corresponding content is
adequately reported anywhere in the paper, including appendices, before
deciding absence -- so that a paper is not marked \texttt{present}
merely for lacking a discussion it never needed, nor for content it
already reports outside the main text. The complete prompt is provided
in
Figure~\ref{fig:abstention_screening_prompt}. The final split contains
$N_{\text{neg}}{=}360$ operationally non-supported pairs retained after
GPT-5.4 screening. As a check on this screen, an author with
peer-review experience inspected a 29-instance random sample of the
retained pairs; the human spot-check agrees with the screening decision
on 27 of 29 sampled pairs (93.1\%).

\begin{figure*}[t]
\centering
\begin{tcolorbox}[
  colback=gray!5,
  colframe=gray!45,
  width=0.96\textwidth,
  fontupper=\small,
  title={Label-Absence Screening Prompt},
  breakable
]
You are evaluating whether a paper exhibits a specific weakness type, for
the purpose of constructing a non-supported (paper, label) evaluation
pair.

You are given:
\begin{enumerate}[leftmargin=*,nosep]
    \item Paper metadata, including title and abstract.
    \item The full paper content, including all appendices.
    \item A target weakness label and its taxonomy definition.
\end{enumerate}

Note: no reviewer raised this weakness label for this paper; you are
judging directly against the paper content, not against a reviewer
complaint.

Your task is to judge whether the paper exhibits the target weakness,
based only on the provided paper content and the label definition. Do
not use or assume access to the author rebuttal or to any reviewer
comments.

\textbf{Conditional labels.}
Some labels are conditional: they apply only if the paper makes a claim,
contribution, or assumption for which the missing content would be
material (for example, a missing-efficiency-analysis label applies only
if the paper makes an efficiency, real-time, or large-scale applicability
claim). For such labels, follow two checks in order:
\begin{enumerate}[leftmargin=*,nosep]
    \item \emph{Materiality.} Does the paper make a claim the missing
    content would be material to? If not, classify the label as
    \texttt{absent}.
    \item \emph{Adequacy.} If it does, search the entire paper --
    including appendices, implementation-detail sections, and
    supplementary results -- for the corresponding content. If it is
    reported there in adequate detail, the paper does not exhibit the
    weakness even though the claim is made; classify the label as
    \texttt{absent}. Only classify \texttt{present} if the claim is made
    and the corresponding content is missing or inadequate everywhere in
    the paper.
\end{enumerate}

\textbf{Decision.}
Return one of:
\begin{itemize}[leftmargin=*]
    \item \texttt{present}: the paper content provides clear evidence of
    the target weakness.
    \item \texttt{absent}: the paper content provides no evidence of the
    target weakness.
    \item \texttt{uncertain}: the paper content is ambiguous or
    insufficient to decide.
\end{itemize}

Return only valid JSON in the following format:
\begin{verbatim}
{
  "decision": "present/absent/uncertain",
  "reason": "short explanation"
}
\end{verbatim}
\end{tcolorbox}
\caption{Label-absence screening prompt used to construct the
non-supported split (Appendix~\ref{app:abstention_construction}).
Adapted from the retrieval-quality screening prompt
(Figure~\ref{fig:retrieval_quality_prompt}); unlike that prompt, this one
outputs a single \texttt{present}/\texttt{absent}/\texttt{uncertain}
decision on whether the paper exhibits the queried label, rather than
pass/fail judgments on retrieval sufficiency.}
\label{fig:abstention_screening_prompt}
\end{figure*}

\paragraph{Scope.}
``Absent'' here means the full-paper context provides no evidence for the
label, verified by an LLM screen and a human sample; it is an
operational, not exhaustive, notion. Systematically characterizing every
weakness a paper does not exhibit is outside the scope of this
evaluation.

\subsection{Metrics}
\label{app:abstention_metrics}

We frame abstention as a binary decision (answer vs.\ abstain) and report,
as percentages:
\begin{itemize}[leftmargin=*,nosep]
  \item \textbf{Answer Rate} (supported): fraction of the 1{,}000 \bench
  Task-1 instances where the model produced at least one claim.
  \item \textbf{False-Abstention Rate} (supported): fraction where the
  model returned \texttt{None} for a reviewer-raised positive instance.
  This is computed on the Task-1 outputs already generated for the main
  experiments and requires no new inference.
  \item \textbf{Abstention Rate} (non-supported): fraction of the
  $N_{\text{neg}}$ instances where the model correctly returned
  \texttt{None}.
  \item \textbf{Over-generation Rate} (non-supported): fraction where the
  model fabricated a claim ($1-$Abstention Rate).
  \item \textbf{Balanced Accuracy}: mean of Answer Rate and Abstention
  Rate.
\end{itemize}
All models receive the same abstention-aware task instruction and the
same input context. After stripping whitespace and formatting tokens,
outputs equal to \texttt{None} are counted as abstentions; all other
non-empty outputs are counted as answers.

\subsection{Results}
\label{app:abstention_results}

\begin{table}[t]
\centering
\small
\setlength{\tabcolsep}{5pt}
\renewcommand{\arraystretch}{1.15}
\begin{tabular}{l cc cc c}
\toprule
\multirow{2}{*}{\textbf{Model}}
& \multicolumn{2}{c}{\textbf{Supported} (\bench, $n{=}1000$)}
& \multicolumn{2}{c}{\textbf{Non-supported} ($n{=}360$)}
& \multirow{2}{*}{\shortstack{\textbf{Balanced}\\\textbf{Acc.}\,$\uparrow$}} \\
\cmidrule(lr){2-3}\cmidrule(lr){4-5}
& \textbf{Answer}\,$\uparrow$ & \textbf{False-Abst.}\,$\downarrow$
& \textbf{Abstention}\,$\uparrow$ & \textbf{Over-gen.}\,$\downarrow$ & \\
\midrule
\rowcolor{grpllm}
GPT-5.1$_{\mathrm{schema}}$      & 98.1 & 1.9 & 31.4 & 68.6 & 64.7 \\
\rowcolor{grpother}
Qwen3-32B                        & 96.7 & 3.3 & 39.7 & 60.3 & 68.2 \\
\rowcolor{grpother}
DeepReviewer-14B                 & 95.0 & 5.0 & 12.2 & 87.8 & 53.6 \\
\midrule
\rowcolor{grpours}
\textsc{\ours}-SFT               & 96.2 & 3.8 & 71.4 & 28.6 & 83.8 \\
\rowcolor{grpours}
\textsc{\ours}-RL                & 95.4 & 4.6 & 78.3 & 21.7 & 86.9 \\
\bottomrule
\end{tabular}
\caption{Zero-shot abstention on non-supported weakness--paper pairs.
Negatives are a held-out split of distractor weakness labels not raised
by any reviewer of the corresponding paper, screened label-absent by
GPT-5.4 with a human
spot-check ($N_{\text{neg}}{=}360$). All values are percentages. \emph{Answer}: produced $\ge 1$ claim
for a reviewer-raised target weakness in \bench; \emph{False-Abstention}:
returned \texttt{None} for a reviewer-raised positive instance (computed
on existing Task~1 outputs, no new inference); \emph{Abstention}: correctly returned \texttt{None};
\emph{Over-generation}: fabricated a claim under a non-supported label;
\emph{Balanced Acc.}: mean of Answer and Abstention rates.}
\label{tab:abstention}
\end{table}

Table~\ref{tab:abstention} shows that all models answer supported
instances reliably: Answer Rate is at least 95\%, and the
False-Abstention Rate is at most 5\%, so no model abstains
indiscriminately. The models diverge sharply on non-supported
instances. Prompt-based and specialized review generators rarely abstain:
GPT-5.1$_{\mathrm{schema}}$ fabricates a weakness under
68.6\% of non-supported labels and DeepReviewer-14B under
87.8\%, consistent with a strong prior that a queried
label always corresponds to a real problem and with the fact that
generic criticism can be produced for almost any label. \textsc{\ours}
abstains far more often: \textsc{\ours}-SFT returns \texttt{None} on
71.4\% of non-supported instances and \textsc{\ours}-RL on
78.3\%, while keeping the False-Abstention Rate on
supported instances at 4.6\%. The gain from RL over SFT
(+6.9 points of Abstention Rate) is consistent with the
rubric's hard penalty on wrong-label and hallucinated claims, which
rewards withholding an unsupported claim. Balanced Accuracy improves
from 68.2\% for the strongest evaluated non-\ours baseline, Qwen3-32B,
to 86.9\% for \textsc{\ours}-RL, an absolute improvement of
18.7 percentage points.

\textsc{\ours}-SFT and \textsc{\ours}-RL exhibit substantially stronger
zero-shot abstention than the evaluated baselines. Moreover, RL improves
the abstention rate over SFT by 6.9 percentage points, consistent with
the grounding-oriented reward design. However, \textsc{\ours}-RL still
over-generates on 21.7\% of screened non-supported pairs, motivating
explicit negative training and evaluation on harder within-category
distractors.

\subsection{Case Study}
\label{app:abstention_cases}

Figure~\ref{fig:abstention_cases} presents two successful abstention
cases and one representative over-generation failure. In Cases 1 and 2,
\textsc{\ours}-RL correctly returns \texttt{None} because the full-paper
context shows the queried weakness does not apply, either because the
paper makes no claim the label would be material to, or because the
paper makes such a claim but adequately supports it elsewhere in the
text. In Case 3, the
model incorrectly infers a formatting-related weakness from the
discussion of Table 7, illustrating the residual over-generation
behavior reflected by the 21.7\% error rate in
Table~\ref{tab:abstention}.

\begin{figure*}[t]
\centering

\begin{tcolorbox}[breakable, colback=graybg, colframe=grayborder,
  fontupper=\small, title={Case 1 -- correct abstention \textnormal{(paper \detokenize{kmbG9iBRIb})}}]
\textbf{Paper:} \emph{Accountability in Offline Reinforcement Learning:
Explaining Decisions with a Corpus of Examples.} Contribution: an offline
controller that produces accountable decisions by selecting a corpus
subset of supporting examples, evaluated in simulated and real-world
healthcare settings.

\textbf{Queried label (not raised by reviewers of this paper):}
\textcolor{errorred}{L2.7.1 -- Missing Computational Cost, Runtime, and
Scalability Analysis.}

\textbf{Full-paper context:} the paper claims an efficient algorithm
that ``circumvented the computational difficulty'' and supports this
claim with a dedicated scalability analysis (Appendix D.3) and a
hardware/running-time report (Appendix D.5: 2 TITAN X GPUs, 32 Intel
E5-2640 CPUs; convex-hull decomposition under 10 seconds with
sub-linear scaling under parallelization). The efficiency claim is
adequately supported, so L2.7.1 does not apply.

\textbf{Reference:} \textcolor{refgreen}{\texttt{None}} \quad\textbar\quad
\textbf{\ours-RL:} \textcolor{refgreen}{\texttt{None}}
\end{tcolorbox}

\begin{tcolorbox}[breakable, colback=graybg, colframe=grayborder,
  fontupper=\small, title={Case 2 -- correct abstention \textnormal{(paper \detokenize{7wk9PqiiW2D})}}]
\textbf{Paper:} \emph{ProsodyBERT: Self-Supervised Prosody Representation
for Style-Controllable TTS.} Contribution: a self-supervised prosody
representation learned by masked prediction over clustered
prosody features, improving controllability and emotion recognition.

\textbf{Queried label (not raised by reviewers of this paper):}
\textcolor{errorred}{L2.7.1 -- Missing Computational Cost, Runtime, and
Scalability Analysis.}

\textbf{Full-paper context:} the paper claims computational efficiency
(``we reduce the model size and feature vector ... for computational
efficiency'') and supports this claim with an architecture and training
cost report (Appendix D: 6-layer, 21M-parameter encoder; trained for 10
iterations on 8 GPUs, taking approximately 6 hours for the emotion
recognition configuration and 10 hours for the TTS configuration). The
efficiency claim is adequately supported, so L2.7.1 does not apply.

\textbf{Reference:} \textcolor{refgreen}{\texttt{None}} \quad\textbar\quad
\textbf{\ours-RL:} \textcolor{refgreen}{\texttt{None}}
\end{tcolorbox}

\begin{tcolorbox}[breakable, colback=graybg, colframe=grayborder,
  fontupper=\small, title={Case 3 -- over-generation failure \textnormal{(paper \detokenize{87Ks7PvYVJi})}}]
\textbf{Paper:} \emph{Offline Decentralized Multi-Agent Reinforcement
Learning.} Contribution: value-deviation and transition-normalization
corrections for offline MARL, with a convergence result and empirical
gains on cooperative tasks.

\textbf{Queried label (not raised by reviewers of this paper):}
\textcolor{errorred}{L2.6.2 -- Formatting, Figures, or Submission
Issues.}

\textbf{Full-paper context:} the paper reports extrapolation-error
results in Table 7; reviewers raised experimental, clarity, novelty, and
theoretical-assumption concerns, but none flagged formatting, figure
quality, or submission compliance.

\textbf{Reference:} \textcolor{refgreen}{\texttt{None}} \quad\textbar\quad
\textbf{\ours-RL:} \textcolor{errorred}{\texttt{Claim 1: The extrapolation-error results in Table 7 are presented without consistent axis labeling or unit annotation, making the reported error magnitudes difficult for readers to interpret at a glance.}}
\end{tcolorbox}

\caption{Examples from the screened non-supported split. Cases 1 and 2
illustrate correct abstention, whereas Case 3 shows a residual
over-generation failure. Queried distractor labels and the wrong
inference are shown in \textcolor{errorred}{red}; correct \texttt{None}
outputs are shown in \textcolor{refgreen}{green}.}
\label{fig:abstention_cases}
\end{figure*}

\end{document}